\documentclass[11pt,twoside]{article}
\usepackage{fullpage}

\usepackage{epsf}
\usepackage{fancyhdr}
\usepackage{graphics}
\usepackage{graphicx}
\usepackage{psfrag}
\usepackage{microtype}
\usepackage{subfigure}
\usepackage{algorithmic}
\usepackage{color,xcolor}
\usepackage{caption}
\usepackage{subcaption}
\usepackage{subfigure}

\usepackage[linesnumbered,ruled]{algorithm2e}
\DontPrintSemicolon

\usepackage{color}

\usepackage{amsthm}
\usepackage{amsfonts}
\usepackage{amsmath}
\usepackage{amssymb,bbm}

\usepackage{mathrsfs}

\usepackage{url}
\usepackage[colorlinks,linkcolor=magenta,citecolor=blue, pagebackref=true,backref=true]{hyperref}
\renewcommand*{\backrefalt}[4]{%
    \ifcase #1 \footnotesize{(Not cited.)}%
    \or        \footnotesize{(Cited on page~#2.)}%
    \else      \footnotesize{(Cited on pages~#2.)}%
    \fi}

\usepackage{nicefrac}
\usepackage{comment}

\usepackage{chngpage}

 \usepackage{tabularx}%
\usepackage{enumitem}
\usepackage{booktabs}
\usepackage{caption}

\usepackage{bm,bbm}
\usepackage{mathtools}
\usepackage{cleveref}

\usepackage{eucal}

\usepackage{scalerel}

\usepackage{final_macros}

\newcommand{\integers}{\mathbb{Z}}

\newcommand{\state}{x}

\newcommand{\action}{a}

\newcommand{\MyState}{X}

\newcommand{\Action}{A}

\newcommand{\actionspace}{\mathbb{A}}

\newcommand{\policy}{\pi}

\renewcommand{\widebar}{\overline}

\newtheorem{assumption}{Assumption}
\newcommand{\numobs}{\ensuremath{n}}

\newcommand{\usedim}{\ensuremath{d}}

\newcommand{\Reward}{R}
\newcommand{\reward}{r}

\newcommand{\mixingtime}{t_{\mathrm{mix}}}

\newenvironment{carlist}
 {\begin{list}{$\bullet$}
 {\setlength{\topsep}{0in} \setlength{\partopsep}{0in}
  \setlength{\parsep}{0in} \setlength{\itemsep}{\parskip}
  \setlength{\leftmargin}{0.07in} \setlength{\rightmargin}{0.08in}
  \setlength{\listparindent}{0in} \setlength{\labelwidth}{0.08in}
  \setlength{\labelsep}{0.1in} \setlength{\itemindent}{0in}}}
 {\end{list}}

\newcommand{\lspace}{\LinSpace}

\newcommand{\projecttolin}{\Pi_{\lspace}}

\newcommand{\projecttolinhil}{\Pi_{\lspace, \mathbb{H}^1}}

\newcommand{\ltwospace}{\mathbb{L}^2}
\newcommand{\statespace}{\mathbb{X}}

\newcommand{\totaltime}{T}

\newcommand{\fbar}{\ensuremath{\bar{f}}}

\newcommand{\stationary}{\xi}

\newcommand{\filtration}{\mathcal{F}}

\newcommand{\fakerefassumelip}[1]{\hyperref[assume:smooth-high-order]{{\color{magenta} {\upshape\textbf (}{\upshape{\textbf{Lip}}#1}{\upshape\textbf )}}} }

\newcommand{\fakerefassumehyper}[2]{\hyperref[assume:hyper-contractivity]{{\color{magenta} {\upshape\textbf (}{\upshape{\textbf{Hyper}}(#1, #2)}{\upshape\textbf )}}} }

\newcommand{\discount}{\beta}

\newcommand{\valuefunc}{f}

\newcommand{\ValueFunc}{\valuefunc}

\newcommand{\ValFun}{\valuefunc}
\newcommand{\ValFunc}{\valuefunc}
\newcommand{\valfun}{\valuefunc}

\DeclareFontFamily{U}{mathx}{}
\DeclareFontShape{U}{mathx}{m}{n}{<-> mathx10}{}
\DeclareSymbolFont{mathx}{U}{mathx}{m}{n}

\DeclareMathAccent{\widecheck}{0}{mathx}{"71}
\newcommand{\BellOp}{\ensuremath{\mathcal{T}}}

\newcommand{\valuebar}{\widebar{\valuefunc}}

\newcommand{\valuehat}{\widehat{\valuefunc}}

\newcommand{\lammin}{\lambda_{\min}}
\newcommand{\lammax}{\lambda_{\max}}

\newcommand{\Event}{\mathcal{E}}

\usepackage{mathbbol}

\long\def\comment#1{}

\newcommand{\mbasis}{m}

\newcommand{\constScaryReg}{\constScary_{\mathrm{reg}}}

\newcommand{\statnorm}[1]{\|#1\|_{\ltwospace(\stationary)}}
\newcommand{\statinprod}[2]{\inprod{#1}{#2}_{\ltwospace(\stationary)}}

\newcommand{\ValPol}{\ensuremath{\ValFun^\star}}
\newcommand{\ValTrue}{\ValPol}

\newcommand{\ValHat}{\ensuremath{\widehat{\ValFun}}}

\newcommand{\StateSpace}{\statespace}
\newcommand{\ActionSpace}{\actionspace}

\newcommand{\SigMkv}{\Sigma_{\mathrm{Mkv}}}
\newcommand{\SigMG}{\Sigma_{\mathrm{MG}}}

\newcommand{\ctrDrift}{g}

\newcommand{\Abar}{\widebar{A}}
\newcommand{\bbar}{\overline{b}}

\newcommand{\Ahat}{\widehat{A}}
\newcommand{\bhat}{\widehat{b}}
\newcommand{\epshat}{\widehat{\varepsilon}}

\usepackage[framemethod=TikZ]{mdframed}

\newcommand{\myframe}[1]{
\begin{mdframed}[backgroundcolor=black!1, roundcorner=5pt]
  #1
\end{mdframed}
}

\newenvironment{narrowpara}
  {\par\addvspace{\smallskipamount}\narrower\noindent\ignorespaces}
  {\par\addvspace{\smallskipamount}}

\newcommand{\coordinate}{e}

\newcommand{\myassumption}[4]{
  \setlist[enumerate,1]{leftmargin=#4}
\myframe{
	\begin{enumerate}[label={ \bf{{{(#1)}}}}]
		\item \label{#2} {#3}
	\end{enumerate}
        }
}

\newcommand{\torus}{\mathbb{T}}

\newcommand{\LinSpace}{\mathbb{K}}

\newcommand{\IdMat}{\mathcal{I}}

\newcommand{\drift}{b}
\newcommand{\covMat}{\Lambda}

\newcommand{\BM}{B}

\newcommand{\Wass}{\mathcal{W}}

\newcommand{\semigroup}{\mathcal{P}}
\newcommand{\generator}{\mathcal{A}}
\newcommand{\constScary}{\widebar{C}}

\newcommand{\numerOrder}{\nu}
\newcommand{\sobonorm}[1]{\vecnorm{#1}{\mathbb{H}^1 (\stationary)}}
\newcommand{\advFunc}{q}

\newcommand{\advFuncHat}{\widehat{\advFunc}}

\newcommand{\soboinprod}[2]{\inprod{#1}{#2}_{\mathbb{H}^1 (\stationary)}}
\newcommand{\State}{\MyState}

\newcommand{\dt}{\stepsize}
\newcommand{\op}{\mathrm{op}}
\newcommand{\ftil}{\widetilde{f}}
\newcommand{\lpstatnorm}[2]{\vecnorm{#1}{\mathbb{L}^{#2} (\stationary)}}
\newcommand{\sobopstatnorm}[2]{\vecnorm{#1}{\mathbb{W}^{1, #2} (\stationary)}}
\newcommand{\constreg}{c_{\mathrm{reg}}}
\newcommand{\sobokpstatnorm}[3]{\vecnorm{#1}{\mathbb{W}^{#2, #3} (\stationary)}}
\newcommand{\sobohilnorm}[2]{\vecnorm{#1}{\mathbb{H}^{#2} (\stationary)}}
\newcommand{\sigstar}{\sigma^*}
\newcommand{\Mkv}{\mathrm{Mkv}}
\newcommand{\highorder}{\mathcal{H}}
\newcommand{\poincare}{\rho_*}

\newcommand{\Tthres}{T_{\mathrm{thres}}}

\newcommand{\fanchor}{f_0}
\newcommand{\thetanchor}{\theta_0}

\newcommand{\Delstar}{\Delta^*}

\begin{document}

\begin{center}
{\bf{\LARGE{Statistical guarantees for continuous-time policy evaluation: blessing of ellipticity and new tradeoffs}}}

\vspace*{.2in}
{\large{
 \begin{tabular}{cc}
  Wenlong Mou$^{ \diamond}$ 
 \end{tabular}

}

\vspace*{.2in}

 \begin{tabular}{c}
 Department of Statistical Sciences, University of Toronto$^{\diamond}$
 \end{tabular}

}

\begin{abstract}
  We study the estimation of the value function for continuous-time Markov diffusion processes using a single, discretely observed ergodic trajectory. Our work provides non-asymptotic statistical guarantees for the least-squares temporal-difference (LSTD) method, with performance measured in the first-order Sobolev norm. Specifically, the estimator attains an $O(1 / \sqrt{T})$ convergence rate when using a trajectory of length $T$; notably, this rate is achieved as long as $T$ scales nearly linearly with both the mixing time of the diffusion and the number of basis functions employed.
  
  A key insight of our approach is that the ellipticity inherent in the diffusion process ensures robust performance even as the effective horizon diverges to infinity. Moreover, we demonstrate that the Markovian component of the statistical error can be controlled by the approximation error, while the martingale component grows at a slower rate relative to the number of basis functions. By carefully balancing these two sources of error, our analysis reveals novel trade-offs between approximation and statistical errors.
\end{abstract}
\end{center}

\section{Introduction}
The past decade has seen great success of reinforcement learning (RL) in various data-driven decision making problems, ranging from video games~\cite{openai2019dota}, robotics~\cite{recht2019tour}, finance~\cite{wang2020continuous}, to fine-tuning of foundation models~\cite{uehara2024understanding}. In many of these applications, the environment evolves continuously over time, and the agent interacts with the environment in a continuous-time manner. Similar to the Markov decision process (MDP) framework in discrete-time, the continuous-time controlled diffusion processes provide a natural framework for modeling such continuous-time decision-making problems. With discrete-time observations from the continuous-time dynamics, the continuous-time RL problem can be viewed as a discrete-time MDP, allowing us to apply standard techniques. In particular, the model-free RL algorithms offers flexibility of function approximation. By fitting the value function and/or control policy with powerful statistical learning models including neural networks, one can efficiently learn the optimal decisions in high-dimensional and complex environments.

Despite the empirical success, however, the theoretical understanding of continuous-time RL algorithms is still in its infancy. In particular, when applied to continuous-time diffusion processes, the statistical guarantees for value learning algorithms are largely unknown. The theoretical gap also leads to practical limitations, as the fundamental tradeoffs in the choice of function approximations, discretization step length, and the trajectory length remain elusive.

In this work, we aim to bridge this gap by providing sharp statistical guarantees for value function estimation in continuous-time diffusion processes. Concretely, we consider the following time-homogeneous Markov diffusion
\begin{align}
  d \MyState_t = \drift (\MyState_t) dt + \covMat (\MyState_t)^{1/2} d \BM_t,\label{eq:cts-time-process}
\end{align}
where $(\BM_t: t \geq 0)$ is a $\usedim$-dimensional standard Brownian motion, and the functions $(\drift, \covMat)$ are coefficients in the equation. Of our interest is the value function $\ValTrue$, defined as the infinite-horizon discounted reward-to-go
\begin{align}
  \ValTrue (\state) \mydefn \Exs \Big[ \int_0^\infty e^{- \discount s} \reward (\MyState_s) ds \Big| \MyState_0 = \state \Big],
\end{align}
for a discount rate $\discount > 0$ and a reward function $\reward$. It is known that the value function $\ValTrue$ satisfies the Bellman equation
\begin{subequations}
\begin{align}
  \ValTrue (\state) = \int_0^t \Exs \big[ e^{- \discount s} \reward (\State_s) \mid \State_0 = \state\big] ds + e^{- \discount t} \Exs \big[ \ValTrue (\MyState_t) \mid \State_0 = \state \big], \quad \mbox{for any }t > 0,\label{eq:defn-bellman-eq}
\end{align}
which also leads to the second-order elliptic equation in the $t \rightarrow 0$ limit
\begin{align}
  \discount \ValTrue = \inprod{\drift}{\nabla \ValTrue} + \frac{1}{2} \mathrm{Tr} \big( \covMat \nabla^2 \ValTrue \big) + \reward.\label{eq:defn-elliptic-eq}
\end{align}
\end{subequations}

We aim to estimate the value function $\ValTrue$ from discrete-time observations of the diffusion process~\eqref{eq:cts-time-process}, with a snapshot observation of the state $\State_t$ and random reward $\Reward_t = \Reward_t (\State_t)$ every $\stepsize > 0$ time interval.

A popular and practical choice for solving this problem is model-free RL with function approximation. This approach allows us to fit the value function $\ValTrue$ directly using a function class, without the need to explicitly estimate the coefficients $(\drift, \covMat)$ and the reward function $\reward$. In particular, the temporal difference (TD) and least-square temporal difference (LSTD) algorithms~\cite{sutton1988learning,bradtke1996linear,boyan2002technical} use empirical data to solve the projected version of the Bellman fixed-point equation~\eqref{eq:defn-bellman-eq}. See the monographs~\cite{sutton2018reinforcement,szepesvari2022algorithms} for a comprehensive overview. When applied to the continuous-time diffusion process~\eqref{eq:cts-time-process}, LSTD-type algorithms have been developed and analyzed with population-level approximation and asymptotic convergence guarantees~\cite{doya1995temporal,jia2022policyeval,kobeissi2023temporal,mou2024bellman}. Nevertheless, in order to provide relevant guidance to practitioners, two key questions remain open:
\begin{enumerate}
  \item As the discretization stepsize $\stepsize$ goes to zero, the effective horizon of the discrete-time RL problem goes to infinity, and the fluctuations in the random observations become relatively large. In such a regime, can we still have meaningful statistical guarantees for the estimated value function?
  \item The choice of function class is used to balance the approximation error and the statistical error. Are the fundamental tradeoffs for continuous-time policy evaluation algorithms different from the discrete-time counterparts?
\end{enumerate}
In this paper, we answer these questions by providing sharp statistical guarantees for the empirical projected LSTD estimator and its variants. Our main contributions are as follows:
\begin{itemize}
  \item We provide the first non-asymptotic analysis of the LSTD algorithm using a single trajectory of the process~\eqref{eq:cts-time-process}. The error is measured under Sobolev norm, a natural norm that facilitates the estimation of advantage functions and policy gradients. By utilizing the elliptic structures in the process~\eqref{eq:cts-time-process}, we show a non-asymptotic convergence rate of order $1 / \sqrt{T}$ with a trajectory of length $T$, as long as $T$ is mildly larger than the mixing time of the process and the number of basis functions. Unlike the discrete-time counterparts, the convergence rate is independent of the effective horizon.
  \item By refined bound on the problem-dependent pre-factor in the convergence rate, our results reveal non-standard tradeoffs between the approximation error and the statistical error for continuous-time policy evaluation. Unlike discrete-time RL~\cite{mou2023optimal,mou2024optimal}, as the number of basis functions increases, the leading-order term (from the Markovian structure) in the statistical rate will \emph{decrease} together with the approximation error. An additional term (from the martingale structure) increases with the number of basis functions, but at a slower rate. This leads to a non-monotonic tradeoff between the approximation error and the statistical errors, and suggests more radical choices in the value function approximation.
\end{itemize}

The rest of this paper is organized as follows: we first discuss notations and related works. We then set up the problem concretely in Section~\ref{sec:preliminaries}. The main results and their implications are presented in Section~\ref{sec:main-results}. We provide the proof of the results in Section~\ref{sec:proofs}, and conclude with discussions in Section~\ref{sec:discussion}.

\paragraph{Notations:} Here we summarize some notation used throughout the paper. We use $(B_t: t \geq 0)$ to denote $d$-dimensional standard Brownian motion. For a
positive integer $m$, we define the set $[m] \defn \{1,2, \cdots,
m\}$. We use $\vecnorm{\cdot}{2}$ to denote the standard Euclidean norm in finite-dimensional spaces. For a matrix $A$, we use $\matsnorm{A}{F}$ and $\opnorm{A}$ to denote the Frobenius norm and operator norm, respectively. We use $\{e_j\}_{j=1}^\usedim$ to denote the standard basis vectors in
the Euclidean space $\real^\usedim$, i.e., $e_i$ is a vector with a $1$ in the $i$-th coordinate and zeros elsewhere. Given a positive semidefinite matrix $A \in \real^{n \times n}$, we denote the weighted norm $\vecnorm{x}{A} \mydefn \sqrt{x^\top A x}$ for any vector $x \in \real^n$, and we denote the weighted operator norm $\matsnorm{B}{\op, A} \mydefn \sup_{\vecnorm{x}{A} \leq  1} \vecnorm{Bx}{A}$ for any other matrix $B \in \real^{n \times n}$.

Given a probability measure $\stationary$, we denote $\statinprod{f}{g} \mydefn \Exs_{X \sim \stationary} \big[ f (X) g (X) \big]$, and we denote $\statnorm{f} \mydefn \statinprod{f}{f}^{1/2}$. We similarly define the Sobolev inner product structures of the first and second order
\begin{align*}
  \soboinprod{f}{g} &\mydefn \statinprod{f}{g} + \Exs_\stationary \big[ \nabla f (X)^\top \nabla g (X) \big], \quad \mbox{and} \\
  \inprod{f}{g}_{\mathbb{H}^2 (\stationary)} &\mydefn \soboinprod{f}{g} + \Exs_\stationary \big[ \mathrm{Tr} \big(\nabla^2 f (X)^\top \nabla^2 g (X) \big) \big],
\end{align*}
and we define the corresponding norms $\sobonorm{f} \mydefn \soboinprod{f}{f}^{1/2}$ and $\sobohilnorm{f}{2} \mydefn \inprod{f}{f}_{\mathbb{H}^2 (\stationary)}^{1/2}$.

For non-Hilbertian Sobolev spaces, for an integer $k \geq 0$ and a scalar $p \geq 1$, we define
\begin{align*}
  \sobokpstatnorm{f}{k}{p} \mydefn \Big\{ \sum_{i = 0}^k \int \matsnorm{\nabla^i f (x)}{F}^p d \stationary (x) \Big\}^{1/p},
\end{align*}
where we slightly abuse the notation to use the Frobenius norm $\matsnorm{\cdot}{F}$ to denote the Euclidean norm for vectors and high-order Frobenius norm for tensors.

\subsection{Additional related works}
It is helpful to discuss some existing literature on statistical analysis of reinforcement learning algorithms, in both the continuous time and the discrete-time settings.

\paragraph{Continuous-time RL:} Due to its versatile application in many real-world problems, RL problems in continuous-time settings have drawn a lot of recent research attention. Asymptotic convergence properties of various RL methods have been established in a recent line of literature~\cite{jia2023q,jia2022policyeval,jia2022policygrad,wang2020reinforcement}, under fairly general settings. In particular, \cite{jia2022policyeval} investigated the population-level convergence and numerical errors for a class of martingale-based Monte Carlo and temporal difference methods. The paper~\cite{kobeissi2023temporal} analyzed the TD stochastic approximation algorithm in a continuous-time setting, under an idealized $\mathrm{i.i.d.}$ observation model. Our results differ from~\cite{kobeissi2023temporal} in two aspects: first, the single-trajectory observation model is fundamentally different from the simulator model, as the samples becomes increasingly inter-dependent with small discretization stepsize (and the notion of ``sample size'' in our setting is the continuous-time trajectory length); second, we provide a refined analysis in the constant pre-factor in the bound, providing new guidance on the tradeoff between approximation and statistical errors. A closely-related paper is our own prior work~\cite{mou2024bellman}. We focused on population-level analysis of approximation and discretization errors in~\cite{mou2024bellman}, while this paper is devoted to a statistical error analysis.

Beyond policy evaluation, a broad range of recent literature addressed various aspects of continuous-time RL. The papers~\cite{zhou2021actor, li2022neural, ruthotto2020machine} investigated machine learning methods for PDEs arising from stochastic control problems. Focusing on fine-tuning of diffusion models, some recent research~\cite{tang2024fine,uehara2024fine,zhao2024scores,han2024stochastic} studied the convergence of reinforcement learning algorithms for a particular class of controlled SDEs. The techniques developed in this paper are potentially applicable to these settings.

\paragraph{Statistical estimation of value functions:} For discrete-time RL problems, theoretical analysis of policy evaluation algorithms is well-studied in literature. \cite{tsitsiklis1997analysis} established approximation error guarantees for TD with function approximation. Non-asymptotic statistical error analysis has been performed under various settings~\cite{srikant2019finite,mou2024optimal,duan2022policy}, for both the batch solution and stochastic approximation algorithms. Many of these theoretical guarantees are instance-dependent, namely, they provide fine-grained analysis adaptive to structures in the problem. This paper shares the same rationale, as we discover new structures in the statistical error of continuous-time problems. On the other hand, there is a key difference between continuous-time and discrete-time RL: a fundamental complexity that governs the optimal rates for discrete-time problems is the effective horizon; in continuous-time problems, the effective horizon diverges as stepsize goes to zero. Instead, we use structures in the diffusion process itself, such as ellipticity, to ensure convergence of the algorithms.

\section{Preliminaries}\label{sec:preliminaries}
Let us first set up the problem concretely. Assuming that the process~\eqref{eq:cts-time-process} has a unique stationary distribution $\stationary$, the learner collects discrete-time observations from a stationary trajectory $(\State_t: t \geq 0)$ with $\State_0 \sim \stationary$. For $k = 0,1,2\cdots, T / \stepsize$, we observe
\begin{align}
  (\State_{k \stepsize}, \Reward_{k \stepsize}) \quad \mbox{where } \Exs \big[ \Reward_{k \stepsize} \mid \State_{k \stepsize} \big] = \reward (\State_{k \stepsize}).
\end{align}
For simplicity of presentation, we assume that the random reward is uniformly bounded, satisfying $\abss{\Reward_t} \leq 1$ almost surely, for any $t$.

\subsection{Time discretization and function approximations}\label{subsec:discretized-projected-fixed-pt}
In the paper~\cite{mou2024bellman}, a higher-order discretization scheme is proposed for the Bellman equation~\eqref{eq:defn-bellman-eq}. Given an integer numerical order $\numerOrder \geq 2$, the time-discretized Bellman operator is defined as
\begin{subequations}
\begin{align}
  \label{eq:defn-high-order-bellman-operator}
  \BellOp^{(\numerOrder)} (\valuefunc) (\state) &\mydefn \sum_{i = 0}^{\numerOrder - 1} \int_0^{(\numerOrder - 1) \stepsize} e^{- \discount s} W_i (s) \Exs \big[ \reward (\MyState_{i \stepsize}) \mid \MyState_0 = \state \big] ds + e^{- \discount (\numerOrder - 1) \stepsize} \Exs \big[ \ValueFunc (\MyState_{(\numerOrder - 1) \stepsize}) \mid \MyState_0 = \state  \big],\\
  \mbox{where}\quad & W_i (s) \mydefn \Big\{ \prod_{j \neq i} \big( \stepsize j - \stepsize i \big) \Big\}^{-1}  \prod_{j \neq i} \big( s - \stepsize i\big), \quad \mbox{for } i = 1,2, \cdots \numerOrder.\label{eq:defn-chebyshev-poly}
\end{align}
\end{subequations}
The paper~\cite{mou2024bellman} further proposed projecting the Bellman operator on a finite-dimensional linear subspace $\LinSpace$, and solving the projected fixed-point equation.
\begin{align}
  \fbar = \projecttolin \circ \BellOp^{(\numerOrder)} (\fbar),\label{eq:defn-projected-discretized-fixed-pt}
\end{align}
where $\projecttolin$ is the orthonormal projection operator onto the linear subspace $\LinSpace$ under $\ltwospace (\stationary)$. By representing the function $\valuebar$ using finite-dimensional basis $\valuebar \mydefn \thetabar^\top \psi$. The projected fixed-point equation~\eqref{eq:defn-projected-discretized-fixed-pt} can be re-written as
\begin{multline}
  \Exs \big[ \psi (\MyState_0) \psi (\MyState_0)^\top \big] \thetabar \\
  = \sum_{i = 0}^{\numerOrder - 1} \int_0^{(\numerOrder - 1) \stepsize} e^{- \discount s} W_i (s) \Exs \big[ \reward (\MyState_{i \stepsize}) \psi (\MyState_0) \big] ds + e^{- \discount (\numerOrder - 1) \stepsize} \Exs \big[ \psi (\MyState_0) \psi (\MyState_{(\numerOrder - 1) \stepsize})^\top \big] \thetabar,\label{eq:projected-fixed-pt-in-subspace-rep}
\end{multline}
which is an $\mbasis$-dimensional linear equation, allowing for efficient computational algorithms.

In general, the projected fixed-point $\fbar$ may not be the best approximation to the true value function $\ValTrue$. As the effective horizon $1 / (1 - e^{- \stepsize (\numerOrder - 1) \discount})$ goes to infinity, the worst-case approximation factors~\cite{tsitsiklis1997analysis,mou2023optimal} can be large. Fortunately, with the help of elliptic structure in the diffusion process~\eqref{eq:cts-time-process}, one can still obtain bounded approximation factors. In doing so, we need to impose the following assumptions on the coefficients $(\drift, \covMat)$ and the reward function $\reward$.

\myassumption{Lip$(\numerOrder)$}{assume:smooth-high-order}{
    There exists positive constants $\big\{\smoothness_i^{\drift}\big\}_{i = 0}^{2\numerOrder - 2}$, $\big\{\smoothness_i^{\covMat}\big\}_{i = 0}^{2\numerOrder - 2}$, and $\big\{\smoothness_i^{\reward}\big\}_{i = 0}^{2 \numerOrder}$, such that
    \begin{subequations}
    \begin{align}
      \abss{\partial^\alpha \drift_k (x)} \leq \smoothness_i^{\drift},& \quad \mbox{for any $k \in [d]$ and multi-index $\alpha ~ \mathrm{s.t.}~|\alpha| \leq i$},\\
      \abss{\partial^\alpha \covMat_{k, \ell} (x)} \leq \smoothness_i^{\covMat}, &\quad \mbox{for any $k, \ell \in [d]$ and multi-index $\alpha ~ \mathrm{s.t.}~|\alpha| \leq i$},\\
      \abss{\partial^\alpha \reward (x)} \leq \smoothness_i^{\reward}, &\quad \mbox{for any multi-index $\alpha ~ \mathrm{s.t.}~|\alpha| \leq i$}.
    \end{align}
    \end{subequations}
}{1.5cm}

\myassumption{SL$(L_\stationary)$}{assume:smooth-stationary}
{
 For any $x \in \StateSpace$, we have $\vecnorm{\nabla \log \stationary (x)}{2} \leq \smoothness_\stationary$.
}{1.5cm}

\myassumption{UE$(\lammin, \lammax)$}{assume:uniform-elliptic}
{
  There exists positive constants $\lammin, \lammax$, such that
  \begin{align*}
    \lammin I_d \preceq \covMat (x) \preceq \lammax I_d, \quad \mbox{for any }x \in \real^d.
  \end{align*}
}{3cm}
These assumptions are widely used in the literature on continuous-time diffusion processes. Additionally, the paper~\cite{mou2024bellman} also imposes the following regularity assumption on the function class $\LinSpace$.
\myassumption{Reg$(c_1, \omega)$}{assume:basis-condition}
{
 For any function $\ValFun \in \LinSpace$ we have
 \begin{align*}
    \statnorm{\nabla \ValFunc} \leq c \mbasis^\omega \statnorm{f}, \quad \mbox{and} \quad \statnorm{\nabla^2 \ValFunc} \leq c \mbasis^{\omega} \statnorm{\nabla f}.
  \end{align*} 
}{2.2cm}
See the paper~\cite{mou2024bellman} for a detailed discussion on examples that satisfy these assumptions.
Under the assumptions, the paper~\cite{mou2024bellman} shows that when the stepsize $\stepsize$ satisfies $\stepsize \leq c_0 \mbasis^{- 4 \omega}$, we have the approximation factor upper bound
\begin{align}
  \sobonorm{\ValTrue - \fbar} \leq c_1 \inf_{\valfun \in \LinSpace} \sobonorm{\ValTrue - \valfun} + c_2 \stepsize^{\numerOrder},\label{eq:mou-zhu-bound}
\end{align}
for positive constants $(c_0, c_1, c_2)$ that depends on the the parameters in the four assumptions.

\subsection{Motivating example: learning advantage functions}
The approximation factor bound~\eqref{eq:mou-zhu-bound}, as well as \Cref{thm:main} to follow, provides guarantees on the estimation of the value function $\ValTrue$, measured in a first-order Sobolev norm. While the gradient of value function is interesting for its own sake, in most RL applications, the advantage function is of more practical interest. In particular, when the policy evaluation method serves as the critic component in policy gradient methods, the policy gradient oracle is given by the advantage function~\cite{konda1999actor,kakade2001natural}. Fortunately, the advantage function can be computed from the value function, and the Sobolev error in the value function is precisely the right quantity to control the estimation error of the advantage function.

To set up the problem, let us consider an action space $\ActionSpace$ and the controlled diffusion process of the form
\begin{align}
  d \MyState_t = \ctrDrift (\MyState_t, \Action_t) dt + \covMat (\MyState_t)^{1/2} d \BM_t.\label{eq:controlled-cts-time-process}
\end{align}

 A policy $\policy$ is a mapping from $\StateSpace$ to a probability distributions over $\ActionSpace$. When taking the policy $\policy$, the dynamics becomes
\begin{align}
  d \MyState_t^\policy = \drift^\policy (\MyState_t^\policy) dt +\covMat (\MyState_t^\policy)^{1/2} d \BM_t, \quad \mbox{where $\drift^\policy (\state) \mydefn \Exs_{\Action \sim \policy (\state)} [\ctrDrift (\state, \Action)]$}\label{eq:policy-cts-time-process}
\end{align}
We use $\ValueFunc^\policy$ to denote the value function under the dynamics~\eqref{eq:policy-cts-time-process}.

An advantage function is the difference between the value under a given action in the first step and the value under the policy $\policy$. In continuous-time setting, we apply the fixed action within an infinitesimal time interval. Similar to \cite{jia2023q}, we define the advantage function as the rate of change. In particular, we define
\begin{align}
  \advFunc (\state, \action) \mydefn \lim_{\Delta t \rightarrow 0}\frac{1}{\Delta t} \Exs_\state \Big[ \int_0^{\Delta t} e^{- \discount t} \reward (\MyState_t^\action) dt + e^{- \discount \Delta t} \ValFun^\policy (\MyState^\action_{\Delta t}) - \ValFun^\policy (\state)\Big].
\end{align}
The following result expresses the advantage function using gradient of the value function.
\begin{proposition}\label{prop:advantage-func}
  Let $\ValueFunc^\policy$ be the value function under policy $\policy$. We have
  \begin{align*}
    \advFunc (\state, \action)  = \inprod{\nabla \ValFun^\policy (\state)}{\ctrDrift (\state, \action) - \drift^\policy (\state)}, \quad \mbox{for any $\action \in \ActionSpace$},
  \end{align*}
  for any $(\state, \action)$ pair.
\end{proposition}
\noindent See \Cref{subsec:app-proof-of-prop-advantage-func} for the proof of this proposition.

A popular class of controlled diffusion processes is the control-affine systems, where the drift term takes the form $\ctrDrift (\state, \action) = \drift (\state) + \action$, for some function $\drift$. Such a model finds applications in both classical control systems~\cite{isidori1985nonlinear} and fine-tuning of diffusion models~\cite{uehara2024fine}. In the control-affine case, the advantage function can be expressed as
\begin{align}
  \advFunc (\state, \action) = \inprod{\nabla \ValFun^\policy (\state)}{\action - \int \action' d \policy (\action' | \state) }, \quad \mbox{for any $\action \in \ActionSpace$}.
\end{align}
When the policy $\policy$ is known, the advantage function can be computed from a value function estimator $\valuehat$ using a simple plug-in approach.
\begin{align*}
  \advFuncHat (\state, \action) = \inprod{\nabla \valuehat (\state)}{\action - \int \action' d \policy (\action' | \state)}, \quad \mbox{for any $\action \in \ActionSpace$},
\end{align*}
which leads to the bound
\begin{align*}
  \Exs_{\State \sim \stationary} \big[ \abss{ \advFuncHat (\State, \action) -   \advFunc (\State, \action)}^2 \big] \leq \mathrm{diam} (\actionspace)^2 \sobonorm{\ValHat - \valuefunc^\policy}^2,
\end{align*}
establishing the fundamental role of the Sobolev norm in the estimation of the advantage function.

\section{Main results}\label{sec:main-results}

Following the idea of least-square temporal difference methods~\cite{boyan2002technical}, we solve the projected fixed-point equation~\eqref{eq:projected-fixed-pt-in-subspace-rep} using empirical observations. In particular,
\begin{subequations}
given a trajectory $(\MyState_{t})_{0 \leq t \leq \totaltime}$ observed at discrete time steps $(k\stepsize: k \geq 0)$, we solve the linear equation
   \begin{align}
     \thetahat_\totaltime = \Big\{ \sum_{k = 0}^{\totaltime / \stepsize - \numerOrder} \psi (\MyState_{k \stepsize})  \cdot \Big( \psi (\MyState_{k \stepsize}) - e^{- \discount (\numerOrder - 1) \stepsize}  \psi (\MyState_{ (k + \numerOrder - 1) \stepsize}) \Big)^\top \Big\}^{-1} \cdot \Big\{ \stepsize \sum_{k = 0}^{\totaltime / \stepsize - \numerOrder}  \sum_{i = 0}^{\numerOrder - 1} \kappa_i \Reward_{(k + i) \stepsize} \psi (\MyState_{k \stepsize}) \Big\}, \label{eq:empirical-lstd}
   \end{align}
where the coefficients $\kappa_i$ are defined as
   \begin{align}
     \kappa_i \mydefn \frac{1}{\stepsize} \int_0^{(\numerOrder - 1) \stepsize} e^{- \discount s} W_i (s) ds, \quad \mbox{for $W_i$ defined in \Cref{eq:defn-chebyshev-poly}}\label{eq:defn-kappa-i-coeff}
   \end{align}
   Finally, the estimated value function is given by
   \begin{align}
     \ValHat (\state) \mydefn \inprod{\thetahat_\totaltime}{\psi (\state)}, \quad\mbox{for any $\state \in \real^\usedim$}.
   \end{align}
\end{subequations}
The rest of this paper is devoted to the statistical analysis of the estimator $\ValHat$.

\subsection{Assumptions}
Let us first discuss the assumptions. Since we perform a joint analysis for the approximation, statistical, and numerical errors, the assumptions discussed in \Cref{subsec:discretized-projected-fixed-pt} will still be needed. Additionally, we need some assumptions to facilitate the statistical analysis.

First, since we work with a single trajectory of a stochastic process. In order to have any meaningful guarantees, we need the underlying process to be ergodic. A popular assumption is the Poincar\'{e} inequality.
\myassumption{PI$(\poincare)$}{assume:markov-mixing}
{
  The stationary distribution $\stationary$ satisfies a Poincar\'{e} inequality
  \begin{align*}
   \var_\stationary \big( f (\State) \big) \leq \frac{1}{\poincare} \Exs_\stationary \big[ \vecnorm{\nabla f (\State)}{2}^2 \big], \quad \mbox{for any $f \in \mathbb{H}^1 (\stationary)$}.
  \end{align*}
}{1.5cm}
Poincar\'{e} inequality is standard in Markov diffusion literature, which can be guaranteed when the function $- \log \stationary$ satisfies certain growth conditions at infinity. See e.g.,~\cite{bakry2006diffusions,bakry2008simple} for examples of such conditions. Note that though \ref{assume:markov-mixing} is made only on the stationary distribution, together with the ellipticity condition~\ref{assume:uniform-elliptic}, the Poincar\'{e} inequality is sufficient for exponential convergence of the diffusion process in $\chi^2$ divergence.

Additionally, we need a hyper-contractivity condition within the linear subspace $\LinSpace$.
\myassumption{Hyper$(q, \tau_q)$}{assume:hyper-contractivity}
{
 For any function $\ValFun \in \LinSpace$ we have
 \begin{align*}
    \lpstatnorm{\ValueFunc}{q} \leq \tau_q \statnorm{\ValFun}, \quad  \sobopstatnorm{\ValueFunc}{q} \leq \tau_q \sobonorm{\ValFun}, \quad  \mbox{and} \quad  \sobokpstatnorm{\ValueFunc}{2}{q} \leq \tau_q \sobohilnorm{\ValFun}{2}.
  \end{align*} 
}{2.5cm}
Hyper-contractivity, or the moment comparison conditions in the form of \ref{assume:hyper-contractivity}, are widely used in statistical learning literature, which are satisfied, for example, by the Gaussian distribution. A popular version is the $\ell^4/\ell^2$-hypercontractivity, where $q = 4$. In our analysis, we do not optimize the exponent $q$ and require the condition to hold for some $q > 4$. With more careful proof arguments, it is possible to make the proof valid for $q$ arbitrarily close to $4$, and the hyper-contractivity assumption can be relaxed to a small subset of $\LinSpace$ \footnote{Indeed, we use the hyper-contractivity condition for all directions in $\LinSpace$ only in the matrix concentration arguments. For the leading order statistical error in \Cref{thm:main} to follow, the constant $\tau$ can be replaced by the hypercontractivity constant for the basis functions $(\coordinate_j^\top H^{-1/2} \psi)_{j = 1}^\mbasis$, which can be easily verified for most bases.} We refer the readers to the papers~\cite{mendelson2015learning,lecue2013learning,jain2018accelerating} for more discussions about this assumption and its applications to statistical learning.

Finally, we require the feature mapping $x \mapsto \psi (x)$ to be uniformly bounded under a particular norm, satisfying the following assumption
\myassumption{Bou$(D_\mbasis)$}{assume:bounded-feature}
{
  Defining the matrix $H_1 \mydefn \Exs_\stationary \big[ \psi (\State) \psi (\State)^\top \big] + \Exs_\stationary \big[ \nabla \psi (\State) \nabla \psi (\State)^\top \big]$, we have
  \begin{align*}
   \sup_{x, y \in \real^\usedim} \vecnorm{H_1^{-1/2} \psi (x)}{2} \cdot \Big\{ \vecnorm{H_1^{-1/2} \psi (y)}{2} + \matsnorm{H_1^{-1/2} \nabla \psi (y)}{F} +   \vecnorm{H_1^{-1/2} \generator \psi (y)}{2} \Big\} \leq D_\mbasis^2.
  \end{align*}
  
}{2cm}
We need boundedness of feature vectors to apply Bernstein-type inequalities in our arguments. While \ref{assume:bounded-feature} simplifies our arguments, it is possible to relax it to become some tail assumptions using more involved concentration arguments (see e.g.~\cite{adamczak2008tail}). We take the the form pre-conditioned by $H_1$ to align with the Sobolev norm used in the paper. When the feature vectors are well-behaved, we could expect $D_\mbasis \lesssim \sqrt{\mbasis}$, while it may grow faster with $\mbasis$ in certain cases.

\subsection{Main statistical guarantees}
Now we are ready to state the main results of the paper. First, let us define the following threshold, which characterizes the trajectory length needed for the theorem to hold true.
\begin{align}
  \Tthres (\mbasis, T_0) \mydefn \tau^4 \mbasis \log (\mbasis / \delta) \cdot \Big\{ \constScaryReg (T_0) \log^2 ( 1/ \stepsize) + \frac{\constScary}{\poincare} e^{- \lammin \poincare T_0 / 4} \mbasis^{2 \omega} \Big\}. \label{eq:defn-threshold-T-thres}
\end{align}
Based on these assumptions, we can derive a non-asymptotic upper bound for the error achieved by the LSTD estimator. Playing a key role in this result is the projection error, defined as
\begin{align*}
  \Delstar \mydefn \ValTrue - \projecttolinhil (\ValTrue), \quad \mbox{where} \quad \projecttolinhil (\ValTrue) = \arg\min_{\valfun \in \LinSpace} \sobonorm{\ValTrue - \valfun}.  
\end{align*}
We also define a pair of $\mbasis \times \mbasis$ matrices, which play a central role in controlling the complexities in the linear subspace.
\begin{align*}
  H_0 \mydefn \Exs_\stationary \big[ \psi (\State_0) \psi (\State_0)^\top \big] \quad \mbox{and} \quad  H_1 \mydefn \Exs_\stationary \big[ \psi (\State_0) \psi (\State_0)^\top \big] +  \Exs_\stationary \big[ \nabla \psi (\State_0) \nabla \psi (\State_0)^\top \big].
\end{align*}
Clearly, we have $H_0 \preceq H_1$, and in many cases, the eigen-values in $H_1$ grows much faster than $H_0$, making the quantity $\mathrm{Tr} (H_1^{-1} H_0)$ growing sublinearly in $\mbasis$.
\begin{theorem}\label{thm:main}
  Under Assumptions~\fakerefassumelip{$(\numerOrder + 1)$},~\ref{assume:smooth-stationary},~\ref{assume:uniform-elliptic}, for any $p > 1$ and $T_0 \geq 3$, there exists a constant $\constScary > 0$ depending on $p$ and the parameters in these assumptions, and a constant $\constScaryReg (T_0) > 0$ depending on $(p, T_0)$ and the underlying diffusion process~\eqref{eq:cts-time-process}, such that with Assumptions~\fakerefassumehyper{$q$}{$\tau$}, for $q = \max( 2p, \tfrac{4p}{4p - 1}, 8)$, and Assumptions~\ref{assume:basis-condition},~\ref{assume:markov-mixing} and~\ref{assume:bounded-feature}, for $\stepsize \leq c \mbasis^{- 4 \omega}$, there is an event $\Event$ such that $\Prob (\Event) \geq 1 - \delta$,  with the bound
  \begin{align*}
    \Exs \Big[ \sobonorm{\valuehat_T - \ValTrue}^2 \bm{1}_\Event \Big] &\leq c_1 \sobonorm{\Delstar}^2 + \frac{\tau^2 \mbasis\log^2 \tfrac{1}{\stepsize}}{T} \constScaryReg (T_0) \sobopstatnorm{\Delstar}{2p}^2  +  \frac{\tau^4 \constScary}{T} \mathrm{Tr} \big( H_1^{-1} H_0 \big) (\stepsize + \statnorm{\nabla \ValTrue}^2)\\
    &\qquad+ \constScary \stepsize^{2 \numerOrder} + \frac{\tau^2 \mbasis}{T} \constScary \Big\{ \poincare^{-1} e^{- \frac{\poincare \lammin}{4} T_0 } \sobokpstatnorm{\Delstar}{2}{2p} \sobopstatnorm{\Delstar}{2p} + \stepsize \sobokpstatnorm{\Delstar}{2}{2p}^2 \Big\},
  \end{align*}
  as long as the trajectory length satisfies the lower bound
  \begin{align*}
    T \geq 2 \Tthres (\mbasis, T_0) + \frac{2 c D_\mbasis^2}{\poincare} \log^{3/2} \big( \frac{ \mbasis}{\delta \stepsize} \big).
  \end{align*}
\end{theorem}
\noindent See Section~\ref{subsec:proof-thm-main} for the proof of this theorem.

A few remarks are in order. First, the function $\constScaryReg (T_0)$ depends on some regularity estimates for log density of the diffusion semigroup. In \Cref{subsec:some-special-cases}, we discuss the implications of a well-controlled $\constScaryReg (T_0)$ and the worst-case scenario. The trajectory length requirement also depends on such regularity estimates. It is linear in $\mbasis$ when both $\constScaryReg (T_0)$ and $D_\mbasis$ are well-controlled, but can be more stringent when we do not have such good bounds. \Cref{thm:main} bounds the mean-squared error on a high-probability event. Using Markov inequality, this implies an error upper bound that holds true with arbitrary constant probability (say, with probability $3/4$). Using some concentration arguments, it is easy to convert it to a high-probability bound with poly-logarithmic dependence on the failure probability $\delta$. We omit the details for simplicity.

Let us now discuss various error terms in \Cref{thm:main}.
\begin{carlist}
  \item The first term $\sobonorm{\Delstar}^2 = \inf_{\valfun \in \LinSpace} \sobonorm{\valfun - \ValTrue}^2$ is the best approximation error to the true solution $\ValTrue$, and therefore unavoidable so long as we are working in the linear subspace $\LinSpace$.
  \item The Markovian component in the statistical error
  \begin{align*}
    \frac{\tau^4 \mbasis}{T}  \Big\{ \constScaryReg (T_0) \sobopstatnorm{\Delstar}{2p}^2 \log^2 (1 / \stepsize) + \frac{\sobokpstatnorm{\Delstar}{2}{2p} \sobopstatnorm{\Delstar}{2p}}{ \poincare e^{\frac{\poincare \lammin}{4} T_0 }} + \stepsize \sobokpstatnorm{\Delstar}{2}{2p}^2 \Big\}.
  \end{align*}
  In addition to the $\mbasis / T$ rate of convergence, the pre-factor in these terms decay together with the approximation error $\sobonorm{\Delstar}^2$. When $p$ is close to unity, the leading-order term $\sobopstatnorm{\Delstar}{2p}^2$ is close to the approximation error $\sobonorm{\Delstar}^2$ itself. The two additional terms depend on approximation to the Hessian $\nabla^2 \ValTrue$. They are either exponentially decaying with $T_0$ or scaling with the stepsize $\stepsize$. Note that $T_0$ is not an algorithmic parameter, but can be arbitrarily chosen to balance the trade-off between the two terms. In \Cref{subsec:some-special-cases}, we will discuss  choices of $T_0$ under both a nice-behaved special case and the worst-case scenarios.
  \item The martingale component of the statistical error
  \begin{align*}
    \frac{\tau^4 \constScary}{T} \mathrm{Tr} \big( H_1^{-1} H_0 \big) (\stepsize + \statnorm{\nabla \ValTrue}^2).
  \end{align*}
  Unlike classical projection estimators~\cite{tsybakov2008introduction}, this term grows at a rate $\mathrm{Tr} \big( H_1^{-1} H_0 \big)$, which can be potentially slower than $\mbasis / T$. As we will see in \Cref{subsec:application-concrete-models}, various bounds on this term could lead to non-classical tradeoffs in the choice of $\mbasis$. See \Cref{subsec:application-concrete-models} for a concrete example of such trade-offs.
  \item The $\numerOrder$-th order numerical approximation error $\constScary \stepsize^{2 \numerOrder}$, which is standard in the literature and unavoidable for any numerical method.
\end{carlist}

\subsubsection{Some special cases}\label{subsec:some-special-cases}
It is helpful to discuss some special cases of the main result in \Cref{thm:main}.
\paragraph{Well-specified model:} When the model is well-specified, i.e., $\Delstar = 0$, the error bound simplifies to
\begin{align*}
  \Exs \Big[ \sobonorm{\valuehat_T - \ValTrue}^2 \bm{1}_\Event \Big] &\leq  \frac{\tau^4 \constScary}{T} \mathrm{Tr} \big( H_1^{-1} H_0 \big) (\stepsize + \statnorm{\nabla \ValTrue}^2) + \constScary \stepsize^{2 \numerOrder},
\end{align*}
so the Markovian component of the statistical error vanishes together with the approximation error. That being said, the trajectory length requirement $\Tthres (\mbasis, T_0)$ is still needed, and still depends on the mixing time of the underlying process.
\paragraph{Assumptions extended to $\mathrm{span} (\LinSpace, \ValTrue)$:} When Assumptions~\ref{assume:basis-condition} and \fakerefassumehyper{$\tfrac{4p}{p - 1}$}{$\tau$} holds true in the $(\mbasis + 1)$-dimensional linear subspace
\begin{align*}
  \mathrm{span} (\LinSpace, \ValTrue) \mydefn \Big\{ \valfun + \alpha \ValTrue: \valfun \in \LinSpace, \alpha \in \real \Big\},
\end{align*}
we have $\sobopstatnorm{\Delstar}{2p} \leq \tau \sobonorm{\Delstar}$ and $\sobokpstatnorm{\Delstar}{2}{2p} \leq \tau \mbasis^\omega \sobonorm{\Delstar}$. Consequently, the Markovian component of the statistical error is dominated by the approximation error $\sobonorm{\Delstar}^2$, as long as the trajectory length exceeds the threshold $\Tthres$. In this case, the error bound simplifies to
\begin{align*}
  \Exs \Big[ \sobonorm{\valuehat_T - \ValTrue}^2 \bm{1}_\Event \Big] &\leq \Big\{ c_1 + \frac{\Tthres}{T} \Big\} \inf_{\valfun \in \LinSpace} \sobonorm{\ValTrue - \valfun}^2 +  \frac{\tau^4 \constScary}{T} \mathrm{Tr} \big( H_1^{-1} H_0 \big) (\stepsize + \statnorm{\nabla \ValTrue}^2) + \constScary \stepsize^{2 \numerOrder}.
\end{align*}
The pre-factor $c_1 + \Tthres / T$ is upper bounded by $c_1 + 1$. So we only need to balance the approximation error and the martingale component of the statistical error.

\paragraph{Well-controlled $\constScaryReg (T_0)$:} Following the proof of \Cref{lemma:sigma-mkv-global-bound}, we have 
\begin{align*}
 \constScary (T_0) \leq c \big( T_0 + \constreg (T_0)\sqrt{T_0} \big),
\end{align*}
where $\constreg (T_0)$ is the constant pre-factor in the $\frac{4p}{p -1}$-th moment regularity estimates for the diffusion semi-group $(\semigroup_t)_{t \geq 0}$. See \Cref{eqs:relate-function-norm-to-basis} in the proof of \Cref{lemma:generator-cross-cov-bound} for details of such regularity estimates. Intuitively, under Assumption~\ref{assume:markov-mixing}, such regularity estimates should not depend on $T_0$, as the regularity of the density should be close to the regularity of the stationary distribution. In the paper~\cite{mou2022improved} (Proposition 2), a uniform-in-time upper bound is derived for the second moment, and the estimate was extended to higher moments in the paper~\cite{li2024error}, without explicit dependence on $T_0$. It remains an open question to derive an upper bound on $\constreg (T_0)$ with explicit dependence on $T_0$.

\begin{subequations}
Suppose that the process is well-behaved such that $\constreg (T_0) \lesssim \sqrt{T_0}$, then we have $\constScaryReg (T_0) \lesssim T_0$. In such a case, by taking $T_0 \asymp \poincare^{-1} \log (\mbasis / \stepsize)$, the bound in \Cref{thm:main} becomes
\begin{multline}
  \Exs \Big[ \sobonorm{\valuehat_T - \ValTrue}^2 \bm{1}_\Event \Big] \leq c_1 \sobonorm{\Delstar}^2 + \constScary \frac{\tau^2 \mbasis\log^3 \tfrac{\mbasis}{\stepsize}}{\poincare T} \sobopstatnorm{\Delstar}{2p}^2  +  \frac{\tau^4 \constScary}{T} \mathrm{Tr} \big( H_1^{-1} H_0 \big)  \statnorm{\nabla \ValTrue}^2\\
  \qquad+  \constScary \stepsize^{2 \numerOrder} + \frac{\constScary \stepsize}{T} \Big\{ \tau^2 \mbasis \sobokpstatnorm{\Delstar}{2}{2p}^2 + \tau^4 \mathrm{Tr} (H_1^{-1} H_0) \Big\},\label{eq:simplified-bound-good-case}
\end{multline}
under the trajectory length requirement
\begin{align}
  T \geq \frac{\constScary  \tau^4}{\poincare} \mbasis \log^4 \big( \frac{\mbasis}{\delta \stepsize} \big) + \frac{c D_\mbasis^2}{\poincare} \log^{3/2} \big( \frac{\mbasis}{\delta \stepsize} \big).\label{eq:sample-size-req-good-case}
\end{align}
When the stepsize $\stepsize$ is sufficiently small, the last two terms in the bound~\eqref{eq:simplified-bound-good-case} are negligible, and the error is dominated by the approximation error $\sobonorm{\Delstar}^2$ and the martingale component of the statistical error. Assuming furthermore that $D_\mbasis \lesssim \sqrt{\mbasis}$, the guarantee only requires a trajectory length of $O \big(\mbasis / \poincare \big)$, up to logarithmic factors.
\end{subequations}

\begin{subequations}
\paragraph{Worst-case $\constScaryReg (T_0)$:} In the worst-case scenario, we do not have a good control on $\constScaryReg (T_0)$ in terms of $T_0$. So we can simply take $T_0 = 3$, and the bound in \Cref{thm:main} becomes
\begin{multline}
  \Exs \Big[ \sobonorm{\valuehat_T - \ValTrue}^2 \bm{1}_\Event \Big] \leq c_1 \sobonorm{\Delstar}^2 +  \frac{\tau^4 \constScary}{T} \mathrm{Tr} \big( H_1^{-1} H_0 \big) (\stepsize + \statnorm{\nabla \ValTrue}^2) + \constScary \stepsize^{2 \numerOrder}\\
  \qquad + \frac{\tau^2 \mbasis \log^2 (1 / \stepsize)}{T} \constScary \Big\{ \poincare^{-1}  \sobokpstatnorm{\Delstar}{2}{2p} \sobopstatnorm{\Delstar}{2p} + \stepsize \sobokpstatnorm{\Delstar}{2}{2p}^2 \Big\},\label{eq:simplified-bound-bad-case}
\end{multline}
under the sample size requirement
\begin{align}
  T \geq \frac{\constScary \tau^4}{\poincare} \mbasis^{1 + 2 \omega} \log^3 \big( \frac{\mbasis}{\delta \stepsize} \big) + \frac{c D_\mbasis^2}{\poincare} \log^{3/2} \big( \frac{\mbasis}{\delta \stepsize} \big).\label{eq:sample-size-req-bad-case}
\end{align}
Compared to \Cref{eq:simplified-bound-good-case}, the Markovian part of the error bound in \Cref{eq:simplified-bound-bad-case} involves the product of first- and second-order Sobolev norm of the approximation error $\Delstar$, and the sample size requirement in \Cref{eq:sample-size-req-bad-case} scales with $\mbasis^{1 + 2 \omega}$, which is worse than the $O (\mbasis)$-scaling in \Cref{eq:sample-size-req-good-case}.

In general, it is an important direction of future research to determine the exact dependence of $\constScaryReg (T_0)$ on $T_0$ based on different assumptions on the underlying diffusion process.
\end{subequations}

\subsection{An illustrative example}\label{subsec:application-concrete-models}
To illustrate the non-standard trade-offs between approximation error and statistical error in \Cref{thm:main}, let us consider the following example. Suppose, for simplicity, that the process~\eqref{eq:cts-time-process} runs in a $\usedim$-dimensional torus $\torus^\usedim$,\footnote{Though the results are formally developed for $\real^\usedim$, all the arguments extends directly to torii.} and suppose that the stationary distribution $\stationary$ is uniform over $\torus^\usedim$. We let the feature mapping $\psi: \torus^\usedim \to \real^\mbasis$ be the (complex) Fourier basis. For notation simplicity, we index the basis functions by multi-indices $\alpha = (\alpha_1, \ldots, \alpha_\usedim) \in \integers^\usedim$. For any $n \geq 1$, the basis functions are given by
\begin{align*}
  \psi_\alpha (\state) \mydefn \exp \Big\{ 2 \pi i \inprod{\alpha}{\state} \Big\}, \quad \mbox{for any $\state \in \torus^\usedim$ and $\vecnorm{\alpha}{1} \leq n$}.
\end{align*}
 We have the following result
\begin{corollary}\label{cor:fourier-example}
  Under the assumptions of \Cref{thm:main}, for a scalar $k \geq 2$, if the value function $\ValTrue$ is order-$k$ H\"{o}lder, when $T \geq  \frac{c'}{\poincare} \mbasis^{1 + 2 / \usedim} \log^{3/2} \big( \frac{\mbasis}{\delta \stepsize} \big)$ we have
  \begin{align*}
    \Exs \Big[ \sobonorm{\valuehat_T - \ValTrue}^2 \bm{1}_\Event \Big] \leq c \Big\{ \mbasis^{- (2 k - 2) / d} + \frac{1}{T} g_\usedim (\mbasis) + \stepsize^{2 \numerOrder} \Big\} ,
  \end{align*}
  where we define
  \begin{align*}
    g_\usedim (\mbasis) \mydefn \begin{cases}
      1 & \usedim = 1,\\
      \log \mbasis & \usedim = 2,\\
      \mbasis^{1 - \frac{ 2}{\usedim}} & \usedim \geq 3.
      \end{cases},
  \end{align*}
   and the constants $(c, c')$ depends on the smoothness, hyper-contractivity and ellipticity parameters in the assumptions.
\end{corollary}
\noindent See \Cref{subsec:proof-cor-fourier-example} for the proof of this corollary. Based on \Cref{cor:fourier-example}, the optimal rates of convergence exhibits a more complicated picture.
\begin{itemize}
  \item[($\usedim = 1$)] the rate is $\mbasis^{- 2(k- 1)} + 1/T + \stepsize^{2 \numerOrder}$ while requiring $T \geq \mbasis^3 \log^{3/2} \mbasis$. So we have
  \begin{align*}
    \Exs \Big[ \sobonorm{\valuehat_T - \ValTrue}^2 \bm{1}_\Event \Big] \lesssim \max \Big( T^{-1}, \big(\tfrac{\log^{3/2} T}{T} \big)^{\frac{2}{3} (k - 1)} \Big) + \stepsize^{2 \numerOrder}.
  \end{align*}
  \item[$(\usedim = 2)$] the rate is $\mbasis^{- (k - 1)} + \log \mbasis / T + \stepsize^{2 \numerOrder}$ while requiring $T \geq \mbasis^2 \log^{3/2} \mbasis$. So we have
  \begin{align*}
    \Exs \Big[ \sobonorm{\valuehat_T - \ValTrue}^2 \bm{1}_\Event \Big] \lesssim \max \Big( \tfrac{\log T}{T}, \big( \tfrac{\log^{3/2} T}{T} \big)^{- \frac{1}{2} (k - 1)} \Big) + \stepsize^{2 \numerOrder}.
  \end{align*}
  \item[$(\usedim \geq 3)$] the rate is $\mbasis^{- 2(k - 1) / \usedim} + \mbasis^{1 - 2/\usedim} / T + \stepsize^{2 \numerOrder}$ while requiring $T \geq \mbasis^{1 + 2 / \usedim} \log^{3/2} \mbasis$. So we have
  \begin{align*}
    \Exs \Big[ \sobonorm{\valuehat_T - \ValTrue}^2 \bm{1}_\Event \Big] \leq T^{\frac{- 2 (k - 1)}{2k + \usedim - 4}} + \stepsize^{2 \numerOrder}.
  \end{align*}
\end{itemize}
Given a sufficiently small stepsize $\stepsize$, these rates are (up to log factors) faster than the standard $O (\numobs^{- \frac{2 (k - 1)}{2k + d}})$ minimax rates for estimating the gradient of a H\"{o}lder function with $\numobs$ observations (c.f.~\cite{tsybakov2008introduction}). This is because the main part of the statistical error shrinks with the approximation error, leading to a non-standard tradeoff. It is also worth noting that the improved rates in \Cref{cor:fourier-example} does not require any growth condition on the function $\constScaryReg$. This is because the trajectory length requirement $T \gtrsim D_\mbasis^2$ ensures that both the leading-order and high-order Markovian statistical errors are dominated by the approximation error.

\subsection{Technical highlight: structures in the Markov chain covariance}

Let us discuss the intuition behind the statistical error bound in \Cref{thm:main}, highlighting the key technical discoveries.

Recall that the estimator $\ValHat$ solves a linear equation using Markov chain trajectory data. 
The paper~\cite{mou2024optimal} characterize the optimal statistical complexities for such problems. In particular, as $T \rightarrow + \infty$ while keeping all other parameters fixed, the asymptotic distribution of $\thetahat_T$ takes the form
\begin{align*}
  \sqrt{T} (\thetahat_T - \thetabar) \xrightarrow{d} \mathcal{N} \big(0, \stepsize \Abar^{-1} (\SigMG^* + \SigMkv^*) \Abar^{- \top} \big),
\end{align*}
where $\Abar = \Exs \big[ \psi (\MyState_{k \stepsize})  \cdot \big( \psi (\MyState_{k \stepsize}) - e^{- \discount (\numerOrder - 1) \stepsize}  \psi (\MyState_{ (k + \numerOrder - 1) \stepsize}) \big]$, and the matrices $\SigMG^*$, $\SigMkv^*$ correspond to the martingale and Markovian parts of the noise variance, respectively. When applied to the continuous-time policy evaluation problem, the two covariance terms take the following form
\begin{subequations}
\begin{carlist}
  \item $\SigMG^*$ is from the covariance of a continuous-time martingale in the form
  \begin{align}
   \psi (\State_0) \int_0^\stepsize e^{- \discount t} \nabla \fbar (\State_t) \covMat (\State_t)^{1/2} dB_t.
  \end{align}
  This contributes to the $\mathrm{Tr} (H_1^{-1} H_0)$ term in \Cref{thm:main}, and is independent of the mixing time of the Markovian diffusion. Note that we do not have gradient on $\psi$ for the covariance, while the pre-conditioner $\Abar$ scales with $H_1$. This makes it possible for the martingale part of the statistical error to grow at a sub-linear rate.
  \item $\SigMkv$ involves the asymptotic covariance matrix in the Markov chain CLT satisfied by sum of the path functional
  \begin{align}
    \psi (\State_{k \stepsize}) \int_0^\stepsize e^{- \discount t} (\discount - \generator) (\ValTrue - \fbar) (\State_{k \stepsize + t}) dt.\label{eq:mkv-part-noise-in-proof-sketch}
  \end{align}
  This quantity involves second derivative of the error function $\ValTrue - \fbar$. In order to control it with a leading-order term depending only on the first-order Sobolev norm (as in \Cref{thm:main}), we seek to ``move'' the gradient operator from the function $\ValTrue - \fbar$ to the basis function $\psi$. This key technical ingredient in our result is a refined analysis on the structures in the Markov chain covariance.
\end{carlist}
\end{subequations}

Concretely, given a pair of functions $f, g$ in the domain of $\generator$, we define the cross-covariance operator between the functionals in the form of~\eqref{eq:mkv-part-noise-in-proof-sketch}
\begin{align}
  \mu_k (f, g) \mydefn \cov \Big( \frac{f (\State_0) }{\stepsize} \int_0^\stepsize e^{- \discount t} (\discount - \generator) g (\State_t) dt,  \frac{f (\State_{k \stepsize}) }{\stepsize}  \int_0^\stepsize e^{- \discount t} (\discount - \generator) g (\State_{k \stepsize + t}) dt \Big).\label{eq:defn-muk-cross-var}
\end{align}
The Markov chain CLT admits the (rescaled) asymptotic covariance structure
\begin{align*}
  \sigstar_{\Mkv} (f, g)^2 = \stepsize \mu_0 (f, g) + 2 \stepsize \sum_{k = 0}^{+ \infty} \mu_k (f, g).
\end{align*}
We have the following lemma
\begin{lemma}\label{lemma:sigma-mkv-global-bound}
  Under the setup of \Cref{thm:main}, for any function $f \in \LinSpace$, we have
  \begin{align*}
    \sigstar_{\Mkv} (f, g)^2 &\leq \stepsize \abss{\mu_0 (f, g)} + 2 \stepsize \sum_{k = 0}^{+ \infty} \abss{\mu_k (f, g)}\\
    &\leq \tau^2 \sobonorm{f}^2 \Big\{ \constScaryReg (T_0) \sobopstatnorm{g}{2p}^2 \log (1 / \stepsize) + \constScary \highorder_{T_0, \stepsize} (g) \Big\},
  \end{align*}
  where the high-order term is given by
  \begin{align*}
    \highorder_{T_0, \stepsize} (g) \mydefn \poincare^{-1} e^{- \frac{\poincare \lammin}{4} T_0 } \sobokpstatnorm{g}{2}{2p} \cdot \sobopstatnorm{g}{2p} + \stepsize \sobokpstatnorm{g}{2}{2p}^2.
  \end{align*}
\end{lemma}
\noindent See \Cref{subsec:proof-lemma-sigma-mkv-global-bound} for the proof of this lemma. Intuitively, this lemma allows us to ``re-distribute'' the gradient operator in \Cref{eq:defn-muk-cross-var}, moving from the function $g$ to $f$. The leading-order term in such a bound depends only on first-order Sobolev norms of both functions. By letting $g = \ValTrue - \fbar$ and relating the error $ \ValTrue - \fbar$ to $\Delstar$ using the elliptic structures, we show that the Markovian component of the noise covariance is controlled by the approximation error itself.

\section{Proofs}\label{sec:proofs}
In this section, we collect the proofs of the sample-based results presented in~\Cref{sec:main-results}.

\subsection{Proof of \Cref{lemma:sigma-mkv-global-bound}}
\label{subsec:proof-lemma-sigma-mkv-global-bound}

We first prove the following lemma.
\begin{lemma}\label{lemma:generator-cross-cov-bound}
  Under Assumptions~\fakerefassumelip{$(2)$},~\ref{assume:smooth-stationary} and~\ref{assume:uniform-elliptic}, for any scalar $p > 1$ and time $T_0 > 0$, there exists a constant $\constreg (p, T_0) > 0$ that depends only on $p, T_0$ and the diffusion process~\eqref{eq:cts-time-process}, such that for functions $f_1, f_2, g_1, g_2 \in D (\generator)$ and any $t \in (0, T_0]$, we have
  \begin{multline*}
   \abss{ \Exs \big[ f_1 (\MyState_0) (\discount - \generator) g_1 (\MyState_0) \cdot f_2 (\MyState_t) (\discount - \generator) g_2 (\MyState_t) \big] } \\
   \leq \constreg (p, T_0) \Big\{1 + \frac{1}{t} \Big\} \sobopstatnorm{g_1 }{2p} \cdot \sobopstatnorm{g_2 }{2p} \cdot \sobopstatnorm{f_1}{\frac{4p}{p - 1}} \cdot \sobopstatnorm{f_2}{\frac{4p}{p - 1}}.
  \end{multline*}
\end{lemma}
\noindent See \Cref{subsubsec:proof-lemma-generator-cross-cov-bound} for the proof of this lemma.

Based on \Cref{lemma:generator-cross-cov-bound}, we can bound the term $\mu_k$ in the asymptotic variance $\sigstar_{\Mkv}$, as given in the following lemma.
\begin{lemma}\label{lemma:muk-approx-error-bound}
   Under Assumptions~\fakerefassumelip{$(2)$},~\ref{assume:smooth-stationary} and~\ref{assume:uniform-elliptic}, for any scalar $p > 1$ and time $T_0 > 0$, there exists a constant $\constreg (p, T_0) > 0$ that depends only on $p, T_0$ and the diffusion process~\eqref{eq:cts-time-process}, and a constant $c' > 0$ depending on the parameters in Assumptions~\fakerefassumelip{$(2)$},~\ref{assume:smooth-stationary} and~\ref{assume:uniform-elliptic}, such that for pair of functions $f, g \in D (\generator)$ and any $k \stepsize \in (0, T_0]$, we have
  \begin{align*}
    |\mu_k| \leq \constreg(p, T_0) \Big\{1 + \frac{1}{k \stepsize} \Big\} \sobopstatnorm{f}{\frac{4p}{p - 1}}^2 \sobopstatnorm{g}{2p}^2 +  c' \stepsize^2  \sobokpstatnorm{f}{2}{\frac{2p}{p - 1}}^2  \sobokpstatnorm{g}{2}{2p}^2
  \end{align*}
\end{lemma}
\noindent See \Cref{subsubsec:proof-lemma-muk-approx-error-bound} for the proof of this lemma.

Note that Lemma~\ref{lemma:generator-cross-cov-bound} and~\ref{lemma:muk-approx-error-bound} are applicable only for $t = k \stepsize \in (0, T_0]$. When the time index is outside this range, we use alternative estimates given by the following lemma.
\begin{lemma}\label{lemma:muk-mixing-bound}
  Under the setup of Lemma~\ref{lemma:sigma-mkv-global-bound}, assuming additionally that \ref{assume:markov-mixing} holds true, there exists a constant $c_0 > 0$ depending on parameters in Assumptions~\fakerefassumelip{$(2)$},~\ref{assume:smooth-stationary} and~\ref{assume:uniform-elliptic}, such that for any $k \geq 2 / \stepsize + 1$, we have
  \begin{align*}
    |\mu_0| &\leq c_0 \lpstatnorm{f}{\frac{2p}{p - 1}}^2 \cdot\sobokpstatnorm{g}{2}{2p}^{2}, \quad \mbox{and}\\
    |\mu_k| &\leq c_0 \exp \Big( - \frac{\lammin \poincare}{4} k \stepsize \Big) \sobopstatnorm{g}{2p} \cdot \sobokpstatnorm{g}{2}{2p} \cdot \sobopstatnorm{f}{\frac{4p}{p - 1}}^2
  \end{align*}
\end{lemma}
\noindent See \Cref{subsubsec:proof-lemma-muk-mixing-bound} for the proof of this lemma.
Combining \Cref{lemma:muk-approx-error-bound} and \Cref{lemma:muk-mixing-bound}, for $T_0 \geq 3$, we have
\begin{align}
  &\stepsize \sum_{k = 0}^{+ \infty} |\mu_k| = \stepsize |\mu_0| + \stepsize\sum_{1 \leq k \leq T_0 / \stepsize} |\mu_k| + \stepsize \sum_{k > T_0 / \stepsize} |\mu_k| \nonumber \\
  &\leq c_0 \stepsize \lpstatnorm{f}{\frac{2p}{p - 1}}^2 \cdot\sobokpstatnorm{g}{2}{2p}^{2} + \constreg(p, T_0)  \sobopstatnorm{f}{\frac{4p}{p - 1}}^2 \sobopstatnorm{g}{2p}^2 \sum_{1 \leq k \leq T_0 / \stepsize} \stepsize \Big\{1 + \frac{1}{k \stepsize} \Big\} \nonumber \\
  &\qquad \qquad + \stepsize^2 T_0 \sobokpstatnorm{f}{2}{\frac{2p}{p - 1}}^2  \sobokpstatnorm{g}{2}{2p}^2 +  c_0  \sobopstatnorm{g}{2p} \sobokpstatnorm{g}{2}{2p}  \sobopstatnorm{f}{\frac{4p}{p - 1}}^2 \sum_{k \geq T_0 / \stepsize} \stepsize e^{ - \frac{\lammin \poincare}{4} k \stepsize } \nonumber \\
  &\leq \sobopstatnorm{f}{\frac{4p}{p - 1}}^2 \Big\{ \constreg(p, T_0) \sobopstatnorm{g}{2p}^2 T_0 \log (1 / \stepsize) +  \frac{4 c_0 e^{- \lammin \poincare T_0 / 4}}{\lammin \poincare} \sobopstatnorm{g}{2p} \cdot \sobokpstatnorm{g}{2}{2p} \Big\} \nonumber \\
  &\qquad \qquad + \Big\{c_0 \stepsize \lpstatnorm{f}{\frac{2p}{p - 1}}^2 + c' \stepsize^2 T_0 \sobokpstatnorm{f}{2}{\frac{2p}{p - 1}}^2 \Big\} \sobokpstatnorm{g}{2}{2p}^2. \label{eq:sum-of-muk-collect-bounds}
\end{align}
By Assumption~\fakerefassumehyper{$\frac{4p}{p - 1}$}{$\tau_{\frac{4p}{p - 1}}$}, we have
\begin{align*}
  \sobopstatnorm{f}{\frac{4p}{p - 1}} \leq \tau_{\frac{4p}{p - 1}} \sobonorm{f}, \quad \lpstatnorm{f}{\frac{2p}{p - 1}} \leq \tau_{\frac{4p}{p - 1}} \statnorm{f}, \quad \mbox{and} \quad  \sobokpstatnorm{f}{2}{\frac{2p}{p - 1}} \leq \tau_{\frac{4p}{p - 1}} \sobohilnorm{f}{2}.
\end{align*}
Furthermore, by the basis function assumption~\ref{assume:basis-condition}, we have
\begin{align*}
  \sobohilnorm{f}{2} \leq 2 c_1 \mbasis^\omega \sobonorm{f}.
\end{align*}
So given the stepsize satisfying $\stepsize \mbasis^\omega \leq 1$, we have
\begin{align*}
  c_0 \stepsize \lpstatnorm{f}{\frac{2p}{p - 1}}^2 + c' \stepsize^2 T_0 \sobokpstatnorm{f}{2}{\frac{2p}{p - 1}}^2 \leq \tau_{\frac{4p}{p - 1}}^2 \stepsize (c_0 + 2 c' c_1) T_0 \sobonorm{f}^2.
\end{align*}
Substituting them to Eq~\eqref{eq:sum-of-muk-collect-bounds}, we conclude that
\begin{align*}
  &\stepsize \sum_{k = 0}^{+ \infty} |\mu_k|\\
  & \leq c \tau_{\frac{4p}{p - 1}}^2 \statnorm{f}^2 \Big\{ T_0 \constreg(p, T_0) \sobopstatnorm{g}{2p}^2 \log(1 / \stepsize) + \frac{e^{- \lammin \poincare T_0 / 4}}{\poincare} \sobopstatnorm{g}{2p} \cdot \sobokpstatnorm{g}{2}{2p} + \stepsize T_0 \sobokpstatnorm{g}{2}{2p}^2 \Big\},
\end{align*}
which completes the proof of \Cref{lemma:sigma-mkv-global-bound}.

\subsubsection{Proof of \Cref{lemma:generator-cross-cov-bound}}\label{subsubsec:proof-lemma-generator-cross-cov-bound}
Given $t > 0$ and $x \in \real^\usedim$, let $p_t ( \cdot | x)$ be the density of $\MyState_t$ for the process~\eqref{eq:cts-time-process} starting from $\MyState_0 = x$, we have
\begin{align*}
  &\Exs \big[ f_1 (\MyState_0) (\discount - \generator) g_1 (\MyState_0) \cdot f_2 (\MyState_t) (\discount - \generator) g_2 (\MyState_t) \big] \\
  &= \iint  f_1 (x) (\discount - \generator) g_1 (x) f_2 (y) (\discount - \generator) g_2 (y) p_t (y | x) \stationary (x) dy  dx.
\end{align*}
For notational convenience, throughout this proof, we use $\nabla_y p_t (\cdot | \cdot)$ to denote the gradient with respect to the first argument, and use $\nabla_x p_t (\cdot | \cdot)$ to denote the gradient with respect to the second argument.

Our proof relies on applying integration-by-parts formula several times to move the gradient operator from the functions $g_1$ and $g_2$ to rest parts of the integral. For notation convenience, for $i \in \{1, 2\}$, we define the following auxiliary functions
\begin{align*}
  h_i (x) \mydefn  g_i (x) \drift (x) + \frac{1}{2}  \covMat (x) \nabla g_i (x), \quad \mbox{and} \quad  \ell_i (x) \mydefn \discount g_i (x) + g_i (x)  \nabla \cdot \drift (x) +  \frac{1}{2} \inprod{\nabla \covMat (x)}{\nabla g_i (x)}.
\end{align*}
Clearly, by the smoothness assumption~\fakerefassumelip{2}, we have
\begin{align*}
   \lpstatnorm{h_i}{p} \leq \big( \smoothness_0^\drift + \smoothness_0^\covMat \big) \usedim \sobopstatnorm{g_i}{p}  \quad \mbox{and} \quad \lpstatnorm{\ell_i}{p} \leq \big(\discount + \smoothness_1^\drift + \smoothness_1^\covMat \big) \usedim \sobopstatnorm{g_i}{p},
\end{align*}
for any scalar $p \geq 1$ and $i \in \{1,2\}$.

To prove \Cref{lemma:generator-cross-cov-bound}, we use the following decomposition result
\begin{lemma}\label{lemma:integration-by-parts-decomp-in-cross-cov-bound}
   Given functions $f_1, g_1, f_2, g_2 \in D (\generator)$, we have
   \begin{align*}
     \Exs \big[ f_1 (\MyState_0) (\discount - \generator) g_1 (\MyState_0) \cdot f_2 (\MyState_t) (\discount - \generator) g_2 (\MyState_t) \big] = E_1 + E_2,
   \end{align*}
   where we define the components
   \begin{align*}
     E_1 & \mydefn \Exs \Bigg[ \Big\{h_1 (\State_0)^\top \big(\nabla f_1 + f_1  \nabla \log \stationary\big) (\State_0) + \ell_1 (\State_0)  f_1 (\State_0)  \Big\}\\
     &\qquad \qquad \cdot \Big\{ h_2 (\State_t)^\top \big( \nabla f_2 (\State_t) + f_2 (\State_t) \nabla_y \log p_t (\State_t \mid \State_0) \big) +  \ell_2 (\State_t) f_2 (\State_t) \Big\} \Bigg],
   \end{align*}
   and
   \begin{align*}
     E_2 & \mydefn \Exs \Bigg[ f_1 (\State_0) h_1^\top (\State_0)\Big\{\nabla_x \log p_t (\State_t| \State_0) \nabla f_2 (\State_t)^\top + f_2 (\State_t) \frac{\nabla_x \nabla_y^\top p_t (\State_t | \State_0)}{p_t (\State_t | \State_0)}\Big\} h_2 (\State_t) \Bigg]\\
     &\qquad \qquad + \Exs \Big[ f_1 (\State_0) f_2 (\State_t) \ell_2 (\State_t) h_1 (\State_0)^\top \nabla \log_x p_t (\State_t | \State_0)  \Big].
   \end{align*}
\end{lemma}
\noindent We prove this lemma at the end of this section. Taking this lemma as given, let us now proceed with the proof of Lemma~\ref{lemma:generator-cross-cov-bound}.

By Cauchy--Schwarz inequality, we have
\begin{multline*}
  E_1^2 \leq \Exs \Big[\Big\{h_1 (\State_0)^\top \big(\nabla f_1 + f_1  \nabla \log \stationary\big) (\State_0) + \ell_1 (\State_0)  f_1 (\State_0)  \Big\}^2 \Big]\\
     \cdot \Exs \Big[ \Big\{ h_2 (\State_t)^\top \big( \nabla f_2 (\State_t) + f_2 (\State_t) \nabla_y \log p_t (\State_t \mid \State_0) \big) +  \ell_2 (\State_t) f_2 (\State_t) \Big\}^2 \Big].
\end{multline*}
By H\"{o}lder inequality, we have
\begin{align*}
  &\Exs \Big[\Big\{h_1 (\State_0)^\top \big(\nabla f_1 + f_1  \nabla \log \stationary\big) (\State_0) + \ell_1 (\State_0)  f_1 (\State_0)  \Big\}^2 \Big]\\
  &\leq 2 \Exs \Big[\abss{h_1 (\State_0)^\top \big(\nabla f_1 + f_1 \nabla \log \stationary\big) (\State_0)}^2 \Big] + 2 \Exs \Big[ \abss{\ell_1 (\State_0)  f_1 (\State_0) }^2 \Big] \\
  &\leq 2 (1 + \smoothness_\stationary) \big( \lpstatnorm{h_1}{2p}^2 + \lpstatnorm{\ell_1}{2p}^2 \big) \cdot \sobopstatnorm{f_1}{\frac{2p}{p - 1}}^2\\
  &\leq 2 (1 + \smoothness_\stationary) \big( \smoothness_0^\drift + \smoothness_0^\covMat + \smoothness_1^\drift + \smoothness_1^\covMat + \discount  \big)^2 d^2 \sobopstatnorm{g_1}{2p}^2 \sobopstatnorm{f_1}{\frac{2p}{p - 1}}^2.
\end{align*}
and applying H\"{o}lder inequality twice yields
\begin{align*}
  &\Exs \Big[ \Big\{ h_2 (\State_t)^\top \big( \nabla f_2 (\State_t) + f_2 (\State_t) \nabla_y \log p_t (\State_t \mid \State_0) \big) +  \ell_2 (\State_t) f_2 (\State_t) \Big\}^2 \Big]\\
  &\leq 2 \Exs \Big[ \abss{ h_2 (\State_t)^\top \big( \nabla f_2 (\State_t) + f_2 (\State_t) \nabla_y \log p_t (\State_t \mid \State_0) \big) }^2 \Big] + 2 \Exs \Big[ \abss{ \ell_2 (\State_t) f_2 (\State_t)}^2  \Big]\\
  &\leq 2 \lpstatnorm{h_2}{2p}^2 \cdot \Exs \Big[ \vecnorm{ \nabla f_2 (\State_t) + f_2 (\State_t) \nabla_y \log p_t (\State_t \mid \State_0)}{2}^{\frac{2p}{p - 1}} \Big]^{\frac{p - 1}{p}} + 2 \lpstatnorm{\ell_2}{2p}^2 \cdot \sobopstatnorm{f_2}{\frac{2p}{p - 1}}^2\\
  &\leq  2\Big\{ 1 + \Exs \Big[ \vecnorm{ \nabla_y \log p_t (\State_t \mid \State_0)}{2}^{\frac{4 p}{p - 1}} \Big]^{\frac{p - 1}{2 p}} \Big\} \lpstatnorm{h_2}{2p}^2 \sobopstatnorm{f_2}{\frac{4p}{p - 1}}^2  + 2 \lpstatnorm{\ell_2}{2p}^2 \cdot \sobopstatnorm{f_2}{\frac{2p}{p - 1}}^2\\
  &\leq  4 \Big\{ 1 + \Exs \Big[ \vecnorm{ \nabla_y \log p_t (\State_t \mid \State_0)}{2}^{\frac{4 p}{p - 1}} \Big]^{\frac{p - 1}{2 p}} \Big\} \big( \smoothness_0^\drift + \smoothness_0^\covMat + \smoothness_1^\drift + \smoothness_1^\covMat + \discount \big)^2 d^2 \sobopstatnorm{g_2}{2p}^2 \sobopstatnorm{f_2}{\frac{4p}{p - 1}}^2
\end{align*}
Putting them together leads to the bound
\begin{align*}
  |E_1| \leq c_1 \Big\{ 1 + \Exs \Big[ \vecnorm{ \nabla_y \log p_t (\State_t \mid \State_0)}{2}^{\frac{4 p}{p - 1}} \Big]^{\frac{p - 1}{4 p}} \Big\} \sobopstatnorm{g_1}{2p} \cdot \sobopstatnorm{g_2}{2p} \cdot \sobopstatnorm{f_1}{\frac{4p}{p - 1}} \cdot \sobopstatnorm{f_2}{\frac{4p}{p - 1}},
\end{align*}
for constant $c_1$ depending on the constants $(\discount, \usedim, \smoothness_0^\drift, \smoothness_0^\covMat, \smoothness_1^\drift, \smoothness_1^\covMat, \smoothness_\stationary)$.

For the term $E_2$, we apply H\"{o}lder's inequality to the three terms to obtain that
\begin{align*}
  &\abss{\Exs \Big[ f_1 (\State_0) h_1 (\State_0)^\top \nabla_x \log p_t (\State_t| \State_0) \nabla f_2(\State_t)^\top h_2 (\State_t) \Big]}\\
  &\leq \sqrt{\Exs \big[ \vecnorm{f_1 (\State_0) h_1 (\State_0)}{2}^2 \big]} \cdot \sqrt{\Exs \big[ \vecnorm{\nabla_x \log p_t (\State_t| \State_0) \nabla f_2 (\State_t)^\top h_2 (\State_t)}{2}^2\big]}\\
  &\leq \lpstatnorm{f_1}{\frac{2p}{p - 1}} \lpstatnorm{h_1}{2p} \cdot \Big( \Exs \big[ \matsnorm{\nabla_x \log p_t (\State_t| \State_0) \nabla f_2 (\State_t)^\top}{F}^{\frac{2p}{p - 1}} \big] \Big)^{\frac{p - 1}{2p}} \cdot \Big(\Exs [ \vecnorm{h_2 (\State_t)}{2}^{2p}]\Big)^{1/(2p)}\\
  &\leq \Exs \big[ \vecnorm{\nabla_x \log p_t (\State_t| \State_0)}{2}^{\frac{4p}{p - 1}} \big]^{\frac{p - 1}{4p}} \cdot \lpstatnorm{f_1}{\frac{4p}{p - 1}} \cdot  \lpstatnorm{f_2}{\frac{4p}{p - 1}} \cdot \lpstatnorm{h_1}{2p} \cdot \lpstatnorm{h_2}{2p},
\end{align*}
and similarly,
\begin{multline*}
  \abss{ \Exs \Big[ f_1 (\State_0) h_1^\top (\State_0) f_2 (\State_t) \frac{\nabla_x \nabla_y^\top p_t (\State_t | \State_0)}{p_t (\State_t | \State_0)}  h_2 (\State_t) \Big]}\\
  \leq \Exs \Big[ \matsnorm{\frac{\nabla_x \nabla_y^\top p_t (\State_t| \State_0)}{p_t (\State_t| \State_0)}}{F}^{\frac{4p}{p - 1}} \Big]^{\frac{p - 1}{4p}} \cdot \lpstatnorm{f_1}{\frac{4p}{p - 1}} \cdot \lpstatnorm{f_2}{\frac{4p}{p - 1}} \cdot \lpstatnorm{h_1}{2p} \cdot \lpstatnorm{h_2}{2p},
\end{multline*}
as well as the last term
\begin{align*}
  &\abss{ \Exs \Big[ f_1 (\State_0) f_2 (\State_t) \ell_2 (\State_t) h_1 (\State_0)^\top \nabla \log_x p_t (\State_t | \State_0)  \Big]}\\
  &\leq \sqrt{\Exs \vecnorm{f_1 (\State_0) h_1 (\State_0)}{2}^2} \cdot \sqrt{\Exs \vecnorm{f_2 (\State_t) \ell_2 (\State_t) \nabla \log_x p_t (\State_t | \State_0)}{2}^2}\\
  &\leq \Exs \big[ \vecnorm{\nabla_x \log p_t (\State_t| \State_0)}{2}^{\frac{4p}{p - 1}} \big]^{\frac{p - 1}{4p}} \cdot \lpstatnorm{f_1}{\frac{4p}{p - 1}} \cdot\lpstatnorm{f_2}{\frac{4p}{p - 1}} \cdot \lpstatnorm{h_1}{2p} \cdot \lpstatnorm{\ell_2}{2p}.
\end{align*}
Putting them together yields
\begin{multline*}
  |E_2| \leq c_2 \Big\{  \Exs \Big[ \matsnorm{ \frac{\nabla_x \nabla_y^\top p_t (\State_t \mid \State_0) }{ p_t (\State_t \mid \State_0)}}{F}^{\frac{4 p}{p - 1}} \Big]^{\frac{p - 1}{4 p}}  + \Exs \Big[ \vecnorm{ \nabla_x \log p_t (\State_t \mid \State_0)}{2}^{\frac{4 p}{p - 1}} \Big]^{\frac{p - 1}{4 p}} \Big\} \\
  \cdot \sobopstatnorm{g_1}{2p} \cdot \sobopstatnorm{g_2}{2p} \cdot \sobopstatnorm{f_1}{\frac{4p}{p - 1}} \cdot \sobopstatnorm{f_2}{\frac{4p}{p - 1}} .
\end{multline*}
Combining the bounds for $E_1$ and $E_2$, we conclude that
\begin{multline}
  \frac{\abss{\Exs \big[ f_1 (\MyState_0) \generator g_1 (\MyState_0) \cdot f_2 (\MyState_t) \generator g_2 (\MyState_t) \big]}}{\sobopstatnorm{g_1}{2p} \cdot \sobopstatnorm{g_2}{2p} \cdot \sobopstatnorm{f_1}{\frac{4p}{p - 1}} \cdot \sobopstatnorm{f_1}{\frac{4p}{p - 1}} }\\
  \leq c' \Big\{ 1 + \Exs \Big[ \vecnorm{ \frac{\nabla_y p_t}{p_t} (\State_t \mid \State_0)}{2}^{\frac{4 p}{p - 1}} \Big] + \Exs \Big[ \vecnorm{ \frac{\nabla_x p_t}{p_t} (\State_t \mid \State_0)}{2}^{\frac{4 p}{p - 1}} \Big] + \Exs \Big[ \matsnorm{ \frac{\nabla_x \nabla_y^\top p_t}{ p_t } (\State_t \mid \State_0)}{F}^{\frac{4 p}{p - 1}} \Big]  \Big\}^{\frac{p - 1}{4 p}} .\label{eq:generator-cross-cov-bound-in-fisher-info}
\end{multline}
It remains to bound the moments of gradients and mixed-derivatives of the log-density $\log p_t (y|x)$. In doing so, we use the following well-known result from the Malliavin calculus literature.
\begin{proposition}[\cite{menozzi2021density}, Theorem 2.1]\label{prop:malliavin}
  Given a pair of integers $j, j' \geq 0$, under Assumptions~\fakerefassumelip{($j+j'$)} and~\ref{assume:uniform-elliptic}, for any scalar $q > 1$ and time $T_0 > 0$, there exists a constant $c_0 (q, j + j')$ that depends only on $(q, j + j', \lammin, \lammax, \usedim, T_0)$ and $(\smoothness_i^\drift, \smoothness_i^{\covMat})_{i = 0}^{j + j'}$, such that for any $t \in (0, T_0]$ and $C^\infty$-smooth function $\phi$, we have
  \begin{align*}
    |\nabla^j \semigroup_t \big( \nabla^{j'} \phi \big) (x)| \leq \frac{c (q, T_0, j + j')}{t^{(j + j') / 2}} \Big\{ \semigroup_t \big( |\phi|^q \big) (x)\Big\}^{1/q}.
  \end{align*}
\end{proposition}
In order to apply this result, we note that
\begin{subequations}
\begin{align}
  \nabla \semigroup_t \phi (x) &= \nabla_x \int \phi (y) p_t (y|x) dy = \int \phi(y) \nabla_x p_t (y|x) dy = \Exs \big[ \phi(\State_t) \frac{\nabla_x p_t}{p_t} (\State_t | x) \mid \State_0 = x \big],\label{eq:semigroup-regularize-to-density-gradx} \\
  \semigroup_t \big( \nabla \phi \big) (x) &=\int \nabla_y \phi (y) p_t (y|x) dy = - \int \phi(y) \nabla_y p_t (y|x) dy = - \Exs \big[ f(\State_t) \frac{\nabla_y p_t}{p_t} (\State_t | x) \mid \State_0 = x \big], \label{eq:semigroup-regularize-to-density-grady} \\
  \nabla \semigroup_t \big( \nabla \phi \big) (x) &=\int \nabla_y \phi (y) \nabla_x p_t (y|x)^\top dy = - \int \phi(y) \nabla_y \nabla_x^\top p_t (y|x) dy \nonumber \\
  &= - \Exs \Big[ \phi(\State_t) \frac{\nabla_y \nabla_x^\top p_t}{p_t} (\State_t | x) \mid \State_0 = x \Big]. \label{eq:semigroup-regularize-to-density-gradxy}
\end{align}
\end{subequations}
\begin{subequations}\label{eqs:log-density-regularity}
Choosing $\phi (y) = \frac{\nabla_x p_t}{p_t} (y|x) \cdot \vecnorm{\frac{\nabla_x p_t}{p_t} (y|x)}{2}^{\frac{2 p + 2}{p - 1}}$ in Eq~\eqref{eq:semigroup-regularize-to-density-gradx}, and applying \Cref{prop:malliavin} with $q = \frac{4p}{3p + 1}$ yields
\begin{align*}
  \Exs_x \Big[  \vecnorm{\frac{\nabla_x p_t}{p_t} (\State_{t}|x)}{2}^{\frac{4p}{p - 1}} \Big] \leq \frac{1}{\sqrt{t}} c \big(\tfrac{4p}{3p + 1}, T_0, 1 \big) \cdot \Big\{
  \Exs_x \Big[  \vecnorm{\frac{\nabla_x p_t}{p_t} (\State_{t}|x)}{2}^{\frac{4p}{p - 1}} \Big] \Big\}^{\frac{3p + 1}{4p}},
\end{align*}
which leads to the bound
\begin{align}
  \Big\{ \Exs_x \Big[  \vecnorm{\frac{\nabla_x p_t}{p_t} (\State_{t}|x)}{2}^{\frac{4p}{p - 1}} \Big]\Big\}^{\frac{p - 1}{4p}} \leq \frac{1}{\sqrt{t}} c \big(\tfrac{4p}{3p + 1}, T_0, 1 \big).\label{eq:log-density-regularity-x}
\end{align}
Similarly, by substituting $\phi (y) = \frac{\nabla_y p_t}{p_t} (y|x) \cdot \vecnorm{\frac{\nabla_y p_t}{p_t} (y|x)}{2}^{\frac{2 p + 2}{p - 1}}$ in Eq~\eqref{eq:semigroup-regularize-to-density-gradx}, we obtain the bound
\begin{align}
  \Big\{ \Exs_x \Big[  \vecnorm{\frac{\nabla_y p_t}{p_t} (\State_{t}|x)}{2}^{\frac{4p}{p - 1}} \Big]\Big\}^{\frac{p - 1}{4p}} \leq \frac{1}{\sqrt{t}} c \big(\tfrac{4p}{3p + 1}, T_0, 1 \big). \label{eq:log-density-regularity-y}
\end{align}
Finally, applying Eq~\eqref{eq:semigroup-regularize-to-density-gradxy} with $\phi (y) = \frac{\nabla_x \nabla_y^\top p_t}{p_t} (y|x) \cdot \matsnorm{\frac{\nabla_x \nabla_y^\top p_t}{p_t} (y|x)}{F}^{\frac{2 p + 2}{p - 1}}$, we have
\begin{align*}
  \Exs_x \Big[  \matsnorm{\frac{\nabla_x \nabla_y^\top p_t}{p_t} (\State_{t}|x)}{F}^{\frac{4p}{p - 1}} \Big] \leq \frac{1}{t} c \big(\tfrac{4p}{3p + 1}, T_0, 2 \big) \cdot \Big\{
  \Exs_x \Big[  \matsnorm{\frac{\nabla_x \nabla_y^\top p_t}{p_t} (\State_{t}|x)}{F}^{\frac{4p}{p - 1}} \Big] \Big\}^{\frac{3p + 1}{4p}},
\end{align*}
which leads to the bound
\begin{align}
  \Big\{ \Exs_x \Big[  \matsnorm{\frac{\nabla_x \nabla_y^\top p_t}{p_t} (\State_{t}|x)}{F}^{\frac{4p}{p - 1}} \Big]\Big\}^{\frac{p - 1}{4p}} \leq \frac{1}{t} c \big(\tfrac{4p}{3p + 1}, T_0, 2 \big). \label{eq:log-density-regularity-xy}
\end{align}
\end{subequations}

Substituting the inequalities \Cref{eq:log-density-regularity-x,eq:log-density-regularity-y,eq:log-density-regularity-xy} to \Cref{eq:generator-cross-cov-bound-in-fisher-info}, we conclude that
\begin{multline*}
  \abss{\Exs \big[ f_1 (\MyState_0) \generator g_1 (\MyState_0) \cdot f_2 (\MyState_t) \generator g_2 (\MyState_t) \big]} \\
  \leq c' \Big\{ 1 + \frac{1}{\sqrt{t}} c \big(\tfrac{4p}{3p + 1}, T_0, 1 \big) + \frac{1}{t} c \big(\tfrac{4p}{3p + 1}, T_0, 2 \big) \Big\} \sobopstatnorm{g_1}{2p} \cdot \sobopstatnorm{g_2}{2p} \cdot \sobopstatnorm{f_1}{\frac{4p}{p - 1}} \cdot \sobopstatnorm{f_2}{\frac{4p}{p - 1}},
\end{multline*}
which completes the proof of \Cref{lemma:generator-cross-cov-bound}.

\paragraph{Proof of Lemma~\ref{lemma:integration-by-parts-decomp-in-cross-cov-bound}:}
Applying integration-by-parts formula with respect to the variable $x$, we obtain that
\begin{align}
  &\iint  f_1 (x) (\discount - \generator) g_1 (x) f_2 (y) (\discount - \generator) g_2 (y) p_t (y | x) \stationary (x) dy  dx \nonumber \\
  &= \iint f_1 (x) \Big\{ \discount g_1 (x) - \inprod{\drift (x)}{\nabla g_1 (x)} - \frac{1}{2} \mathrm{Tr}\big( \covMat (x) \nabla^2 g_1 (x) \big) \Big\} f_2 (y) (\discount - \generator) g_2 (y) p_t (y | x) \stationary (x) dy  dx\nonumber \\
  &= \iint \Big\{\discount f_1 (x) g_1 (x) p_t (y | x) \stationary (x) + \nabla_x^\top \big( f_1 (x) p_t (y | x) \stationary (x) \drift (x) \big)  g_1 (x) + \frac{1}{2} \inprod{ \nabla_x \big( f_1 (x) p_t (y | x) \stationary (x) \covMat (x) \big) }{ \nabla g_1 (x)} \Big\} \nonumber\\
  &\qquad \qquad \qquad \qquad \cdot f_2 (y) (\discount - \generator) g_2 (y)  dy  dx\nonumber \\
  &= \iint  \Bigg[ \Big\{ \nabla f_1 (x) + f_1 (x) \nabla_x \log p_t (y | x) + f_1 (x) \nabla \log \stationary (x) \Big\}^\top \Big\{ g_1 (x) \drift (x) + \frac{1}{2}  \covMat (x) \nabla g_1 (x) \Big\}\nonumber \\
  &\qquad \qquad \qquad \qquad + f_1 (x)  \Big\{ g_1 (x) \big(\discount + \nabla \cdot \drift (x) \big) + \frac{1}{2} \inprod{\nabla \covMat (x)}{\nabla g_1 (x)} \Big\} \Bigg] \cdot f_2 (y) (\discount - \generator) g_2 (y) p_t (y | x) \stationary (x) dy  dx\nonumber \\
  &= \int  \Big[ h_1 (x)^\top \Big\{ \nabla f_1 (x) + f_1 (x) \nabla \log \stationary (x) \Big\}  + f_1 (x) \ell_1 (x) \Big] \int f_2 (y) (\discount - \generator) g_2 (y) p_t (y | x)  dy  \stationary (x) dx \nonumber \\
  &\qquad + \iint  h_1 (x)^\top f_1 (x) \nabla_x \log p_t (y | x)  f_2 (y) (\discount - \generator) g_2 (y) p_t (y | x) \stationary (x) dy  dx.
   \label{eq:big-integration-by-parts-in-cross-cov-decomp-proof}
\end{align}
Now we study the inner integral (with respect to the variable $y$). For the two terms, we apply integration-by-parts to each of them and obtain
\begin{align}
  &\int f_2 (y) (\discount - \generator) g_2 (y) p_t (y | x) dy \nonumber \\
  &= \int f_2 (y) p_t (y|x) \Big\{ \discount g_2 - \inprod{\nabla g_2}{\drift} - \frac{1}{2} \mathrm{Tr} \big( \covMat \nabla^2 g_2 \big) (y) \Big\} dy \nonumber\\
  &= \int \Bigg[ \Big\{ \nabla f_2 (y) + f_2 (y) \nabla_y \log p_t (y|x) \Big\}^\top \Big\{ g_2 (y) \drift (y) + \frac{1}{2} \covMat (y) \nabla g_2 (y) \big) \Big\} \nonumber\\
  &\qquad \qquad \qquad + f_2 (y) g_2 (y) \big(\discount + \nabla \cdot \drift (y) \big) + \frac{1}{2} f_2 (y) \inprod{\nabla \covMat (y)}{\nabla g_2 (y)} \Bigg] p_t (y | x)dy \nonumber\\
  & = \int \Bigg[ h_2 (y)^\top \Big\{ \nabla f_2 (y) + f_2 (y) \nabla_y \log p_t (y|x) \Big\} + f_2 (y) \ell_2 (y) \Bigg] p_t (y | x)dy.\label{eq:y-integration-by-parts-in-cross-cov-decomp-proof}
\end{align}
and for each $j \in [\usedim]$, the same argument yields
\begin{align*}
  &\int f_2 (y) (\discount - \generator) g_2 (y) \partial_{x_j} p_t (y | x) dy \\
  &= \int \Bigg[ \Big\{ \nabla f_2 (y) \cdot \partial_{x_j} \log p_t (y| x) + f_2 (y) \partial_{x_j} \nabla_y \log p_t (y | x) + f_2 (y) \partial_{x_j} \log p_t (y| x)  \nabla_y \log p_t (y|x) \Big\}^\top h_2 (y)\\
  &\qquad \qquad \qquad  + \partial_{x_j} \log p_t (y| x) f_2 (y) \ell_2 (y) \Bigg] p_t (y | x)dy.
\end{align*}
By concatenating the coordinates together, we obtain
\begin{align}
  &\int f_2 (y) (\discount - \generator) g_2 (y) p_t (y | x) \nabla_x \log p_t (y | x) dy \nonumber \\
  &= - \int \Bigg[ \Big\{\nabla_x \log p_t (y| x) \nabla^\top f_2 (y) + f_2 (y) \nabla_x \nabla_y^\top \log p_t (y | x) + f_2 (y) \nabla_x \log p_t (y| x)  \nabla_y \log p_t (y|x)^\top \Big\} h_2 (y) \nonumber\\
  &\qquad \qquad \qquad +\nabla_x \log p_t (y| x) f_2 (y) \ell_2 (y) \Bigg] p_t (y | x)dy.\label{eq:ylogp-integration-by-parts-in-cross-cov-decomp-proof}
\end{align}
To complete the proof, we substitute \Cref{eq:y-integration-by-parts-in-cross-cov-decomp-proof} to the first term of \Cref{eq:big-integration-by-parts-in-cross-cov-decomp-proof}, and obtain
\begin{align*}
   &\int  \Big[ h_1 (x)^\top \Big\{ \nabla f_1 (x) + f_1 (x) \nabla \log \stationary (x) \Big\}  + f_1 (x) \ell_1 (x) \Big] \int f_2 (y) (\discount - \generator) g_2 (y) p_t (y | x)  dy  \stationary (x) dx\\
   &= \iint  \Big\{ h_1 (x)^\top \big( \nabla f_1 (x) + f_1 (x) \nabla \log \stationary (x) \big)  + f_1 (x) \ell_1 (x) \Big\} \\
   &\qquad \qquad \qquad \cdot \Big\{ h_2 (y)^\top \big( \nabla f_2 (y) + f_2 (y) \nabla \log p_t (y | x) \big)  + f_2 (y) \ell_2 (y) \Big\} \stationary (x) p_t (y|x) dy dx\\
   &=: E_1.
\end{align*}
Similarly, substituting \Cref{eq:ylogp-integration-by-parts-in-cross-cov-decomp-proof} to the second term of \Cref{eq:big-integration-by-parts-in-cross-cov-decomp-proof} yields
\begin{align*}
   &\iint  h_1 (x)^\top f_1 (x) \nabla_x \log p_t (y | x)  f_2 (y) (\discount - \generator) g_2 (y) p_t (y | x) \stationary (x) dy  dx  \\
   & =  \iint f_1 (x)  p_t (y | x) \stationary (x) h_1 (x)^\top \\
   &\qquad \qquad \qquad \cdot \Bigg[ \Big\{\nabla_x \log p_t (y| x) \nabla f_2 (y)^\top + f_2 (y) \frac{\nabla_x \nabla_y^\top p_t (y | x)}{p_t (y | x)} \Big\} h_2 (y) + \nabla_x \log p_t (y| x) f_2 (y) \ell_2 (y) \Bigg] dx dy\\
   &=: E_2.
\end{align*}
combining them completes the proof of \Cref{lemma:integration-by-parts-decomp-in-cross-cov-bound}.

\subsubsection{Proof of \Cref{lemma:muk-approx-error-bound}}\label{subsubsec:proof-lemma-muk-approx-error-bound}
By definition, we note that
\begin{align*}
  \mu_k &= \frac{1}{\stepsize^2} \int_0^\stepsize \int_0^\stepsize e^{- \discount (t_1 + t_2)} \Exs \big[ f (\State_0) (\discount - \generator) g (\State_{t_1}) \cdot f (\State_{k \stepsize}) \big( \semigroup_{t_2}(\discount - \generator) g\big) (\State_{k \stepsize}) \big] dt_1 dt_2\\
  & \qquad \qquad - \Big( \frac{1}{\stepsize} \int_0^\stepsize e^{- \discount t} \Exs \big[ f (\State_0) (\discount - \generator) g (\State_{t}) \big] dt \Big)^2\\
  &\leq \frac{1}{\stepsize^2} \int_0^\stepsize \int_0^\stepsize e^{- \discount (t_1 + t_2)} \Exs \big[ f (\State_0) (\discount - \generator) g (\State_{t_1}) \cdot f (\State_{k \stepsize}) \big( \semigroup_{t_2}(\discount - \generator) g\big) (\State_{k \stepsize}) \big] dt_1 dt_2.
\end{align*}
We seek to bound the integrand for any fixed pair $(t_1, t_2)$.

Since the generator is interchangeable with the diffusion semigroup, for any $t_2 > 0$, we have $\semigroup_{t_2}(\discount - \generator) g = (\discount - \generator) \semigroup_{t_2} g$. Define the auxiliary function
\begin{align*}
  L (x) \mydefn (\discount - \generator) g (x) \cdot \Exs \Big[ f (\State_{k \stepsize}) \cdot \big( (\discount - \generator) \semigroup_{t_2} g \big) (\State_{k \stepsize}) \mid \State_{t_1} = x \Big].
\end{align*}
By It\^{o}'s formula, we note that
\begin{align*}
  L (\State_{t_1}) = L (\State_0) + \int_0^{t_1} \generator L (\State_t) dt + \int_0^{t_1} \nabla L (\State_{t})^\top \covMat (\State_t)^{1/2} dB_t.
\end{align*}
Note that the last term is a martingale in $t_1$. By substituting it back to the expectation term of interest, we have
\begin{align*}
  \Exs \big[ f (\State_0) L (\State_{t_1}) \big] &= \Exs \big[ f (\State_{t_1}) L (\State_{t_1}) \big] + \int_0^{t_1} \Exs \big[f(\State_t) \generator L (\State_{t_1}) \big] dt\\
  &= \Exs \big[ f (\State_0) L (\State_0) \big] + \int_0^{t_1} \Exs \big[f(\State_t) \generator \semigroup_{t_1 - t} L (\State_{t}) \big] dt.
\end{align*}
Applying integration-by-parts formula, we note that
\begin{align*}
  &\Exs \big[f(\State_t) \generator \semigroup_{t_1 - t} L (\State_{t}) \big]\\
   &= \int f(x) \stationary (x) \Big\{ \inprod{\nabla \semigroup_{t_1 - t} L}{\drift} (x) + \frac{1}{2} \mathrm{Tr} \big( \covMat \nabla^2 \semigroup_{t_1 - t} L \big) (x) \Big\} dx\\
  &= \int \Big\{ - \nabla \cdot \big( f \stationary \drift \big) (x) + \frac{1}{2} \nabla^2 \cdot \big( f \stationary \covMat \big) (x) \Big\} \semigroup_{t_1 - t} L  (x) dx.
\end{align*}
According to the Fokker--Planck equation, the stationary distribution $\stationary$ satisfies $- \nabla \cdot (\stationary \drift) + \frac{1}{2} \nabla^2 \cdot (\stationary \covMat) = 0$, so we have
\begin{align*}
   - \nabla \cdot \big( f \stationary \drift \big) (x) + \frac{1}{2} \nabla^2 \cdot \big( f \stationary \covMat \big) (x) = \stationary (x) \Big\{ - \inprod{\nabla f}{\drift} + \frac{1}{2} \mathrm{Tr} \big(\covMat \nabla^2 f \big) + \inprod{\nabla f }{\covMat \nabla \log \stationary + \nabla \cdot \covMat} \Big\} (x).
\end{align*}
So by H\"{o}lder's inequality, we obtain that
\begin{align*}
  &\abss{\Exs \big[f(\State_t) \generator \semigroup_{t_1 - t} L (\State_{t}) \big]} \\
  &\leq \int \abss{ - \inprod{\nabla f}{\drift} + \frac{1}{2} \mathrm{Tr} \big(\covMat \nabla^2 f \big) + \inprod{\nabla f }{\covMat \nabla \log \stationary + \nabla \cdot \covMat} \Big\} (x)} \cdot \abss{\semigroup_{t_1 - t} L  (x)}  \stationary (x)  dx\\
  &\leq  \big( \smoothness_0^\drift + \smoothness_0^\covMat + \smoothness_\stationary + \smoothness_1^\covMat \big) \usedim \int \Big\{ \vecnorm{\nabla f (x)}{2} + \matsnorm{\nabla^2 f (x)}{F} \Big\} \cdot \abss{\semigroup_{t_1 - t} L  (x)}  \stationary (x)  dx\\
  &\leq \big( \smoothness_0^\drift + \smoothness_0^\covMat + \smoothness_\stationary + \smoothness_1^\covMat \big) \usedim  \cdot \sobokpstatnorm{f}{2}{\frac{2p}{p - 1}} \cdot \lpstatnorm{\semigroup_{t_1 - t} L}{\frac{2p}{p + 1}}.
\end{align*}
By Jensen's inequality, since the function $x \mapsto x^{\frac{2p}{p + 1}}$ is convex for positive $x$, we have
\begin{align*}
  \lpstatnorm{\semigroup_{t_1 - t} L}{\frac{2p}{p + 1}}^{\frac{2p}{p + 1}} = \Exs \Big[ \abss{\Exs \big[ L (\State_{t_1 - t}) \mid \State_0 \big]}^{\frac{2p}{p + 1}} \Big] \leq \Exs \Big[ \abss{L (\State_{t_1 - t})}^{\frac{2p}{p + 1}} \Big] = \lpstatnorm{L}{\frac{2p}{p + 1}}^{\frac{2p}{p + 1}}.
\end{align*}
Applying H\"{o}lder's inequality once more, we obtain that
\begin{align*}
  \lpstatnorm{L}{\frac{2p}{p + 1}} \leq  \lpstatnorm{(\discount - \generator) g}{2p} \cdot \statnorm{\semigroup_{k \stepsize - t_1} \Big(f \cdot (\discount - \generator) \semigroup_{t_2} g \Big)}
\end{align*}
We claim that
\begin{align}
  \statnorm{\semigroup_{k \stepsize - t_1} \Big(f \cdot (\discount - \generator) \semigroup_{t_2} g \Big)} \leq  \frac{\constreg (p, T_0)}{\sqrt{k \stepsize - t_1}} \sobokpstatnorm{g}{1}{2p} \cdot \sobokpstatnorm{f}{1}{\frac{4p}{p - 1}}.\label{eq:additional-regularity-in-muk-approx-lemma}
\end{align}
We prove this bound at the end of this section. Taking this bound as given, let us proceed with the proof of \Cref{lemma:muk-approx-error-bound}.

Substituting \Cref{eq:additional-regularity-in-muk-approx-lemma} to the bound above, we have
\begin{align*}
  &\abss{\Exs \big[f(\State_t) \generator \semigroup_{t_1 - t} L (\State_{t}) \big]}\\
  &\leq \frac{c' \constreg (p, T_0)}{\sqrt{k \stepsize - t_1}}  \sobokpstatnorm{f}{2}{\frac{2p}{p - 1}} \cdot \lpstatnorm{(\discount - \generator) g}{2p} \cdot \sobokpstatnorm{g}{1}{2p} \cdot \sobokpstatnorm{f}{1}{\frac{4p}{p - 1}}\\
  &\leq \frac{c'' \constreg (p, T_0)}{\sqrt{k \stepsize - t_1}}  \sobokpstatnorm{f}{2}{\frac{2p}{p - 1}} \cdot \sobokpstatnorm{g}{2}{2p} \cdot \sobokpstatnorm{g}{1}{2p} \cdot \sobokpstatnorm{f}{1}{\frac{4p}{p - 1}},
\end{align*}
where the constants $c'$ and $c''$ depend on the smoothness parameters in Assumption~\fakerefassumelip{(2)}.

Therefore, we have that
\begin{align}
  &\abss{\Exs \big[ f (\State_0) L (\State_{t_1}) \big]} \nonumber \\
  &\leq \abss{\Exs \big[f (\State_0) L (\State_0) \big]} + \int_0^{t_1} \abss{\Exs \big[f(\State_t) \generator \semigroup_{t_1 - t} L (\State_{t}) \big]} dt \nonumber\\
  &\leq \abss{\Exs \big[f (\State_0) L (\State_0) \big]} + \frac{c'' \stepsize}{\sqrt{k \stepsize - t_1}} \constreg (p, T_0) \sobokpstatnorm{f}{2}{\frac{2p}{p - 1}} \cdot \sobokpstatnorm{g}{2}{2p} \cdot \sobokpstatnorm{g}{1}{2p} \cdot \sobokpstatnorm{f}{1}{\frac{4p}{p - 1}}.\label{eq:additional-term-final-in-muk-approx-lemma}
\end{align}
By \Cref{lemma:generator-cross-cov-bound}, we have
\begin{align*}
  \abss{\Exs \big[f (\State_0) L (\State_0) \big]} &= \abss{\Exs \Big[ f (\State_0) \cdot (\discount - \generator) g (\State_0) \cdot f (\State_{k \stepsize}) \cdot (\discount - \generator) (\semigroup_{t_2} g) (\State_{k \stepsize}) \Big]}\\
  &\leq \constreg (p, T_0) \Big\{1 + \frac{1}{k \stepsize} \Big\} \sobopstatnorm{f}{\frac{4p}{p - 1}}^2 \cdot \sobopstatnorm{g}{2p} \cdot \sobopstatnorm{\semigroup_{t_2} g}{2p}.
\end{align*}
Finally, it remains to relate the Sobolev norm of $\semigroup_{k \stepsize - t_1}  g$ to the function $g$ itself. By note that by Jensen's inequality, we have $\Exs [|\semigroup_{t_2} g (\State_0)|^{2p}] \leq \Exs [ |g (\State_{t_2})|^{2p} ] = \lpstatnorm{g}{2p}^{2p}$. As for the gradient, we use an existing regularity result by~\cite{wang2005character}.

\begin{proposition}[\cite{wang2005character}, Proposition 1.2]\label{prop:semigroup-grad-estimate-local-grow}
  Under above setup, for any bounded Lipschitz function $u$ on $\real^\usedim$ and $t > 0$, we have
  \begin{align*}
    \vecnorm{\nabla \semigroup_t u (x)}{2}^2 \leq e^{2 c t} \semigroup_t \Big( \vecnorm{\nabla u }{2}^2 \Big) (x), \quad \mbox{for any $x \in \StateSpace$},
  \end{align*}
  for any constant $c$ satisfying
  \begin{align*}
    c \geq \frac{1}{\lammin} \sup_{x \in \real^\usedim} \Big\{ \sum_{i, j \in [\usedim]} \vecnorm{\nabla \covMat_{i,j} (x)}{2}^2 + \opnorm{\nabla \cdot \drift (x)}^2 \Big\}.
  \end{align*}
\end{proposition}
Applying \Cref{prop:semigroup-grad-estimate-local-grow} to the function $g$, for $t_2 \leq \stepsize \leq 1/c$, we have
\begin{align*}
  \Exs \big[ \vecnorm{\nabla \semigroup_{t_2} g (\State_0)}{2}^{2p} \big] \leq e^2 \Exs \Big[ \abss{\semigroup_{t_2} \big( \vecnorm{\nabla g}{2}^2\big) (\State_0) }^p \Big] \leq e^2 \Exs \big[ \vecnorm{\nabla g (\State_{t_2})}{2}^{2p} \big] = e^2 \lpstatnorm{\nabla g}{2p}^{2p}.
\end{align*}
We therefore have the bound
\begin{align}
  \sobopstatnorm{\semigroup_{t_2} g}{2p} \leq e \sobopstatnorm{g}{2p}, \label{eq:semigroup-sobo-norm-growth-bound}
\end{align}
and consequently,
\begin{align*}
  \abss{\Exs \big[f (\State_0) L (\State_0) \big]} \leq e \constreg (p, T_0) \Big\{1 + \frac{1}{k \stepsize} \Big\} \sobopstatnorm{f}{\frac{4p}{p - 1}}^2 \cdot \sobopstatnorm{g}{2p}^2.
\end{align*}
Substituting back to \Cref{eq:additional-term-final-in-muk-approx-lemma}, we have
\begin{align*}
  &\abss{\Exs \big[f (\State_0) (\discount - \generator) g (\State_{t_1}) \cdot f (\State_{k \stepsize}) \big( \semigroup_{t_2}(\discount - \generator) g\big) (\State_{k \stepsize}) \big]}\\
  &\leq e \constreg (p, T_0) \Big\{1 + \frac{1}{k \stepsize} \Big\} \sobopstatnorm{f}{\frac{4p}{p - 1}}^2 \sobopstatnorm{g}{2p}^2 \\
  &\qquad \qquad +  \frac{c'' \constreg (p, T_0) \stepsize}{\sqrt{k \stepsize - t_1}}  \sobokpstatnorm{f}{2}{\frac{2p}{p - 1}}  \sobokpstatnorm{g}{2}{2p}  \sobokpstatnorm{g}{1}{2p}  \sobokpstatnorm{f}{1}{\frac{4p}{p - 1}}.
\end{align*}
Taking average over $t_1$ and $t_2$, we conclude that
\begin{align*}
  |\mu_k| &\leq e \constreg (p, T_0) \Big\{1 + \frac{1}{k \stepsize} \Big\} \sobopstatnorm{f}{\frac{4p}{p - 1}}^2 \sobopstatnorm{g}{2p}^2\\
  &\qquad \qquad +  \sobokpstatnorm{f}{2}{\frac{2p}{p - 1}}   \sobokpstatnorm{g}{2}{2p}  \sobokpstatnorm{g}{1}{2p}  \sobokpstatnorm{f}{1}{\frac{4p}{p - 1}}\int_0^\stepsize \frac{c'' \constreg (p, T_0) \stepsize}{\sqrt{k \stepsize - t_1}} dt_1\\
  &\leq e \constreg (p, T_0) \Big\{1 + \frac{1}{k \stepsize} \Big\} \sobopstatnorm{f}{\frac{4p}{p - 1}}^2 \sobopstatnorm{g}{2p}^2 \\
  &\qquad \qquad +  \frac{c'' \constreg (p, T_0) \stepsize}{\sqrt{k \stepsize}}  \sobokpstatnorm{f}{2}{\frac{2p}{p - 1}}  \sobokpstatnorm{g}{2}{2p}  \sobokpstatnorm{g}{1}{2p}  \sobokpstatnorm{f}{1}{\frac{4p}{p - 1}}\\
  &\leq \big( e \constreg (p, T_0) + \constreg (p, T_0)^2 \big) \Big\{1 + \frac{1}{k \stepsize} \Big\} \sobopstatnorm{f}{\frac{4p}{p - 1}}^2 \sobopstatnorm{g}{2p}^2 + (c'')^2 \stepsize^2  \sobokpstatnorm{f}{2}{\frac{2p}{p - 1}}  \sobokpstatnorm{g}{2}{2p}^2,
\end{align*}
which completes the proof of Lemma~\ref{lemma:muk-approx-error-bound}.

\paragraph{Proof of \Cref{eq:additional-regularity-in-muk-approx-lemma}:}
Note that
\begin{align*}
  &\semigroup_{k \stepsize - t_1} \Big(f \cdot (\discount - \generator) \semigroup_{t_2} g \Big) (x) \\
  &= \int p_{k \stepsize - t_1} (y | x)  f(y) \Big\{ \discount \semigroup_{t_2} g - \inprod{\drift}{\nabla \semigroup_{t_2} g} - \frac{1}{2} \mathrm{Tr} \big( \covMat \nabla^2 \semigroup_{t_2} g \big) \Big\} (y) dy\\
  &= \discount \int p_{k \stepsize - t_1} (y | x)  f(y)  \semigroup_{t_2} g (y) dy \\
  &\qquad \qquad + \frac{1}{2} \int p_{k \stepsize - t_1} (y | x) \inprod{\nabla \semigroup_{t_2} g (y)}{\covMat (y) \nabla f (y) +  f (y) \Big( \nabla \cdot \covMat (y) + \nabla_y \log p_{k \stepsize - t_1} (y|x) - 2  \drift (y) \Big)} dy \\
  &= \discount \Exs_x \big[ f(x) \semigroup_{t_2} g (\State_{k \stepsize - t_1}) \big] + \frac{1}{2} \Exs_x \big[ \inprod{\nabla \semigroup_{t_2} g (\State_{k \stepsize - t_1})}{\covMat (\State_{k \stepsize - t_1}) \nabla f (\State_{k \stepsize - t_1})} \big] \\
  &\qquad \qquad +  \frac{1}{2} \Exs_x \big[ \inprod{\nabla \semigroup_{t_2} g (\State_{k \stepsize - t_1})}{ \nabla \cdot \covMat (\State_{k \stepsize - t_1}) + \nabla_y \log p_{k \stepsize - t_1} (\State_{k \stepsize - t_1}|x) - 2  \drift (\State_{k \stepsize - t_1}) } f (\State_{k \stepsize - t_1}) \Big].
\end{align*}
Therefore, we have the error decomposition
\begin{align*}
  &\statnorm{\semigroup_{k \stepsize - t_1} \Big(f \cdot (\discount - \generator) \semigroup_{t_2} g \Big)}^2 = \Exs \Big[ \semigroup_{k \stepsize - t_1} \Big(f \cdot (\discount - \generator) \semigroup_{t_2} g \Big) (\State_0)^2 \Big] \\
  &\leq 3 \discount^2 \Exs \big[ f(\State_0)^2 \semigroup_{t_2} g (\State_{k \stepsize - t_1})^2 \big]  + \lammax^2 \Exs \big[ \vecnorm{\nabla \semigroup_{t_2} g (\State_{k \stepsize - t_1})}{2}^2 \cdot \vecnorm{\nabla f (\State_{k \stepsize - t_1})}{2}^2 \big] \\
  &\qquad \qquad +  \Exs \Big[ \Big\{ (\smoothness_0^\drift)^2 \usedim + (\smoothness_1^\covMat)^2 \usedim + \vecnorm{\nabla_y \log p_{k \stepsize - t_1} (\MyState_{k \stepsize - t_1} | \MyState_0)}{2}^2 \Big\} \vecnorm{\nabla \semigroup_{t_2} g (\State_{k \stepsize - t_1})}{2}^2 f (\State_{k \stepsize - t_1})^2 \Big].
\end{align*}
Applying H\"{o}lder's inequality, we obtain that
\begin{align*}
  &\statnorm{\semigroup_{k \stepsize - t_1} \Big(f \cdot (\discount - \generator) \semigroup_{t_2} g \Big)}^2 \\
  &\leq \big(3 \discount^2 + \lammax^2 + (\smoothness_0^\drift)^2 \usedim + (\smoothness_1^\covMat)^2 \usedim \big) \cdot \sobokpstatnorm{\semigroup_{k \stepsize - t_1}  g}{1}{2p}^2 \sobokpstatnorm{f}{1}{\frac{2p}{p - 1}}^2\\
  &\qquad \qquad  + \Big\{\Exs \big[ \vecnorm{\nabla_y \log p_{k \stepsize - t_1} (\MyState_{k \stepsize - t_1} | \MyState_0)}{2}^{\frac{4p}{p - 1}} \big] \Big\}^{\frac{p - 1}{4p}} \cdot \sobokpstatnorm{\semigroup_{t_2}  g}{1}{2p}^2 \cdot \lpstatnorm{f}{\frac{4p}{p - 1}}^2.
\end{align*}
By \Cref{eq:log-density-regularity-y}, for $t \in [0, T_0]$, we have
\begin{align*}
   \Big\{\Exs \big[ \vecnorm{\nabla_y \log p_{k \stepsize - t_1} (\MyState_{k \stepsize - t_1} | \MyState_0)}{2}^{\frac{4p}{p - 1}} \big] \Big\}^{\frac{p - 1}{4p}} \leq \frac{c \big( \tfrac{4p}{3p + 1}, T_0, 1\big)}{\sqrt{k \stepsize - t_1}}.
\end{align*}
Putting them together, we conclude that
\begin{align*}
  \statnorm{\semigroup_{k \stepsize - t_1} \Big(f \cdot (\discount - \generator) \semigroup_{t_2} g \Big)} \leq \frac{\constreg (p, T_0)}{\sqrt{k \stepsize - t_1}} \sobokpstatnorm{\semigroup_{t_2}  g}{1}{2p} \cdot \lpstatnorm{f}{\frac{4p}{p - 1}}.
\end{align*}
By \Cref{eq:semigroup-sobo-norm-growth-bound}, we have $\sobokpstatnorm{\semigroup_{t_2}  g}{1}{2p} \leq e \sobokpstatnorm{g}{1}{2p}$. Substituting it to the bound above completes the proof of \Cref{eq:additional-regularity-in-muk-approx-lemma}.

\subsubsection{Proof of \Cref{lemma:muk-mixing-bound}}\label{subsubsec:proof-lemma-muk-mixing-bound}
We first bound the term $\mu_0$. By Cauchy--Schwarz inequality, we have
\begin{align*}
  \mu_0 \leq \Exs \Big[ f (\State_0)^2 \cdot \Big\{ \frac{1}{\stepsize} \int_0^\stepsize e^{- \discount t} (\discount - \generator) g (\State_{t}) dt \Big\}^2 \Big] \leq \frac{1}{\stepsize} \int_0^\stepsize \Exs \Big[ f (\State_0)^2 \Big( (\discount - \generator) g (\State_{t}) \Big)^2 \Big] dt.
\end{align*}
Applying H\"{o}lder's inequality to the integrand, we have
\begin{align*}
  \Exs \big[ \abss{f (\State_0) (\discount - \generator) g (\State_t)}^2 \big] \leq \lpstatnorm{(\discount - \generator) g}{2 p}^2 \cdot \lpstatnorm{f}{\frac{2p}{p - 1}}^2,
\end{align*}
Note that
\begin{align*}
  \lpstatnorm{(\discount - \generator ) g}{2 p}^{2p} &= \Exs \Big[ \abss{ \discount g - \inprod{\drift}{\nabla g} - \frac{1}{2} \mathrm{Tr} \big(\covMat \nabla^2 g \big) }^{2p} \Big]\\
  &\leq 3^{2p} \Big\{ \discount^{2p} \Exs \big[ |g|^{2p} \big] + \Exs \big[ \abss{\inprod{\drift}{\nabla g}}^{2p} \big] + \Exs \big[ \abss{\mathrm{Tr} \big(\covMat \nabla^2 g \big)}^{2p} \big] \Big\}\\
  &\leq (2 \usedim )^{2p}  \Big\{ \discount^{2p} \Exs \big[ |g|^{2p} \big] + (\smoothness_0^\drift)^{2p} \Exs \big[ \vecnorm{\nabla g}{2}^{2p} \big] + (\smoothness_0^\covMat)^{2p} \Exs \big[ \matsnorm{\nabla^2 g}{2}^{2p} \big] \Big\}\\
  &\leq \big(2 \usedim (\discount + \smoothness_0^\drift + \smoothness_0^\covMat) \big)^{2p} \sobokpstatnorm{g}{2}{2p}^{2p}.
\end{align*}
Substituting back to the inequality above, we have that
\begin{align}
  \Exs \big[ \abss{f (\State_0) (\discount - \generator) g (\State_t)}^2 \big] \leq c \lpstatnorm{f}{\frac{2p}{p - 1}}^2 \cdot\sobokpstatnorm{g}{2}{2p}^{2},\label{eq:generator-one-time-var-bound}
\end{align}
for any $t > 0$. As a result, we have
\begin{align*}
  |\mu_0| \leq c \lpstatnorm{f}{\frac{2p}{p - 1}}^2 \cdot\sobokpstatnorm{g}{2}{2p}^{2}.
\end{align*}
Now let us bound the term $\mu_k$ for $k \geq 1 + 1 / \stepsize$. We first need the following technical lemma, which establishes a mixing condition based on the Poincar\'{e} inequality~\ref{assume:markov-mixing}.
\begin{lemma}\label{lemma:poincare-contraction}
  Under Assumption~\ref{assume:markov-mixing}, for any $t > 0$, any zero random variable $Y \in \sigma (X_s: s \geq t)$ satisfies
  \begin{align*}
   \int \abss{\Exs_x \big[ Y \big]}^2 \stationary (x) dx \leq e^{- \lammin \poincare t } \Exs \big[ |Y|^2 \big].
  \end{align*}
\end{lemma}
We prove this lemma at the end of this section. Taking this lemma as given, let us now continue to bound the term $\mu_k$. By definition, we can rewrite $\mu_k$ as
\begin{align*}
  \mu_k &= \cov\Big( f (\State_0) \cdot \frac{1}{\stepsize} \int_0^{\stepsize} e^{- \discount t} (\discount - \generator) g (\State_t) dt, \frac{1}{\stepsize} \int_0^\stepsize e^{- \discount t} \Big\{ f \cdot (\discount - \generator) \semigroup_t g \Big\} (\State_{k \stepsize}) dt \Big) \\
   &= \cov\Big( f (\State_0) \cdot \frac{1}{\stepsize} \int_0^{\stepsize} e^{- \discount t} (\discount - \generator) g (\State_t) dt, \frac{1}{\stepsize} \int_0^\stepsize e^{- \discount t} \semigroup_{1} \Big\{ f \cdot (\discount - \generator) \semigroup_t g \Big\} (\State_{k \stepsize - 1}) dt \Big).
\end{align*}
Note that we have the measurability condition
\begin{align*}
  f (\State_0) \cdot \frac{1}{\stepsize} \int_0^{\stepsize} e^{- \discount t} (\discount - \generator) g (\State_t) dt &\in \sigma (X_t: 0 \leq t \leq \stepsize), \quad\mbox{and}\\
    \int_0^\stepsize e^{- \discount t} \semigroup_{1} \Big\{ f \cdot (\discount - \generator) \semigroup_t g \Big\} (\State_{k \stepsize - 1}) dt &\in \sigma (X_t: t \geq k \stepsize - 1).
\end{align*}
\Cref{lemma:poincare-contraction} yields the bound
\begin{align*}
  &\var \Bigg( \Exs \Big[ \int_0^\stepsize \semigroup_{1} \Big\{ f \cdot (\discount - \generator) \semigroup_t g \Big\} (\State_{k \stepsize - 1}) dt  \mid X_\stepsize \Big]\Bigg) \\
  &\leq e^{- \lammin \poincare (k \stepsize - 1 - \stepsize)} \var \Big(  \int_0^\stepsize \semigroup_{1} \Big\{ f \cdot (\discount - \generator) \semigroup_t g \Big\} (\State_{k \stepsize - 1}) dt \Big)\\
  &\leq e^{- \lammin \poincare (k \stepsize - 1 - \stepsize)} \Exs \Big[ \abss{ \int_0^\stepsize \semigroup_{1} \Big\{ f \cdot (\discount - \generator) \semigroup_t g \Big\} (\State_{k \stepsize - 1}) dt}^2 \Big].
\end{align*}
By Cauchy--Schwarz inequality, we have
\begin{align*}
  \mu_k^2 &\leq \var\Big( f (\State_0) \cdot \frac{1}{\stepsize} \int_0^{\stepsize} e^{- \discount t} (\discount - \generator) g (\State_t) dt \Big) \cdot \var \Big( \Exs \Big[ 
  \frac{1}{\stepsize} \int_0^\stepsize \semigroup_{1} \Big\{ f \cdot (\discount - \generator) \semigroup_t g \Big\} (\State_{k \stepsize - 1}) dt  \mid X_\stepsize \Big] \Big)\\
  &\leq e^{- \lammin \poincare (k \stepsize - 1 - \stepsize)} \Exs \Big[\abss{ f (\State_0) \cdot \tfrac{1}{\stepsize} \int_0^{\stepsize} e^{- \discount t} (\discount - \generator) g (\State_t) dt}^2 \Big] \cdot \Exs \Big[ \abss{ \tfrac{1}{\stepsize} \int_0^\stepsize \semigroup_{1} \Big\{ f \cdot (\discount - \generator) \semigroup_t g \Big\} (\State_{k \stepsize - 1}) dt}^2 \Big].
\end{align*}
By \Cref{eq:generator-one-time-var-bound}, we have
\begin{align*}
  \Exs \Big[ \abss{f (\State_0) \cdot \frac{1}{\stepsize} \int_0^{\stepsize} e^{- \discount t} (\discount - \generator) g (\State_t) dt }^2 \leq c \lpstatnorm{f}{\frac{2p}{p - 1}}^2 \cdot\sobokpstatnorm{g}{2}{2p}^{2}.
\end{align*}
On the other hand, invoking \Cref{eq:additional-regularity-in-muk-approx-lemma} in the proof of \Cref{lemma:muk-approx-error-bound} (with $T_0 = 1$ and $t_2 = t$), we have
\begin{align*}
  &\Exs \Big[ \abss{ \frac{1}{\stepsize}\int_0^\stepsize e^{- \discount t} \semigroup_{1} \Big\{ f \cdot (\discount - \generator) \semigroup_t g \Big\} (\State_{k \stepsize - 1}) dt}^2 \Big] \\
  &\leq  \frac{1}{\stepsize}\int_0^\stepsize  \Exs \Big[ \abss{\semigroup_{1} \Big\{ f \cdot (\discount - \generator) \semigroup_t g \Big\} (\State_{k \stepsize - 1}) }^2 \Big] dt\\
  & \leq \constreg (p, 1)^2 \sobopstatnorm{g}{2p}^2 \cdot \sobopstatnorm{f}{\frac{4p}{p - 1}}^2.
\end{align*}
Putting them together, for $k \geq 2 + 2 / \stepsize$, we have
\begin{align*}
  |\mu_k| &\leq  c \cdot \constreg (p, 1) \exp \Big( - \frac{\lammin \poincare}{4} k \stepsize \Big)  \sobopstatnorm{g}{2p} \cdot \sobopstatnorm{f}{\frac{4p}{p - 1}} \cdot \lpstatnorm{f}{\frac{2p}{p - 1}} \cdot \sobokpstatnorm{g}{2}{2p}\\
  &\leq  c_0 \exp \Big( - \frac{\lammin \poincare}{4} k \stepsize \Big)\sobopstatnorm{g}{2p} \cdot \sobokpstatnorm{g}{2}{2p} \cdot \sobopstatnorm{f}{\frac{4p}{p - 1}}^2,
\end{align*}
which completes the proof of \Cref{lemma:muk-mixing-bound}.

\paragraph{Proof of \Cref{lemma:poincare-contraction}:}
For any function $f \in \mathbb{H}^1 (\stationary)$, following the derivation in Lemma 5 of the paper~\cite{mou2024bellman}, we note that
\begin{align*}
  - \statinprod{f}{\generator f} &= \int f (x) \Big\{ \inprod{\drift}{\nabla f} + \frac{1}{2} \mathrm{Tr} \big( \nabla^2 f \big) \Big\} (x) \stationary (x) dx\\
  &= \frac{1}{2} \int \nabla f (x)^\top \covMat (x) \nabla f (x) \stationary (x) dx - \frac{1}{2} \int f (x)^2 \Big\{ - \nabla \cdot (\stationary \drift) + \frac{1}{2} \nabla^2 \cdot (\covMat^2 \stationary) \Big\} (x) dx\\
  &\geq \frac{\lammin}{2} \statnorm{\nabla f}^2,
\end{align*}
where in the last step, we use the Fokker--Planck equation $- \nabla \cdot (\stationary \drift) + \frac{1}{2} \nabla^2 \cdot (\covMat^2 \stationary) = 0$ for the stationary distribution $\stationary$.

By Assumption~\ref{assume:markov-mixing}, if $\Exs_\stationary [f (\State)] = 0$, we have
\begin{align}
  - \statinprod{f}{\generator f} \geq \frac{\lammin}{2} \statnorm{\nabla f}^2 \geq \frac{\lammin \poincare}{2} \statnorm{f}^2.\label{eq:poincare-ineq-for-diffusion-generator}
\end{align}

Since $Y$ is measurable in the $\sigma$-field $\sigma (X_s: s \geq t)$, by Markov property, the conditional expectation $\Exs \big[ Y | \sigma (X_s: 0 \leq s \leq t) \big]$ is a deterministic function of $X_t$. We use $f$ to denote this function, i.e.,
\begin{align*}
  f (X_t) = \Exs \big[ Y | \sigma (X_s: 0 \leq s \leq t) \big].
\end{align*}
Clearly we have $\int f (x) \stationary (x) dx = \Exs [Y] = 0$. Furthermore, we note that
\begin{align*}
  \semigroup_t f (x) = \Exs \big[ f (\State_t )\mid \State_0 = x \big] = \Exs_x [Y].
\end{align*}
Letting $h_s \mydefn \statnorm{\semigroup_s f}^2$ for any $s > 0$, we have
\begin{align*}
  \frac{d h_s}{ds} \overset{(i)}{=} 2 \statinprod{\semigroup_s f}{\generator \semigroup_s f} \overset{(ii)}{\leq} - \lammin \poincare \statnorm{\semigroup_s f}^2 = - 2 \poincare h_s,
\end{align*}
where in step $(i)$, we interchange the order of the semigroup $\semigroup_t$ and its generator $\generator$, and in step $(ii)$, we use \Cref{eq:poincare-ineq-for-diffusion-generator}. By Gr\"{o}nwall inequality, we conclude that
\begin{align*}
  h_t \leq e^{- 2 \poincare} h_0 = e^{- \poincare \lammin} \Exs \big[ f (X_0)^2 \big] \leq e^{- 2 \poincare} \Exs \big[ Y^2 \big],
\end{align*}
which completes the proof of this lemma.

\subsection{Proof of~\Cref{thm:main}}\label{subsec:proof-thm-main}
Define the random matrices and vectors
\begin{align*}
  A_k &\mydefn \frac{1}{\stepsize} \psi (\MyState_{k \stepsize})  \cdot \Big\{ \psi (\MyState_{k \stepsize})^\top - e^{- \discount (\numerOrder - 1) \stepsize}  \psi (\MyState_{ (k + \numerOrder - 1) \stepsize})^\top \Big\},\\
  b_k &\mydefn  \frac{1}{\stepsize}\sum_{i = 0}^{\numerOrder - 1} \kappa_i \reward \big( \MyState_{(k + i) \stepsize} \big) \psi (\MyState_{k \stepsize}), \quad \mbox{and} \\
  \varepsilon_k & \mydefn \sum_{i = 0}^{\numerOrder - 1} \kappa_i \Big\{ \Reward_{(k + i) \stepsize} -  \reward \big( \MyState_{(k + i) \stepsize} \big) \Big\} \psi (\MyState_{k \stepsize}).
\end{align*}
Denote $N \mydefn \totaltime / \stepsize - \numerOrder + 1$. The estimator $\thetahat_\totaltime$ takes the form
\begin{align*}
  \thetahat_\totaltime = \Big( \underbrace{\frac{1}{N} \sum_{k = 0}^{N - 1} A_k}_{=: \Ahat_N} \Big)^{-1} \cdot \Big\{\underbrace{ \frac{1}{N} \sum_{k = 0}^{N - 1} b_k }_{=: \bhat_N} + \underbrace{ \frac{1}{N} \sum_{k = 0}^{N - 1} \varepsilon_k }_{=: \epshat_N} \Big\}.
\end{align*}
Defining the population-level objects
\begin{align*}
  \Abar \mydefn \Exs_{\MyState_k \sim \stationary} \big[ A_k \big], \quad \mbox{and} \quad \bbar \mydefn \Exs_{\MyState_k \sim \stationary} \big[ b_k \big],
\end{align*}
we let $\thetabar \mydefn \Abar^{-1} \bbar$.

Note that the Hilbert inner product structures in the space value functions can be related to quadratic forms under the basis function representations. In particular, for any pair $\theta_1, \theta_2 \in \real^\mbasis$ of vectors, we have
 \begin{subequations}\label{eqs:relate-function-norm-to-basis}
   \begin{align}
     \statinprod{\theta_1^\top \psi}{\theta_2^\top \psi} &= \theta_1^\top \Exs [\psi (\MyState) \psi (\MyState)^\top]\theta_2 = \theta_1^\top H_0 \theta_2, \quad \mbox{and}\\
     \soboinprod{\theta_1^\top \psi}{\theta_2^\top \psi} &= \theta_1^\top \Big\{ \Exs [\psi (\MyState) \psi (\MyState)^\top] + \Exs [\nabla \psi (\MyState)  \nabla \psi (\MyState)^\top] \Big\} \theta_2 = \theta_1^\top H_1 \theta_2.
   \end{align}
 \end{subequations}
 By triangle inequality, the error of interest can be written as
 \begin{align*}
   \sobonorm{\valuehat_\totaltime - \ValTrue} &\leq \sobonorm{\valuebar - \ValTrue} + \sobonorm{\valuehat_\totaltime - \valuebar} \\
   &\leq \sobonorm{\valuebar - \ValTrue} + \vecnorm{\thetahat_\totaltime - \thetabar}{H_1}.
 \end{align*}
 It remains to bound the estimation error $\vecnorm{\thetahat_\totaltime - \thetabar}{H_1}$ in the Euclidean space. By definition, we find that
\begin{align}
  \Abar \big(\thetahat_\totaltime - \thetabar \big) = \Big\{ \bhat_T - \bbar + \epshat_T - (\Ahat_T - \Abar) \thetabar \Big\} - (\Ahat_T - \Abar) \big(\thetahat_\totaltime - \thetabar \big).\label{eq:basic-ineq-for-sample-based-bellman}
\end{align}
Therefore, in order to bound the estimation error, we need to provide PSD lower bounds on the matrix $\Abar$, and upper bounds on the random fluctuations on the right hand side.

We first establish the following lemma, which controls the conditioning of the population-level matrix $\Abar$ under the $\vecnorm{\cdot}{H_1}$-geometry.
\begin{lemma}\label{lemma:geometry-of-abar}
  Under the setup of Theorem~\ref{thm:main}, the population-level matrix $\Abar$ satisfies
  \begin{align*}
    \opnorm{H_1^{1/2} \Abar^{-1} H_1^{1/2} } \leq  \frac{e^2 \lammax}{ \lammin} \max (1, 1/\discount, 1/\lammin) 
  \end{align*}
\end{lemma}
\noindent See Section~\ref{subsubsec:proof-lemma-abar-geometry} for the proof of this lemma.

We use the following lemma to establish the concentration behavior of the random matrix
\begin{lemma}\label{lemma:matrix-conc-for-bellman-eq}
  Under above setup, for any $\delta \in (0, 1)$ with probability $1 - \delta$, we have
  \begin{multline*}
    \matsnorm{\Abar^{-1} \big( \Ahat_N - \Abar \big)}{\op, H_1}\\
     \leq \tau^2 \sqrt{\frac{\mbasis}{T} \log (\mbasis / \delta)} \Big\{  \constScaryReg (T_0) \log ( 1/ \stepsize) + \frac{\constScary}{\poincare} e^{- \lammin \poincare T_0 / 4} \mbasis^{2 \omega} \Big\}^{1/2} + \frac{c D_\mbasis^2}{\poincare T} \log^{3/2} \big( \frac{T \mbasis}{\stepsize \delta} \big).
  \end{multline*}
\end{lemma}
\noindent See Section~\ref{subsubsec:proof-lemma-matrix-conc-for-bellman-eq} for the proof of this lemma.

As for the main error term in Eq~\eqref{eq:basic-ineq-for-sample-based-bellman}, we have
\begin{lemma}\label{lemma:main-err-term-in-sample-based-bellman}
  Under the setup of \Cref{thm:main}, we have
  \begin{align*}
    &\Exs \Big[ \vecnorm{H_1^{-1/2}\big( \bhat_N - \bbar + \epshat_N - (\Ahat_N - \Abar) \thetabar \big)}{2}^2 \Big]\\
    & \leq  \frac{\tau^2 \mbasis}{T} \Big\{ \constScaryReg (T_0) \sobopstatnorm{\Delstar}{2p}^2 \log (1 / \stepsize) + \constScary \highorder_{T_0, \stepsize} (\Delstar) \Big\}\\
    &\qquad \qquad +  \frac{\Tthres (\mbasis, T_0)}{T} \big( \sobonorm{\Delstar}^2 + \stepsize^{2 \numerOrder} \big) + \frac{\tau^4 \constScary}{T} \mathrm{Tr} \big( H_1^{-1} H_0 \big) (\stepsize + \sobonorm{\fbar}^2).
  \end{align*}
\end{lemma}
\noindent See Section~\ref{subsubsec:proof-lemma-main-err-term-in-sample-based-bellman} for its proof.

Taking these lemmas as given, let us proceed with the proof of \Cref{thm:main}. By \Cref{lemma:geometry-of-abar}, we have
\begin{align*}
  &\vecnorm{\Abar^{-1} \big( \bhat_N - \bbar + \epshat_N - (\Ahat_N - \Abar) \thetabar \big)}{H_1} = \vecnorm{H_1^{1/2} \Abar^{-1} \big( \bhat_N - \bbar + \epshat_N - (\Ahat_N - \Abar) \thetabar \big)}{2}\\
  & \leq \opnorm{H_1^{1/2} \Abar^{-1} H_1^{1/2}} \cdot \vecnorm{H_1^{-1/2} \big( \bhat_N - \bbar + \epshat_N - (\Ahat_N - \Abar) \thetabar \big)}{2}\\
  &\leq  \frac{e^2 \lammax}{ \lammin} \max (1, 1/\discount, 1/\lammin)  \vecnorm{H_1^{-1/2} \big( \bhat_N - \bbar + \epshat_N - (\Ahat_N - \Abar) \thetabar \big)}{2}.
\end{align*}
Define the event
\begin{align*}
  \Event \mydefn \Big\{ \mbox{The bound in \Cref{lemma:matrix-conc-for-bellman-eq} holds true} \Big\}.
\end{align*}
By \Cref{lemma:matrix-conc-for-bellman-eq}, we have $\Prob (\Event) \geq 1 - \delta$. On the event $\Event$, given the trajectory length satisfying $T \geq 2 \Tthres (\mbasis, T_0) + \tfrac{2 c D_\mbasis^2}{\poincare} \log^{3/2} \big( \tfrac{T \mbasis}{\delta \stepsize} \big)$, we have
\begin{align*}
  \matsnorm{\Abar^{-1} \big( \widehat{A}_N - \Abar \big)}{\op, H_1} \leq \frac{1}{2}.
\end{align*}
By the decomposition in Eq~\eqref{eq:basic-ineq-for-sample-based-bellman}, on the event $\Event$, we have
\begin{align*}
  \vecnorm{\thetahat_\totaltime - \thetabar}{H_1} &\leq \vecnorm{\Abar^{-1} \big( \bhat_N - \bbar + \epshat_N - (\Ahat_N - \Abar) \thetabar \big)}{H_1} + \matsnorm{\Abar^{-1} \big( \Ahat_N - \Abar \big)}{op, H_1} \vecnorm{\thetahat_\totaltime - \thetabar}{H_1}\\
  &\leq \frac{e^2 \lammax}{ \lammin} \max (1, 1/\discount, 1/\lammin)  \vecnorm{H_1^{-1/2} \big( \bhat_N - \bbar + \epshat_N - (\Ahat_N - \Abar) \thetabar \big)}{2} + \frac{1}{2} \vecnorm{\thetahat_\totaltime - \thetabar}{H_1}.
\end{align*}
Consequently, we have the moment bound
\begin{align*}
  &\Exs \big[ \vecnorm{\thetahat_\totaltime - \thetabar}{H_1}^2 \bm{1}_{\Event} \big]\\
  &\leq \frac{4 e^4 \lammax^2}{ \lammin^2} \max (1, 1/\discount, 1/\lammin)^2 \Exs \big[ \vecnorm{H_1^{-1/2} \big( \bhat_N - \bbar + \epshat_N - (\Ahat
  _N - \Abar) \thetabar \big)}{2}^2  \big].
\end{align*}
Invoking \Cref{lemma:main-err-term-in-sample-based-bellman}, we have
\begin{align*}
  \Exs \big[ \vecnorm{\thetahat_\totaltime - \thetabar}{H_1}^2 \bm{1}_{\Event} \big] &\leq \frac{\tau^2 \mbasis}{T} \Big\{ \constScaryReg (T_0) \sobopstatnorm{\Delstar}{2p}^2 \log (1 / \stepsize) + \constScary \highorder_{T_0, \stepsize} (\Delstar) \Big\}\\
    &\qquad \qquad +  \frac{\Tthres (\mbasis, T_0)}{T} \big( \sobonorm{\Delstar}^2 + \stepsize^{2 \numerOrder} \big) + \frac{\tau^4 \constScary}{T} \mathrm{Tr} \big( H_1^{-1} H_0 \big) (\stepsize + \sobonorm{\fbar}^2)
\end{align*}
Now that we have the statistical error bound. As for the approximation error bound, we invoke the population-level analysis in Theorem 3 of the paper \cite{mou2024bellman} to obtain
\begin{align*}
  \sobonorm{\fbar - \ValTrue} \leq c_1 \sobonorm{\Delstar} + \constScary \stepsize^{\numerOrder}.
\end{align*}
Given $T \geq \Tthres$, we have
\begin{align*}
  \Exs \Big[ \sobonorm{\valuehat_T - \ValTrue}^2 \bm{1}_\Event \Big] &\leq c_1 \sobonorm{\Delstar}^2 + \constScary \stepsize^{2 \numerOrder}\\
  &\qquad+ \frac{\tau^2 \mbasis}{T} \Big\{ \constScaryReg (T_0) \sobopstatnorm{\Delstar}{2p}^2 \log (1 / \stepsize) + \constScary \highorder_{T_0, \stepsize} (\Delstar) \Big\}\\
  &\qquad \qquad + \frac{\tau^4 \constScary}{T} \mathrm{Tr} \big( H_1^{-1} H_0 \big) (\stepsize + \sobonorm{\ValTrue}^2),
\end{align*}
which completes the proof of \Cref{thm:main}.

\subsubsection{Proof of Lemma~\ref{lemma:geometry-of-abar}}\label{subsubsec:proof-lemma-abar-geometry}
By Lemma 5 and Lemma 7 of \cite{mou2024bellman}, given a stepsize satisfying the condition $\stepsize \leq c \mbasis^{- 4 \omega}$, for any $\theta \in \real^\mbasis$, we have
 \begin{align}
   \theta^\top \Abar \theta &= \stepsize^{-1} \Big\{ \Exs \big[ (\theta^\top \psi) (X_{k \stepsize})^2 \big] - e^{- \discount (\numerOrder - 1) \stepsize}  \Exs \big[ (\theta^\top \psi) (X_{k \stepsize}) \cdot (\theta^\top \psi) (X_{ (k + \numerOrder - 1) \stepsize} ) \big] \Big\} \nonumber \\
   &= \stepsize^{-1} \statinprod{\theta^\top \psi}{\big( \IdMat - e^{- \discount (\numerOrder - 1) \stepsize} \semigroup_{(\numerOrder - 1)  \stepsize} \big) (\theta^\top \psi)} \nonumber \\
   &\geq \frac{\lammin}{e^2 \lammax} \min (1, \discount, \lammin) \vecnorm{\theta}{H_1}^2.\label{eq:abar-psd-lower-bound}
 \end{align}
Given a pair of vectors $(x, y)$ satisfying $y = H_1^{1/2} \Abar^{-1} H_1^{1/2} x$, by Eq~\eqref{eq:abar-psd-lower-bound}, we have
\begin{align*}
  x^\top y = y^\top H_1^{- 1/2} \Abar H_1^{- 1/2} y \geq \frac{\lammin}{e^2 \lammax} \min (1, \discount, \lammin) \vecnorm{H_1^{-1/2} y}{H_1}^2 = \frac{\lammin}{e^2 \lammax} \min (1, \discount, \lammin) \vecnorm{y}{2}^2.
\end{align*}
Consequently, we have $\vecnorm{x}{2} \geq \frac{\lammin}{e^2 \lammax} \min (1, \discount, \lammin) \vecnorm{y}{2}$. Since the choice of $x$ is arbitrary, we conclude that
\begin{align*}
  \opnorm{H_1^{1/2} \Abar^{-1} H_1^{1/2}} = \sup_{\vecnorm{x}{2} \leq 1} \vecnorm{H_1^{1/2} \Abar^{-1} H_1^{1/2} y} \leq  \frac{e^2 \lammax}{ \lammin} \max (1, 1/\discount, 1/\lammin).
\end{align*}
\subsubsection{Proof of Lemma~\ref{lemma:matrix-conc-for-bellman-eq}}\label{subsubsec:proof-lemma-matrix-conc-for-bellman-eq}
By \Cref{lemma:geometry-of-abar}, we have
\begin{align}
  &\matsnorm{\Abar^{-1} \big( \Ahat_N - \Abar \big)}{\op, H_1} = \opnorm{H_1^{1/2} \Abar^{-1} \big( \Ahat_N - \Abar \big) H_1^{-1/2}} \nonumber \\
  &\leq  \opnorm{H_1^{1/2} \Abar^{-1} H_1^{1/2}} \cdot \opnorm{H_1^{-1/2} \big( \Ahat_N - \Abar \big) H_1^{-1/2}} \nonumber \\
  & \leq \frac{e^2 \lammax}{ \lammin} \max (1, 1/\discount, 1/\lammin)\opnorm{H_1^{-1/2} \big( \Ahat_N - \Abar \big) H_1^{-1/2}}. \label{eq:matrix-concentration-op-norm-conversion}
\end{align}
Applying It\^{o}'s formula, we note that
\begin{multline*}
  A_k = \stepsize^{-1} \psi (\MyState_{k \stepsize}) \cdot \int_0^{(\numerOrder - 1) \stepsize} e^{- \discount t} \Big\{\discount \psi (\MyState_{k \stepsize + t}) - \generator \psi (\MyState_{k \stepsize + t}) \Big\}^\top dt\\
   + \stepsize^{-1} \psi (\MyState_{k \stepsize})  \Big\{ \int_0^{(\numerOrder - 1) \stepsize} e^{- \discount t} \nabla \psi (\MyState_{k \stepsize + t}) \covMat^{1/2} (\MyState_{k \stepsize + t}) dB_{k \stepsize + t} \Big\}^\top.
\end{multline*}
Define the summations
\begin{align*}
  I_1 &\mydefn \stepsize^{-1} \sum_{k = 0}^{N - 1}\psi (\MyState_{k \stepsize}) \cdot \int_0^{(\numerOrder - 1) \stepsize} e^{- \discount t} \Big\{\discount \psi (\MyState_{k \stepsize + t}) - \generator \psi (\MyState_{k \stepsize + t}) \Big\}^\top dt,\\
  I_2 &\mydefn \stepsize^{-1} \sum_{k = 0}^{N - 1} \psi (\MyState_{k \stepsize})  \Big\{ \int_0^{(\numerOrder - 1) \stepsize} e^{- \discount t} \nabla \psi (\MyState_{k \stepsize + t}) \covMat^{1/2} (\MyState_{k \stepsize + t}) dB_{k \stepsize + t} \Big\}^\top.
\end{align*}
We have $\Exs[I_1] = N \Abar$ and $\Exs [I_2] = 0$. It suffices to bound the matrix fluctuation errors $N^{-1} I_1 - \Abar$ and $N^{-1} I_2$, respectively.

\paragraph{Concentration bounds for the term $I_1$:} We start by decomposing the summation into $(\numerOrder - 1)$ terms. We define
\begin{align*}
  I_1^{(j)} \mydefn \sum_{\substack{k \in [0, N - 1] \\ k \equiv j \mod (\numerOrder - 1)} } \psi (\MyState_{k \stepsize}) \cdot \stepsize^{-1} \int_0^{(\numerOrder - 1) \stepsize} e^{- \discount t} \Big\{\discount \psi (\MyState_{k \stepsize + t}) - \generator \psi (\MyState_{k \stepsize + t}) \Big\}^\top dt, \quad \mbox{for $j = 0,1,\cdots \numerOrder - 2$}.
\end{align*}
Note that for a fixed index $j$, the time integrals in each term of $I_1^{(j)}$ are non-overlapping.

In order to bound the concentration of such an integral, we use the following technical lemma on concentration of random matrices in Markov processes.
\begin{lemma}\label{lemma:matrix-conc-markov}
  Let $(X_t)_{t \geq 0}$ be a stationary Markov process satisfying Assumption~\ref{assume:markov-mixing}. Given a matrix-valued functional $\Upsilon: C ([0, \dt]; \real^\usedim) \rightarrow \real^{m_1 \times m_2}$, such that $\opnorm{\Upsilon (\{x_t: 0 \leq t \leq \dt\})} \leq B$ for any input process $x$, we define $Y_k \mydefn \Upsilon \big( \{X_{t + k \dt}: 0 \leq t \leq \dt\} \big) $ for $k = 0,1,2,\cdots$. If $\Exs [Y_0] = 0$, for any $\delta \in (0, 1)$, with probability $1 - \delta$, we have
  \begin{align*}
    \opnorm{\frac{1}{N}\sum_{k = 0}^N Y_k}
    \leq 8 \Big( \sum_{i = 0}^{+ \infty} \big( \opnorm{\Exs \big[ Y_0 Y_i^\top \big]} + \opnorm{\Exs \big[ Y_0^\top Y_i \big]} \big) \Big)^{1/2} \sqrt{\frac{\log \tfrac{m_1 + m_2}{\delta}}{N}} + \frac{320 B}{\lammin \poincare \stepsize N} \log \tfrac{m_1 + m_2}{\delta}.
  \end{align*}
\end{lemma}
\noindent See \Cref{subsec:proof-lemma-matrix-conc-markov} for the proof of this lemma. Let us now use this lemma to bound the random matrix $I_1^{(j)}$. We apply this lemma with step length $\stepsize$ replaced by $(\numerOrder - 1) \stepsize$, and the random matrices
\begin{align*}
  \widetilde{Y}_k \mydefn \frac{\psi (\MyState_{\ell (k) \stepsize})}{\stepsize} \int_0^{(\numerOrder - 1) \stepsize} e^{- \discount t} \Big\{\discount \psi (\MyState_{\ell (k) \stepsize + t}) - \generator \psi (\MyState_{\ell (k)  \stepsize + t}) \Big\}^\top dt,
\end{align*}
where we define the time index $\ell (k) \mydefn k (\numerOrder - 1) + j$. We further let $Y_k \mydefn \widetilde{Y}_k - \Exs [\widetilde{Y}_k]$.

In order to apply the lemma, we start by noting that Assumption~\ref{assume:bounded-feature} implies that
\begin{align*}
  \opnorm{H_1^{-1/2} Y_k H_1^{-1/2}} \leq (1 + \discount) D_\mbasis^2.
\end{align*}
As for the cross-covariance, for any vector $u \in \sphere^{\mbasis - 1}$, we note that
\begin{align*}
  &u^\top H_1^{-1/2} \Exs [Y_0 H_1^{-1} Y_k^\top] H_1^{-1/2} u  \\
  &= \sum_{j = 1}^\mbasis \cov \Big( \tfrac{u^\top H_1^{-1/2} \psi (\MyState_{\ell (0) \stepsize})}{\stepsize} \int_0^{(\numerOrder - 1) \stepsize} e^{- \discount t} \big(\discount  - \generator \big) \coordinate_j^\top H_1^{-1/2} \psi (\MyState_{\ell (0)  \stepsize + t}) dt,\\
   &\qquad \qquad\qquad\qquad\qquad \frac{u^\top H_1^{-1/2} \psi (\MyState_{\ell (k) \stepsize})}{\stepsize} \int_0^{(\numerOrder - 1) \stepsize} e^{- \discount t} \big(\discount  - \generator \big) \coordinate_j^\top H_1^{-1/2} \psi (\MyState_{\ell (k)  \stepsize + t}) dt \Big)\\
  &= (\numerOrder - 1)^2 \sum_{j = 1}^\mbasis \mu_k \big( u^\top H_1^{-1/2} \psi, \coordinate_j^\top H_1^{-1/2} \psi_j \big),
\end{align*}
where we slightly abuse the notation in the definition of $\mu_k$ by taking the steplength as $(\numerOrder - 1) \stepsize$ instead of $\stepsize$. By \Cref{lemma:muk-approx-error-bound,lemma:muk-mixing-bound} in conjunction with the hyper-contractivity assumption~\fakerefassumehyper{$\frac{4p}{p - 1}$}{$\tau$}, we have
\begin{align*}
  &\abss{\mu_k \big( u^\top H_1^{-1/2} \psi, \coordinate_j^\top  H_1^{-1/2} \psi \big)} \\
  &\leq \begin{dcases}
    c_0 \tau^4 \statnorm{u^\top H_1^{-1/2} \psi}^2 \sobohilnorm{\coordinate_j^\top H_1^{-1/2} \psi}{2}^2, &k = 0,\\
   \tfrac{\constreg (p, T_0) \tau^4}{k \stepsize} \sobonorm{u^\top H_1^{-1/2} \psi}^2 \sobonorm{\coordinate_j^\top  H_1^{-1/2} \psi}^2 + c' \stepsize^2 \tau^4 \sobohilnorm{u^\top H_1^{-1/2} \psi}{2}^2 \sobohilnorm{\coordinate_j^\top  H_1^{-1/2} \psi}{2}^2, &1 \leq k \leq \tfrac{T_0}{\stepsize},\\
   c_0 \tau^4 \exp \Big( - \frac{\lammin \poincare}{4} k \stepsize \Big) \sobonorm{u^\top H_1^{-1/2} \psi}^2 \sobonorm{\coordinate_j^\top  H_1^{-1/2} \psi} \sobohilnorm{\coordinate_j^\top  H_1^{-1/2} \psi}{2} ,&k \geq T_0 / \stepsize.
  \end{dcases}
\end{align*}
Since $\vecnorm{u}{2} = 1$, we have that
\begin{align*}
   \sobonorm{u^\top H_1^{-1/2} \psi}^2 &= \Exs_\stationary \Big[ \abss{u^\top H_1^{-1/2} \psi (\State)}^2 \Big] + \Exs_\stationary \Big[ \vecnorm{ \nabla \big( u^\top H_1^{-1/2} \psi \big) (\State)}{2}^2 \Big] \\
   &= u^\top H_1^{-1/2} \Big\{  \Exs [\psi (\MyState) \psi (\MyState)^\top] + \Exs [\nabla \psi (\MyState)  \nabla \psi (\MyState)^\top] \Big\} H_1^{-1/2} u\\
   &= \vecnorm{u}{2}^2 = 1.
\end{align*}
Similarly, we have $\sobonorm{\coordinate_j^\top H_1^{-1/2} \psi} =1$ for each $j \in [\mbasis]$. By the basis function growth condition~\ref{assume:basis-condition}, we have
\begin{align*}
  \sobohilnorm{u^\top H_1^{-1/2} \psi}{2} \leq \mbasis^{\omega}, \quad \sobohilnorm{\coordinate_j^\top H_1^{-1/2} \psi}{2} \leq \mbasis^{\omega}.
\end{align*}
Substituting back to the bound on $|\mu_k|$, we conclude that
\begin{align*}
  \opnorm{H_1^{-1/2} \Exs [Y_0 H_1^{-1} Y_k^\top] H_1^{-1/2}} &= \sup_{u \in \sphere^{\mbasis - 1}} \abss{u^\top H_1^{-1/2} \Exs [Y_0 H_1^{-1} Y_k^\top] H_1^{-1/2} u}\\
  &\leq (\numerOrder - 1)^2 \sum_{j = 1}^\mbasis \sup_{u \in \sphere^{\mbasis - 1}} \abss{\mu_k \big( u^\top H_1^{-1/2} \psi, \coordinate_j^\top H_1^{-1/2} \psi_j \big)}\\
  &\leq \tau^4 \numerOrder^2 \mbasis \cdot \begin{dcases}
    c_0  \mbasis^{2 \omega} & k = 0,\\
    \frac{\constreg (p, T_0)}{k \stepsize} + c' \stepsize^2 \mbasis^{4 \omega} & 1 \leq k \leq T_0 / \stepsize,\\
    c_0 \exp \Big( - \frac{\lammin \poincare}{4} k \stepsize \Big) \mbasis^{2 \omega} & k \geq T_0 / \stepsize.
  \end{dcases}
\end{align*}
Similarly, we note that
\begin{align*}
  &u^\top H_1^{-1/2} \Exs [Y_0^\top H_1^{-1} Y_k] H_1^{-1/2} u\\
    &= \sum_{j = 1}^\mbasis \cov \Big( \tfrac{\coordinate_j^\top H_1^{-1/2} \psi (\MyState_{\ell (0) \stepsize})}{\stepsize} \int_0^{(\numerOrder - 1) \stepsize} e^{- \discount t} \big(\discount  - \generator \big) u^\top H_1^{-1/2} \psi (\MyState_{\ell (0)  \stepsize + t}) dt,\\
   &\qquad \qquad\qquad\qquad\qquad \frac{\coordinate_j^\top H_1^{-1/2} \psi (\MyState_{\ell (k) \stepsize})}{\stepsize} \int_0^{(\numerOrder - 1) \stepsize} e^{- \discount t} \big(\discount  - \generator \big) u^\top H_1^{-1/2} \psi (\MyState_{\ell (k)  \stepsize + t}) dt \Big)\\
  &= (\numerOrder - 1)^2 \sum_{j = 1}^\mbasis \mu_k \big( \coordinate_j^\top H_1^{-1/2} \psi, u^\top H_1^{-1/2} \psi_j \big).
\end{align*}
Following the same arguments, we have
\begin{align*}
  \opnorm{H_1^{-1/2} \Exs [Y_0^\top H_1^{-1} Y_k] H_1^{-1/2}} \leq\tau^4 \numerOrder^2 \mbasis \cdot \begin{dcases}
    c_0  \mbasis^{2 \omega} & k = 0,\\
    \frac{\constreg (p, T_0)}{k \stepsize} + c' \stepsize^2 \mbasis^{4 \omega} & 1 \leq k \leq T_0 / \stepsize,\\
    c_0 \exp \Big( - \frac{\lammin \poincare}{4} k \stepsize \Big) \mbasis^{2 \omega} & k \geq T_0 / \stepsize.
  \end{dcases}
\end{align*}
Combining the bounds, for stepsize satisfying $\stepsize \leq \mbasis^{- 2 \omega}$, we have the bound
\begin{align*}
  &\sum_{i = 0}^{+ \infty} \big(  \opnorm{H_1^{-1/2} \Exs [Y_0 H_1^{-1} Y_k^\top ] H_1^{-1/2}} +  \opnorm{H_1^{-1/2} \Exs [Y_0^\top H_1^{-1} Y_k] H_1^{-1/2}}  \big) \\
  &\leq  2 c_0 \numerOrder^2 \tau^4 \mbasis^{1 + 2 \omega} + 2 c' \stepsize T_0 \mbasis^{1 + 4 \omega} + \frac{2 T_0 \constreg (p, T_0)}{\stepsize} \mbasis \log (T_0 / \stepsize) + \frac{4 c_0}{\stepsize \lammin \poincare} e^{- \lammin \poincare T_0 / 4} \mbasis^{1 + 2 \omega}\\
  &\leq \frac{\mbasis}{\stepsize} \tau^4 \Big\{ \constScaryReg (T_0) \log ( 1/ \stepsize) + \frac{\constScary}{\poincare} e^{- \lammin \poincare T_0 / 4} \mbasis^{2 \omega} \Big\}.
\end{align*}
Now we apply \Cref{lemma:matrix-conc-markov} with the variance bounds and the almost-sure bounds, with probability $1- \delta$, we have
\begin{multline*}
  \opnorm{H_1^{-1/2} \Big( \frac{\numerOrder - 1}{N} I_1^{(j)} - \Abar \Big) H_1^{-1/2} }\\
   \leq  \tau^2 \sqrt{\frac{\mbasis}{T} \log (\mbasis / \delta)} \Big\{ \constScaryReg (T_0) \log ( 1/ \stepsize) + \frac{\constScary}{\poincare} e^{- \lammin \poincare T_0 / 4} \mbasis^{2 \omega} \Big\}^{1/2} + \frac{c D_\mbasis^2}{\poincare T} \log (\mbasis / \delta),
\end{multline*}
where the constant $c$ depends on $\lammin$ and $\numerOrder$. Taking union bound over $j \in 0,1,\cdots, \numerOrder - 2$, we conclude that
\begin{multline}
  \opnorm{H_1^{- 1/2} (N^{-1} I_1 - \Abar) H_1^{-1/2}} \\
  \leq \tau^2 \sqrt{\frac{\mbasis}{T} \log (\mbasis / \delta)} \Big\{ \constScaryReg (T_0) \log ( 1/ \stepsize) + \frac{\constScary}{\poincare} e^{- \lammin \poincare T_0 / 4} \mbasis^{2 \omega} \Big\}^{1/2} + \frac{c D_\mbasis^2}{\poincare T} \log (\mbasis / \delta),\label{eq:matrix-conc-markov-final-bound}
\end{multline}
with probability $1 - \delta$.

\paragraph{Concentration bounds for the term $I_2$:} Similar to the case of $I_1$, we start by decomposing the summation into $(\numerOrder - 1)$ terms. For $n \geq 0$ and $j \in \{ 0,1,\cdots \numerOrder - 2\}$, we define
\begin{align*}
  I_2^{(j)} (n) \mydefn \stepsize^{-1} \sum_{\substack{k \in [0, n - 1] \\ k \equiv j \mod (\numerOrder - 1)} }\psi (\MyState_{k \stepsize})  \Big\{ \int_0^{(\numerOrder - 1) \stepsize} e^{- \discount t} \nabla \psi (\MyState_{k \stepsize + t}) \covMat^{1/2} (\MyState_{k \stepsize + t}) dB_{k \stepsize + t} \Big\}^\top.
\end{align*}
For each index $j = 0,1,2\cdots, \numerOrder - 2$, the process $\big( I_2^{(j)} (n) \big)_{n \geq 0}$ is a discrete-time martingale, while each term is a continuous-time martingale.

To bound the concentration of such matrix martingales, we use the following matrix Freedman inequality from~\cite{tropp2011freedman}.
\begin{proposition}[\cite{tropp2011freedman}, Corollary 1.3]\label{prop:matrix-freedman}
  Let $(Y_k)_{k \geq 0}$ be a discrete-time matrix martingale in $\real^{m_1 \times m_2}$ adapted to the filtration $(\filtration_j)_{j \geq 0}$, and $X_k = Y_k - Y_{k- 1}$ for $k \geq 1$. Assume that $\opnorm{X_k} \leq R$ for $k = 1,2,\cdots$, and define the predictable quadratic variation processes
  \begin{align*}
    W_k \mydefn \sum_{j = 1}^k \Exs \big[ X_j X_j^\top \mid \filtration_{j - 1} \big], \quad \mbox{and} \quad \widetilde{W}_k \mydefn \sum_{j = 1}^k \Exs \big[ X_j^\top X_j \mid \filtration_{j - 1} \big].
  \end{align*}
  Then for any $t \geq 0$ and $\sigma > 0$, we have
  \begin{align*}
    \Prob \Big( \exists k \geq 0 ~:~ \opnorm{Y_k} \geq t, \quad \mbox{and} \quad \max \big( \opnorm{W_k}, \opnorm{\widetilde{W}_k} \big) \leq \sigma^2 \Big) \leq  (m_1 + m_2) \exp \Big( \frac{- t^2 / 2}{\sigma^2 + Rt /3}\Big).
  \end{align*}
\end{proposition}
Now let us bound the martingale $H_1^{-1/2} I_2^{(j)} H_1^{-1/2}$ using this concentration result. Given $j \in \{0,1,2\cdots, \numerOrder - 2\}$ fixed, letting $\ell (k) \mydefn (\numerOrder - 1) k + j$, we consider the discrete-time quadratic variation processes
\begin{align*}
  W_n &\mydefn H_1^{-1/2} \sum_{k = 0}^{n - 1}\psi (\MyState_{\ell (k) \stepsize})  \Exs \Big[ \vecnorm{ \int_0^{(\numerOrder - 1) \stepsize} e^{- \discount t} \nabla \psi (\MyState_{\ell (k) \stepsize + t}) \covMat^{1/2} (\MyState_{\ell (k) \stepsize + t}) dB_{\ell (k) \stepsize + t} }{H_1^{-1}}^2  \mid \MyState_{\ell (k) \stepsize}  \Big]\\
  &\qquad \qquad \qquad \cdot \psi (\MyState_{\ell (k) \stepsize})^\top H_1^{-1/2}\\
  &= \sum_{j = 0}^{n - 1} H_1^{-1/2} \psi (\MyState_{\ell (k) \stepsize}) \psi (\MyState_{\ell (k) \stepsize})^\top H_1^{-1/2} \cdot \int_0^{(\numerOrder - 1) \stepsize} e^{- 2 \discount t} \Exs \big[ \vecnorm{  \nabla \psi (\MyState_{\ell (k) \stepsize + t}) \covMat^{1/2} (\MyState_{\ell (k) \stepsize + t})}{H_1^{-1}}^2 \mid \State_{\ell (k) \stepsize} \big] dt\\
  &\preceq \lammax \sum_{k = 0}^{n - 1} H_1^{-1/2} \psi (\MyState_{\ell (k) \stepsize}) \psi (\MyState_{\ell (k) \stepsize})^\top H_1^{-1/2} \cdot \int_0^{(\numerOrder - 1) \stepsize} \semigroup_t \Big( \mathrm{Tr} \big( \nabla \psi H_1^{-1} \nabla \psi \big) \Big) (\MyState_{\ell (k) \stepsize}) dt,
\end{align*}
as well as
\begin{align*}
  \widetilde{W}_n &\mydefn \sum_{k = 0}^{n - 1} \vecnorm{\psi (\MyState_{\ell (k) \stepsize})}{H_1^{-1}}^2  H_1^{-1/2} \Exs \Big[ \Big\{ \int_0^{(\numerOrder - 1) \stepsize} e^{- \discount t} \nabla \psi (\MyState_{\ell (k) \stepsize + t}) \covMat^{1/2} (\MyState_{\ell (k) \stepsize + t}) dB_{\ell (k) \stepsize + t}  \Big\}\\
    &\qquad \qquad \qquad \qquad \qquad \cdot \Big\{ \int_0^{(\numerOrder - 1) \stepsize} e^{- \discount t} \nabla \psi (\MyState_{\ell (k) \stepsize + t}) \covMat^{1/2} (\MyState_{\ell (k) \stepsize + t}) dB_{\ell (k) \stepsize + t}  \Big\}^\top \mid \MyState_{\ell (k) \stepsize} \Big] H_1^{-1/2}\\
    &=  \sum_{k = 0}^{n - 1} \vecnorm{\psi (\MyState_{\ell (k) \stepsize})}{H_1^{-1}}^2  H_1^{-1/2} \int_0^{(\numerOrder - 1) \stepsize} e^{- 2 \discount t}  \Exs \Big[ \nabla \psi (\MyState_{\ell (k) \stepsize + t})  \covMat (\MyState_{\ell (k) \stepsize + t}) \nabla \psi (\MyState_{\ell (k) \stepsize + t})^\top \mid \MyState_{\ell (k) \stepsize} \Big] H_1^{-1/2} dt  \\
    &\preceq \lammax \sum_{k = 0}^{n - 1} \vecnorm{\psi (\MyState_{\ell (k) \stepsize})}{H_1^{-1}}^2 \int_0^{(\numerOrder - 1) \stepsize} H_1^{-1/2}  \Exs \Big[ \nabla \psi (\MyState_{\ell (k) \stepsize + t}) \nabla \psi (\MyState_{\ell (k) \stepsize + t})^\top \mid \MyState_{\ell (k) \stepsize} \Big] H_1^{-1/2}  dt.
\end{align*}

The following lemma establishes uniform upper bounds on the operator norms of matrix-valued processes $(W_n)_{n \geq 0}$ and $(\widetilde{W}_n)_{n \geq 0}$.
\begin{lemma}\label{lemma:variance-matrix-conc-in-martingale-proof}
  Under the setup of \Cref{thm:main}, given a trajectory length $T \geq \frac{c \mbasis }{\lammin \poincare} \log (\mbasis N / \delta)$, with probability $1 - \delta$, we have
  \begin{align}
    \max_{0 \leq n \leq N - 1} \max \big( \opnorm{W_n}, \opnorm{\widetilde{W}_n} \big) \leq 2 \tau^4 \numerOrder \mbasis T + c \numerOrder \stepsize D_\mbasis^4 \log (\mbasis N / \delta).\label{eq:lemma-variance-matrix-conc-in-martingale-proof}
  \end{align}
\end{lemma}
\noindent See \Cref{subsec:proof-lemma-variance-matrix-conc-in-martingale-proof} for the proof of this lemma. Taking this lemma as given, let us now proceed to bound the quantity $I_2^{(j)}$.

The terms in the martingale $I_2^{(j)} (n)$ are not uniformly bounded, however, we could still establish high-probability bounds and apply a truncation argument.

Applying the vector-valued Burkholder--Davis--Gundy inequality~\cite{marinelli2016maximal} conditionally on $\MyState_{\ell (k) \stepsize}$, for every $p \geq 2$, we have
\begin{align*}
  &\Exs \Big[ \sup_{0 \leq t \leq (\numerOrder - 1) \stepsize} \vecnorm{\int_0^t e^{- \discount s} H_1^{-1/2} \nabla \psi (\MyState_{\ell (k) \stepsize + s}) \covMat^{1/2} (\MyState_{\ell (k) \stepsize + s}) dB_{\ell (k) \stepsize + s} }{2}^p \mid \State_{\ell (k) \stepsize} \Big] \\
  &\leq (cp \lammax)^{p/2} \Exs \Big[ \Big\{ \int_0^{(\numerOrder- 1) \stepsize} \mathrm{Tr} \big( \nabla \psi^\top H_1^{-1} \nabla \psi \big) (\State_{\ell (k) \stepsize + t}) dt \Big\}^{p/2} \mid \State_{\ell (k) \stepsize} \Big].
\end{align*}
For fixed $\delta \in (0, 1)$, we take $p = \log (1 / \delta)$ and apply Markov inequality. Conditionally on $X_{\ell (k) \stepsize}$, with probability $1 - \delta$, we have 
\begin{align*}
  &\sup_{0 \leq t \leq (\numerOrder - 1) \stepsize} \vecnorm{\int_0^t e^{- \discount s} H_1^{-1/2} \nabla \psi (\MyState_{\ell (k) \stepsize + s}) \covMat^{1/2} (\MyState_{\ell (k) \stepsize + s}) dB_{\ell (k) \stepsize + s} }{2} \\
  &\leq \Big\{ \int_0^{(\numerOrder- 1) \stepsize} \sup_{y \in \real^\usedim} \mathrm{Tr} \big( \nabla \psi^\top H_1^{-1} \nabla \psi \big) (y) dt \Big\}^{1/2} \cdot \sqrt{c \lammax \log (1 / \delta)}\\
  &\leq \sqrt{c \lammax \numerOrder \stepsize \log (1 / \delta) \cdot \sup_{y \in \real^\usedim} \mathrm{Tr} \big( \nabla \psi^\top H_1^{-1} \nabla \psi \big) (y) }.
\end{align*}
Therefore, with probability $1 - \delta$, we have
\begin{align*}
  &\opnorm{H_1^{-1/2} \psi (\State_{k (\ell) \stepsize})  \int_0^{(\numerOrder - 1) \stepsize} e^{- \discount t} \nabla \psi (\MyState_{\ell (k) \stepsize + t}) \covMat^{1/2} (\MyState_{\ell (k) \stepsize + t}) dB_{\ell (k) \stepsize + t}^\top H_1^{-1/2}} \\
  &\leq \sqrt{c \lammax \numerOrder \stepsize \log (1 / \delta)} \cdot \sup_{ \in \real^\usedim} \vecnorm{H_1^{-1/2} \psi (y)}{2} \cdot \sup_{y \in \real^\usedim} \matsnorm{H_1^{-1/2} \nabla \psi (y)}{F}\\
  &\leq D_\mbasis^2 \sqrt{c \lammax \numerOrder \stepsize \log (1 / \delta)},
\end{align*}
where the last inequality is due to Assumption~\ref{assume:bounded-feature}.

Applying union bound over $\ell = 1,2,\cdots, N / (\numerOrder - 1)$, we conclude that
\begin{multline}
  \max_{1 \leq \ell \leq N / (\numerOrder - 1)} \opnorm{H_1^{-1/2} \psi (\State_{k (\ell) \stepsize}) \Big\{ \int_0^{(\numerOrder - 1) \stepsize} e^{- \discount t} \nabla \psi (\MyState_{\ell (k) \stepsize + t}) \covMat^{1/2} (\MyState_{\ell (k) \stepsize + t}) dB_{\ell (k) \stepsize + t} \Big\}^\top H_1^{-1/2}} \\
  \leq D_\mbasis^2 \sqrt{c \lammax \numerOrder \stepsize \log \big(\tfrac{T}{\stepsize \delta} \big)},\label{eq:high-prob-bound-for-each-term-in-matrix-martingale-concentration}
\end{multline}
with probability $1 - \delta$.

Now we apply \Cref{prop:matrix-freedman} with the variance bound and and high-probability bound derived above. Define the event
\begin{align*}
  \Event \mydefn \Big\{ \mbox{\Cref{eq:lemma-variance-matrix-conc-in-martingale-proof,eq:high-prob-bound-for-each-term-in-matrix-martingale-concentration} hold true} \Big\}.
\end{align*}
We have shown that $\Prob (\Event) \geq 1 - 2 \delta$.

By \Cref{prop:matrix-freedman}, we have
\begin{align*}
  \Prob \Big( \Big\{\opnorm{H_1^{-1/2} \frac{I_2^{(j)} (N)}{N} H_1^{-1/2}} \geq 4 \tau^2 \sqrt{\frac{ \numerOrder \mbasis}{T} \log (\mbasis / \delta)} + c' \frac{D_m^2}{T} \log (\mbasis/ \delta) \sqrt{\stepsize \log \big(\tfrac{T}{\stepsize \delta} \big)} \Big\} \cap \Event \Big) \leq \delta.
\end{align*}
Now we take union bound over the $(\numerOrder - 1)$ indices $j \in \{0,1,\cdots, \numerOrder - 1\}$, by replacing $\delta$ with $\tfrac{\delta}{3 (\numerOrder - 1)}$, we conclude that
\begin{align}
\opnorm{H_1^{-1/2} \frac{I_2 (N)}{N} H_1^{-1/2}} \geq c \tau^2 \sqrt{\frac{ \mbasis}{T} \log (\mbasis / \delta)} + c' \frac{D_m^2}{T} \sqrt{\stepsize \log^3 \big(\tfrac{T}{\stepsize \delta} \big)},\label{eq:i2-martingale-bound-in-matrix-concentration}
\end{align}
with probability $1 - \delta$.

\paragraph{Putting them together:} Combining \Cref{eq:matrix-conc-markov-final-bound,eq:i2-martingale-bound-in-matrix-concentration}, with probability $1 - \delta$, we have
\begin{align*}
  &\opnorm{H_1^{-1/2} \frac{1}{N} \Big\{\sum_{k = 0}^{N - 1} A_k - \Abar \Big\} H^{-1/2}}\\
  &\leq \opnorm{H_1^{-1/2}  \Big\{\frac{1}{N} I_1 - \Abar \Big\} H^{-1/2}} + \opnorm{H_1^{-1/2} I_2 H^{-1/2}}\\
  &\leq  \tau^2 \sqrt{\frac{\mbasis}{T} \log (\mbasis / \delta)} \Big\{ 1 + \constScaryReg (T_0) \log ( 1/ \stepsize) + \frac{\constScary}{\poincare} e^{- \lammin \poincare T_0 / 4} \mbasis^{2 \omega} \Big\}^{1/2} + \frac{c D_\mbasis^2}{\poincare T} \log^{3/2} \big( \frac{T \mbasis}{\stepsize \delta} \big).
\end{align*}
Combining with \Cref{eq:matrix-concentration-op-norm-conversion}, we complete the proof of \Cref{lemma:matrix-conc-for-bellman-eq}.

\subsubsection{Proof of Lemma~\ref{lemma:main-err-term-in-sample-based-bellman}}\label{subsubsec:proof-lemma-main-err-term-in-sample-based-bellman}
Define the interpolated rewards
\begin{align*}
  \widebar{\Reward}_{ k (\numerOrder - 1) \stepsize + t} \mydefn \sum_{i = 0}^{\numerOrder - 1} W_i (t) \Reward_{ k (\numerOrder - 1) \stepsize + i \stepsize}, \quad \mbox{and} \quad 
  \widebar{\reward}_{ k (\numerOrder - 1) \stepsize + t} \mydefn \sum_{i = 0}^{\numerOrder - 1} W_i (t) \reward (\MyState_{ k (\numerOrder - 1) \stepsize + i \stepsize}),
\end{align*}
for $t \in [0, (\numerOrder - 1) \stepsize]$ and $k = 0,1,2,\cdots$.

By It\^{o}'s formula, we have
\begin{multline*}
  A_k \thetabar - b_k = \psi (X_{k \stepsize}) \cdot \stepsize^{-1} \int_0^{(\numerOrder - 1) \stepsize} e^{- \discount t} \Big\{\discount \psi (\MyState_{k \stepsize + t})^\top \thetabar - \generator \psi (\MyState_{k \stepsize + t})^\top \thetabar - \widebar{\Reward}_t \Big\} dt\\
 + \psi (\MyState_{k \stepsize}) \cdot \stepsize^{-1} \int_0^{(\numerOrder - 1) \stepsize} e^{- \discount t} \thetabar^\top \nabla \psi (\MyState_{k \stepsize + t}) \covMat^{1/2} (\MyState_{k \stepsize + t}) dB_{k \stepsize + t}.
\end{multline*}
We can then decompose the summation into 3 parts.
\begin{align*}
  &\frac{1}{N}\sum_{k = 0}^{N - 1} H_1^{-1/2} \big( A_k \thetabar - b_k + \varepsilon_k \big)\\
  &= \frac{1}{N \stepsize} \sum_{k = 0}^{N - 1} H_1^{-1/2} \psi (X_{k \stepsize}) \int_0^{(\numerOrder - 1) \stepsize} e^{- \discount t} \big( \discount \valuebar (\MyState_{k \stepsize + t}) - \generator \valuebar (\MyState_{k \stepsize + t}) - \bar{\reward}_t \big)  dt \\
  & \qquad +  \frac{1}{N \stepsize} \sum_{k = 0}^{N - 1} H_1^{-1/2} \psi (X_{k \stepsize})  \int_0^{(\numerOrder - 1) \stepsize} e^{- \discount t} \thetabar^\top \nabla \psi (\MyState_{k \stepsize + t}) \covMat^{1/2} (\MyState_{k \stepsize + t}) dB_{k \stepsize + t} \\ 
  &\qquad \qquad -  \frac{1}{N} \sum_{k = 0}^{N - 1} H_1^{-1/2}  \varepsilon_k  \\
  &=:  J_1 (N) + J_2 (N) + J_3 (N).
\end{align*}

\paragraph{Upper bound for $J_1 (N)$:}
To start with, we define the auxiliary vector
\begin{align*}
  \thetatil \mydefn \Abar^{-1}  \frac{1}{\stepsize} \int_0^{(\numerOrder - 1) \stepsize} e^{- \discount t} \Exs \big[ \reward (\MyState_t) \psi (\MyState_0) \big] dt.
\end{align*}
Defining the interpolated reward function $\widebar{\reward}_t = \sum_{i = 0}^{\numerOrder - 1} W_i (t) \reward (\MyState_{\stepsize i})$, for the functions $\fbar \mydefn \inprod{\thetabar}{\psi}$ and $\ftil \mydefn \inprod{\thetatil}{\psi}$, we have
\begin{align*}
 & \Exs \Big[ g (\MyState_0) \int_0^{(\numerOrder - 1) \stepsize} e^{- \discount t} \big( \discount \fbar (\MyState_t) - \generator \fbar (\MyState_t) - \widebar{\reward}_t \big) dt \Big] = 0,\quad \mbox{and}\\
  &\Exs \Big[ g (\MyState_0) \int_0^{(\numerOrder - 1) \stepsize}  e^{- \discount t}  \big( \discount \ftil (\MyState_t) - \generator \ftil (\MyState_t) - \reward (\MyState_t) \big) dt \Big] = 0,
\end{align*}
for any function $g \in \LinSpace$.

We define the operator defect errors
\begin{align*}
  \widebar{\zeta}_k &\mydefn \frac{1}{\stepsize} \int_0^{(\numerOrder - 1) \stepsize} e^{- \discount t} \big( \discount \valuebar (\MyState_{k \stepsize + t}) - \generator \valuebar (\MyState_{k \stepsize + t}) - \bar{\reward}_{k \stepsize + t} \big)  dt,\\
  \widetilde{\zeta}_k &\mydefn \frac{1}{\stepsize} \int_0^{(\numerOrder - 1) \stepsize} e^{- \discount t} \big( \discount \ftil (\MyState_{k \stepsize + t}) - \generator \ftil (\MyState_{k \stepsize + t}) - \reward (\State_{k \stepsize + t}) \big)  dt.
\end{align*}
Since the true value function $\ValTrue$ satisfies the elliptic equation $(\discount - \generator) \ValTrue = \reward$. We have
\begin{align*}
  \widetilde{\zeta}_k = \frac{1}{\stepsize} \int_0^{(\numerOrder - 1) \stepsize} e^{- \discount t} \big(\discount - \generator \big) (\ftil - \ValTrue) (\MyState_{k \stepsize + t}) dt.
\end{align*}

 Similar to the proof of \Cref{lemma:matrix-conc-for-bellman-eq}, we decompose the summation into integrals with non-overlapping blocks.
\begin{align*}
  J_1^{(j)} (N) \mydefn \frac{1}{N} \sum_{\substack{ 0 \leq k \leq N - 1 \\ k \equiv j \mod (\numerOrder - 1) }} H_1^{-1/2} \psi (X_{k \stepsize}) \widebar{\zeta}_k,
\end{align*}
for $j = 0,1,2,\cdots, \numerOrder - 1$.

Defining $\ell (k) \mydefn k (\numerOrder - 1)$, by Young's inequality, we have
\begin{align}
  &\Exs \big[ \vecnorm{J_1^{(j)} (N)}{2}^2 \big] \nonumber \\
  &\leq 2 \Exs \big[ \vecnorm{ \frac{1}{N} \sum_{k = 0 }^{N / (\numerOrder - 1)} H_1^{-1/2} \psi (X_{{\ell (k)} \stepsize}) \widetilde{\zeta}_{\ell (k)}}{2}^2 \big] + 2 \Exs \big[ \vecnorm{ \frac{1}{N} \sum_{k = 0 }^{N / (\numerOrder - 1)} H_1^{-1/2} \psi (X_{{\ell (k)} \stepsize}) \big( \widetilde{\zeta}_{\ell (k)} - \widebar{\zeta}_{\ell (k)} \big)}{2}^2 \big]. \label{eq:decomposition-of-j1-in-main-variance-bound}
\end{align}
Now we bound the two terms separately. For the first term, we note that
\begin{align*}
  &\Exs \Big[ \vecnorm{ \frac{1}{N} \sum_{k = 0 }^{N / (\numerOrder - 1)} H_1^{-1/2} \psi (\State_{\ell(k) \stepsize}) \widetilde{\zeta}_{\ell (k)}}{2}^2 \Big]  \\
  &\leq \frac{1}{N^2} \sum_{k = 0}^{N / (\numerOrder - 1)} (N / (\numerOrder - 1) - k) \abss{\Exs \Big[ \psi (\State_{\ell (0) \stepsize})^\top H_1^{-1} \psi (\State_{\ell (k) \stepsize}) \widetilde{\zeta}_{\ell (0)}  \widetilde{\zeta}_{\ell (k)} \Big]}\\
  &\leq \frac{1}{N} \sum_{j = 1}^\mbasis \sum_{k = 0}^{+ \infty} \abss{\mu_k \big( \coordinate_j^\top H_1^{-1/2} \psi, \ftil - \ValTrue  \big)},
\end{align*}
where we slightly abuse the definition~\eqref{eq:defn-muk-cross-var} in $\mu_k$ by replacing the step length $\stepsize$ with $(\numerOrder - 1) \stepsize$.

In order to bound such a term, we invoke \Cref{lemma:sigma-mkv-global-bound}. For each $j \in [\usedim]$, we have
\begin{align*}
  &\stepsize (\numerOrder - 1) \sum_{k = 0}^{+ \infty} \abss{\mu_k \big( \coordinate_j^\top H_1^{-1/2} \psi, \ftil - \ValTrue  \big)} \\
  &\leq \tau^2 \sobonorm{\coordinate_j^\top H_1^{-1/2} \psi}^2 \cdot \Big\{ \constScaryReg (T_0) \sobopstatnorm{\ftil - \ValTrue}{2p}^2 \log (1 / \stepsize) + \constScary \highorder_{T_0, \stepsize} (\ftil - \ValTrue) \Big\}.
\end{align*}
Note that $\sobonorm{\coordinate_j^\top H_1^{-1/2} \psi}^2 = \coordinate_j^\top H_1^{-1/2} H_1 H_1^{-1/2} \coordinate_j = 1$. Substituting to the variance bound, we have
\begin{multline}
  \Exs \Big[ \vecnorm{ \frac{1}{N} \sum_{k = 0 }^{N / (\numerOrder - 1)} H_1^{-1/2} \psi (\State_{\ell(k) \stepsize}) \widetilde{\zeta}_{\ell (k)}}{2}^2 \Big] \\ 
  \leq \frac{\tau^2 \mbasis}{T} \Big\{  \constScaryReg (T_0) \sobopstatnorm{\ftil - \ValTrue}{2p}^2 \log (1 / \stepsize) + \constScary \highorder_{T_0, \stepsize} (\ftil - \ValTrue)\Big\}.\label{eq:main-var-main-part-bound}
\end{multline}
We use the following lemma to bound the second term in the decomposition~\eqref{eq:decomposition-of-j1-in-main-variance-bound}.
\begin{lemma}\label{lemma:variance-additional-term-bound}
  Under the setup of \Cref{lemma:main-err-term-in-sample-based-bellman}, we have
  \begin{multline*}
    \Exs \big[ \vecnorm{ \frac{1}{N} \sum_{k = 0 }^{N / (\numerOrder - 1)} H_1^{-1/2} \psi (X_{{\ell (k)} \stepsize}) \big( \widetilde{\zeta}_{\ell (k)} - \widebar{\zeta}_{\ell (k)} \big)}{2}^2 \big] \\
    \leq \frac{1}{T} \Big\{ \sobonorm{\ftil - \fbar}^2 \Tthres (\mbasis, T_0) + \tau^4 \constScary \mathrm{Tr} \big( H_1^{-1} H_0 \big) ( \stepsize^2 +  \poincare^{-1} \stepsize^{1 + \numerOrder}) \Big\}.
  \end{multline*}
\end{lemma}
\noindent See \Cref{subsec:app-proof-of-lemma-variance-additional-term-bound} for the proof of this lemma.

Since the bounds in \Cref{eq:main-var-main-part-bound,lemma:variance-additional-term-bound} rely on the approximation errors associated to the projected fixed-points. We use the following lemma to control such errors.

\begin{lemma}\label{lemma:relate-approx-error-to-best-err}
  Under the setup of \Cref{thm:main}, for $q \in (2, \tfrac{4p}{p - 1}]$, we have
  \begin{align*}
    \sobopstatnorm{\ftil - \ValTrue}{q} &\leq  \sobopstatnorm{\Delstar}{q} + c \tau \sobonorm{\Delstar}, \\
    \sobokpstatnorm{\ftil - \ValTrue}{2}{q} & \leq c \tau \mbasis^\omega \sobonorm{\Delstar} + \sobokpstatnorm{\Delstar}{2}{q},\\
    \sobokpstatnorm{\valuebar - \ftil}{1}{q} &\leq \constScary \tau \stepsize^\numerOrder,
  \end{align*}
  where the constant $c > 0$ depends on the parameters in Assumptions~\ref{assume:uniform-elliptic}, \ref{assume:smooth-stationary}, and \fakerefassumelip{$(2)$}, and the constant $\constScary > 0$ depends on the parameters in \fakerefassumelip{$(\numerOrder + 1)$}.
\end{lemma}
\noindent See~\Cref{subsec:app-proof-of-lemma-relate-approx-error-to-best-err} for the proof of this lemma.

Applying \Cref{lemma:relate-approx-error-to-best-err} to the bounds in \Cref{eq:main-var-main-part-bound,lemma:variance-additional-term-bound}, we conclude that
\begin{align*}
  &\Exs \big[ \vecnorm{ \frac{1}{N} \sum_{k = 0 }^{N / (\numerOrder - 1)} H_1^{-1/2} \psi (X_{{\ell (k)} \stepsize}) \big( \widetilde{\zeta}_{\ell (k)} - \widebar{\zeta}_{\ell (k)} \big)}{2}^2 \big] \\
    &\leq \frac{1}{T} \Big\{ \stepsize^{2 \numerOrder} \Tthres (\mbasis, T_0) + \tau^4 \constScary \mathrm{Tr} \big( H_1^{-1} H_0 \big) ( \stepsize^2 +  \poincare^{-1} \stepsize^{1 + \numerOrder}) \Big\}.
\end{align*}
as well as
\begin{align*}
  &\Exs \Big[ \vecnorm{ \frac{1}{N} \sum_{k = 0 }^{N / (\numerOrder - 1)} H_1^{-1/2} \psi (\State_{\ell(k) \stepsize}) \widetilde{\zeta}_{\ell (k)}}{2}^2 \Big] \\
  &\leq \frac{\tau^2 \mbasis}{T} \Big\{  \constScaryReg (T_0) \sobopstatnorm{\Delstar}{2p}^2 \log (1 / \stepsize) + \constScary \highorder_{T_0, \stepsize} (\Delstar) \Big\} + \frac{\Tthres (\mbasis, T_0)}{T} \sobonorm{\Delstar}^2.
\end{align*}
Putting them together and combining the bounds for $j = 0,1,\cdots, \numerOrder - 1$, we conclude that
\begin{multline}
  \Exs [ \vecnorm{J_1 (N)}{2}^2 ] \leq \frac{\tau^2 \mbasis}{T} \Big\{ \constScaryReg (T_0) \sobopstatnorm{\Delstar}{2p}^2 \log (1 / \stepsize) + \constScary \highorder_{T_0, \stepsize} (\Delstar) \Big\}\\
  +  \frac{\Tthres (\mbasis, T_0)}{T} \big( \sobonorm{\Delstar}^2 + \stepsize^{2 \numerOrder} \big) + \frac{\tau^4 \constScary }{T}\mathrm{Tr} \big( H_1^{-1} H_0 \big) ( \stepsize^2 +  \poincare^{-1} \stepsize^{1 + \numerOrder}).\label{eq:j1-final-bound-in-main-err-lemma}
\end{multline}

\paragraph{Upper bound for $J_2 (N)$:} For $\ell \in \{0,1,\cdots ,\numerOrder - 2\}$, we define
\begin{align*}
  J_2^{(\ell)} (N) \mydefn \frac{1}{N \stepsize}\sum_{\substack{0 \leq k \leq N - 1\\ k \mod (\numerOrder - 1) = \ell}} H_1^{-1/2} \psi (X_{k \stepsize})  \int_0^{(\numerOrder - 1) \stepsize} e^{- \discount t} \thetabar^\top \nabla \psi (\MyState_{k \stepsize + t}) \covMat^{1/2} (\MyState_{k \stepsize + t}) dB_{k \stepsize + t}
\end{align*}
For each $\ell \in \{0, 1, \cdots, \numerOrder - 2\}$, the stochastic integrals in each terms of $J_3^{(\ell)} (N)$ do not overlap with each other. By It\^{o} isometry, we have
\begin{align*}
  \Exs \big[ \vecnorm{J_2^{(\ell)} (N)}{2}^2 \big] &= \frac{1}{(N \stepsize)^2} \sum_{\substack{0 \leq k \leq N - 1\\ k \mod (\numerOrder - 1) = \ell}} \Exs \Big[ \vecnorm{ H_1^{-1/2} \psi (X_{k \stepsize})}{2}^2 \int_0^{(\numerOrder - 1) \stepsize} e^{- 2 \discount t} \nabla \fbar (\MyState_{k \stepsize + t})^\top \covMat (\MyState_{k \stepsize + t})\nabla \fbar (\MyState_{k \stepsize + t}) dt \Big]\\
  &\leq \frac{\lammax}{(N \stepsize)^2} \sum_{\substack{0 \leq k \leq N - 1\\ k \mod (\numerOrder - 1) = \ell}} \int_0^{(\numerOrder - 1) \stepsize}  \Exs \big[ \vecnorm{ H_1^{-1/2} \psi (X_{k \stepsize}) }{2}^2 \cdot \vecnorm{ \nabla \fbar (\MyState_{k \stepsize + t}) }{2}^2 \big] dt\\
  &\leq \sum_{j = 1}^\mbasis \frac{\lammax}{(N \stepsize)^2} \cdot \frac{N}{\numerOrder - 1} \int_0^{(\numerOrder - 1) \stepsize}  \Exs \big[ (\coordinate_j H_1^{-1/2} \psi) (X_0)^2 \cdot \vecnorm{ \nabla \fbar (\MyState_{t}) }{2}^2 \big] dt\\
  &\leq  \sum_{j = 1}^\mbasis \frac{\lammax}{N \stepsize} \cdot  \sqrt{\Exs \big[ (\coordinate_j H_1^{-1/2} \psi) (X_0)^4]} \cdot \sqrt{ \Exs \big[ \vecnorm{ \nabla \fbar (\MyState_{t}) }{2}^4 \big] }.
\end{align*}
By the hyper-contractivity assumption~\fakerefassumehyper{4}{$\tau_4$}, we have
\begin{align*}
   \sqrt{\Exs \big[ (\coordinate_j H_1^{-1/2} \psi) (X_0)^4]} \leq \tau_4^2 \Exs \big[ (\coordinate_j H_1^{-1/2} \psi) (X_0)^2], \quad \mbox{and} \quad  \sqrt{\Exs \big[ \vecnorm{ \nabla \fbar (\MyState_{t}) }{2}^4 \big]} \leq \tau_4^2 \sobonorm{\fbar}^2.
\end{align*}
Substituting back to the bound above, we conclude that
\begin{align*}
  \Exs \big[ \vecnorm{J_2^{(\ell)} (N)}{2}^2 \big] \leq \frac{\tau_4^4 \lammax}{T}  \sobonorm{\fbar}^2 \sum_{j = 1}^\mbasis \Exs \big[ (\coordinate_j H_1^{-1/2} \psi) (X_0)^2] = \frac{\tau_4^4 \lammax}{T} \sobonorm{\fbar}^2 \mathrm{Tr} \big( H_1^{-1} H_0 \big).
\end{align*}
Consequently, we have
\begin{align}
  \Exs \big[ \vecnorm{J_2 (N)}{2}^2 \big] \leq  \frac{\tau_4^4 \numerOrder \lammax}{T} \sobonorm{\fbar}^2 \mathrm{Tr} \big( H_1^{-1} H_0 \big).\label{eq:j2-final-bound-in-main-err-lemma}
\end{align}

\paragraph{Upper bound for $J_3 (N)$:} Similar to the bounds on $J_1$ and $J_2$, we decompose the summation into non-overlapping blocks.
\begin{align*}
  J_3^{(\ell)} (N) \mydefn \frac{1}{N} \sum_{\substack{0 \leq k \leq N - 1\\ k \mod (\numerOrder - 1) = \ell}}\sum_{i = 0}^{\numerOrder - 1} \kappa_i \Big\{ \Reward_{(k + i) \stepsize} -  \reward \big( \MyState_{(k + i) \stepsize} \big) \Big\}  H_1^{-1/2} \psi (\MyState_{k \stepsize})
\end{align*}
For each $\ell = 0,1,2,\cdots, \numerOrder - 1$, the term $J_3^{(\ell)}$ is a martingale. So we have
\begin{align*}
  \Exs \big[ \vecnorm{J_3^{(\ell)} (N)}{2}^2 \big] &= \frac{1}{N^2} \sum_{\substack{0 \leq k \leq N - 1\\ k \mod (\numerOrder - 1) = \ell}} \Exs \Big[ \vecnorm{ \sum_{i = 0}^{\numerOrder - 1} \kappa_i \Big\{ \Reward_{(k + i) \stepsize} -  \reward \big( \MyState_{(k + i) \stepsize} \big) \Big\}  H_1^{-1/2} \psi (\MyState_{k \stepsize})}{2}^2 \Big]\\
  & \leq \frac{\stepsize}{T} \Big( \sum_{i = 0}^{\numerOrder - 1} |\kappa_i| \Big)^2 \mathrm{Tr} \big( H_1^{-1} H_0 \big)\\
  &\leq c \frac{\stepsize}{T} \mathrm{Tr} \big( H_1^{-1} H_0 \big).
\end{align*}
Combining the bounds for $\ell = 0,1,2,\cdots, \numerOrder - 2$ using Cauchy--Schwarz inequality, we have
\begin{align}
  \Exs \big[ \vecnorm{J_3 (N)}{2}^2 \big] \leq c \numerOrder \frac{\stepsize}{T} \mathrm{Tr} \big( H_1^{-1} H_0 \big).\label{eq:j3-final-bound-in-main-err-lemma}
\end{align}

\paragraph{Putting them together:} Combining \Cref{eq:j1-final-bound-in-main-err-lemma,eq:j2-final-bound-in-main-err-lemma,eq:j3-final-bound-in-main-err-lemma}, we have
\begin{align*}
  &\Exs \big[ \vecnorm{ \frac{1}{N} \sum_{k = 0}^{N - 1} H_1^{-1/2} \big\{ A_k \thetabar - b_k + \varepsilon_k \big\}}{2}^2 \big]\\
  & \leq \frac{\tau^2 \mbasis}{T} \Big\{ \constScaryReg (T_0) \sobopstatnorm{\Delstar}{2p}^2 \log (1 / \stepsize) + \constScary \highorder_{T_0, \stepsize} (\Delstar) \Big\}\\
  &\qquad \qquad +  \frac{\Tthres (\mbasis, T_0)}{T} \big( \sobonorm{\Delstar}^2 + \stepsize^{2 \numerOrder} \big) \\
  &\qquad \qquad \qquad \qquad+ \frac{\tau^4 \constScary}{T} \mathrm{Tr} \big( H_1^{-1} H_0 \big) ( \stepsize^2 +  \poincare^{-1} \stepsize^{1 + \numerOrder} + \stepsize + \sobonorm{\fbar}^2).
\end{align*}
Given a stepsize satisfying $\stepsize^\numerOrder \leq \poincare$, we conclude the proof of this lemma.

\section{Discussion}\label{sec:discussion}
In this paper, we undertake a study on the statistical errors for policy evaluation in continuous-time Markovian diffusions, using single-trajectory data. We establish sharp non-asymptotic guarantees for the LSTD estimator accompanied with a high-order numerical discretization scheme. When measuring the error in first-order Sobolev norm, the statistical analysis is enabled by the underlying ellipticity structures of the diffusion process, achieving the optimal $O (1/T)$ convergence rate. By fine-grained analysis on the problem-dependent pre-factors in the convergence rate, we reveal a new phenomena: the Markovian part of the statistical error can be controlled using the approximation error itself, and the martingale part of the statistical error grows sub-linearly in the number of basis functions. This structural finding renders a non-standard trade-off between approximation and statistical error.

Moving forward, this paper opens up several important directions in statistical analysis of continuous-time RL algorithms.
\begin{itemize}
  \item On the technical side, there are several important open questions unresolved. First, an explicit bound on the high-order regularity estimate for log-density of the diffusion semigroup will allow us to obtain bounds in the form of \Cref{eq:simplified-bound-good-case}, making a clearer picture of the approximation-vs-statistical tradeoff. Moreover, it is interesting to investigate the information-theoretic optimality of the upper bounds in \Cref{thm:main}. In particular, it is interesting to see whether the best choice of $\mbasis$ that balance the three parts of the errors leads to an optimal non-parametric estimator, assuming that $\ValTrue$ lies in some smoothness classes.
  \item In general, it is interesting to extends our study to a broader class of RL algorithms for continuous-time dynamics, including $Q$-learning and actor-critic methods. Our analysis provides two key technical insights that are applicable in general: first, the elliptic structures play a key role in our statistical analysis. In particular, since the effective horizon diverges to infinity as stepsize decreases, most classical discrete-time RL theory becomes inapplicable. Second, through a new characterization of the asymptotic covariance structure for functionals of Markov chain, we discover a non-standard tradeoff in continuous-time RL. It is interesting to see whether the such a phenomenon arise from other RL algorithms.
\end{itemize}

\section*{Acknowledgement}
This work was partially supported by NSERC grant RGPIN-2024-05092 and a Connaught New Researcher Award to WM. WM would like to thank Yuhua Zhu for helpful discussion at early stage of this work, and thanks to Lei Li for pointing out connections to Malliavin calculus.

\bibliographystyle{alpha}
\bibliography{references}

\newcommand{\etalchar}[1]{$^{#1}$}
\begin{thebibliography}{MPWB24}

\bibitem[Ada08]{adamczak2008tail}
R.~Adamczak.
\newblock A tail inequality for suprema of unbounded empirical processes with applications to markov chains.
\newblock {\em Electronic Journal of Probability}, 13:1000--1034, 2008.

\bibitem[BB96]{bradtke1996linear}
S.~J. Bradtke and A.~G. Barto.
\newblock Linear least-squares algorithms for temporal difference learning.
\newblock {\em Machine Learning}, 22(1-3):33--57, 1996.

\bibitem[BBCG08]{bakry2008simple}
D.~Bakry, F.~Barthe, P.~Cattiaux, and A.~Guillin.
\newblock A simple proof of the {P}oincar{\'e} inequality for a large class of probability measures.
\newblock {\em Electronic Communications in Probability}, 13:60--66, 2008.

\bibitem[B{\'E}06]{bakry2006diffusions}
D.~Bakry and M.~{\'E}mery.
\newblock Diffusions hypercontractives.
\newblock In {\em S{\'e}minaire de Probabilit{\'e}s XIX 1983/84: Proceedings}, pages 177--206. Springer, 2006.

\bibitem[Boy02]{boyan2002technical}
J.~A. Boyan.
\newblock Technical update: Least-squares temporal difference learning.
\newblock {\em Machine learning}, 49(2-3):233--246, 2002.

\bibitem[Doy95]{doya1995temporal}
K.~Doya.
\newblock Temporal difference learning in continuous time and space.
\newblock {\em Advances in neural information processing systems}, 8, 1995.

\bibitem[Duo24]{duoandikoetxea2024fourier}
J.~Duoandikoetxea.
\newblock {\em Fourier analysis}, volume~29.
\newblock American Mathematical Society, 2024.

\bibitem[DW22]{duan2022policy}
Yaqi Duan and Martin~J Wainwright.
\newblock Policy evaluation from a single path: Multi-step methods, mixing and mis-specification.
\newblock {\em arXiv preprint arXiv:2211.03899}, 2022.

\bibitem[Gan81]{ganzburg1981multidimensional}
M.~I. Ganzburg.
\newblock Multidimensional jackson theorems.
\newblock {\em Siberian Mathematical Journal}, 22(2):223--231, 1981.

\bibitem[HRX24]{han2024stochastic}
Yinbin Han, Meisam Razaviyayn, and Renyuan Xu.
\newblock Stochastic control for fine-tuning diffusion models: Optimality, regularity, and convergence.
\newblock {\em arXiv preprint arXiv:2412.18164}, 2024.

\bibitem[Isi85]{isidori1985nonlinear}
A.~Isidori.
\newblock {\em Nonlinear control systems: an introduction}.
\newblock Springer, 1985.

\bibitem[JKK{\etalchar{+}}18]{jain2018accelerating}
P.~Jain, S.~M. Kakade, R.~Kidambi, P.~Netrapalli, and A.~Sidford.
\newblock Accelerating stochastic gradient descent for least squares regression.
\newblock In {\em Conference On Learning Theory}, pages 545--604. PMLR, 2018.

\bibitem[JZ22a]{jia2022policyeval}
Y.~Jia and X.~Y. Zhou.
\newblock Policy evaluation and temporal-difference learning in continuous time and space: A martingale approach.
\newblock {\em The Journal of Machine Learning Research}, 23(1):6918--6972, 2022.

\bibitem[JZ22b]{jia2022policygrad}
Y.~Jia and X.~Y. Zhou.
\newblock Policy gradient and actor-critic learning in continuous time and space: Theory and algorithms.
\newblock {\em The Journal of Machine Learning Research}, 23(1):12603--12652, 2022.

\bibitem[JZ23]{jia2023q}
Y.~Jia and X.~Y. Zhou.
\newblock {q}-learning in continuous time.
\newblock {\em Journal of Machine Learning Research}, 24(161):1--61, 2023.

\bibitem[Kak01]{kakade2001natural}
S.~M. Kakade.
\newblock A natural policy gradient.
\newblock {\em Advances in neural information processing systems}, 14, 2001.

\bibitem[KB23]{kobeissi2023temporal}
Z.~Kobeissi and F.~Bach.
\newblock Temporal difference learning with continuous time and state in the stochastic setting.
\newblock 2023.

\bibitem[KT99]{konda1999actor}
V.~Konda and J.~N. Tsitsiklis.
\newblock Actor-critic algorithms.
\newblock {\em Advances in neural information processing systems}, 12, 1999.

\bibitem[LM13]{lecue2013learning}
G.~Lecu{\'e} and S.~Mendelson.
\newblock Learning sub{G}aussian classes: Upper and minimax bounds.
\newblock {\em arXiv preprint arXiv:1305.4825}, 2013.

\bibitem[LVR22]{li2022neural}
X.~Li, D.~Verma, and L.~Ruthotto.
\newblock A neural network approach for stochastic optimal control.
\newblock {\em arXiv preprint arXiv:2209.13104}, 2022.

\bibitem[LWW24]{li2024error}
L.~Li, M.~Wang, and Y.~Wang.
\newblock Error estimates of the {E}uler's method for stochastic differential equations with multiplicative noise via relative entropy.
\newblock {\em arXiv preprint arXiv:2409.04991}, 2024.

\bibitem[Men15]{mendelson2015learning}
S.~Mendelson.
\newblock Learning without concentration.
\newblock {\em Journal of the ACM (JACM)}, 62(3):1--25, 2015.

\bibitem[MFWB22]{mou2022improved}
W.~Mou, N.~Flammarion, M.~J. Wainwright, and P.~L. Bartlett.
\newblock Improved bounds for discretization of {L}angevin diffusions: Near-optimal rates without convexity.
\newblock {\em Bernoulli}, 28(3):1577--1601, 2022.

\bibitem[MPW23]{mou2023optimal}
W.~Mou, A.~Pananjady, and M.~J. Wainwright.
\newblock Optimal oracle inequalities for projected fixed-point equations, with applications to policy evaluation.
\newblock {\em Mathematics of Operations Research}, 48(4):2308--2336, 2023.

\bibitem[MPWB24]{mou2024optimal}
W.~Mou, A.~Pananjady, M.~J. Wainwright, and P.~L. Bartlett.
\newblock Optimal and instance-dependent guarantees for markovian linear stochastic approximation.
\newblock {\em Mathematical Statistics and Learning}, 2024.

\bibitem[MPZ21]{menozzi2021density}
S.~Menozzi, A.~Pesce, and X.~Zhang.
\newblock Density and gradient estimates for non-degenerate {B}rownian {SDE}s with unbounded measurable drift.
\newblock {\em Journal of Differential Equations}, 272:330--369, 2021.

\bibitem[MR16]{marinelli2016maximal}
C.~Marinelli and M.~R{\"o}ckner.
\newblock On the maximal inequalities of {B}urkholder, {D}avis and {G}undy.
\newblock {\em Expositiones Mathematicae}, 34(1):1--26, 2016.

\bibitem[MZ24]{mou2024bellman}
W.~Mou and Y.~Zhu.
\newblock On {B}ellman equations for continuous-time policy evaluation {I}: discretization and approximation.
\newblock {\em arXiv preprint arXiv:2407.05966}, 2024.

\bibitem[NSW24]{neeman2024concentration}
J.~Neeman, B.~Shi, and R.~Ward.
\newblock Concentration inequalities for sums of {M}arkov-dependent random matrices.
\newblock {\em Information and Inference: A Journal of the IMA}, 13(4):iaae032, 2024.

\bibitem[O{\etalchar{+}}19]{openai2019dota}
OpenAI et~al.
\newblock Dota 2 with large scale deep reinforcement learning.
\newblock {\em arXiv preprint arXiv: 1912.06680}, 2019.

\bibitem[Rec19]{recht2019tour}
B.~Recht.
\newblock A tour of reinforcement learning: The view from continuous control.
\newblock {\em Annual Review of Control, Robotics, and Autonomous Systems}, 2(1):253--279, 2019.

\bibitem[ROL{\etalchar{+}}20]{ruthotto2020machine}
L.~Ruthotto, S.~J. Osher, W.~Li, L.~Nurbekyan, and S.~W. Fung.
\newblock A machine learning framework for solving high-dimensional mean field game and mean field control problems.
\newblock {\em Proceedings of the National Academy of Sciences}, 117(17):9183--9193, 2020.

\bibitem[SB18]{sutton2018reinforcement}
R.~S. Sutton and A.~G. Barto.
\newblock Reinforcement learning: An introduction.
\newblock {\em A Bradford Book}, 2018.

\bibitem[Sut88]{sutton1988learning}
R.~S. Sutton.
\newblock Learning to predict by the methods of temporal differences.
\newblock {\em Machine Learning}, 3(1):9--44, 1988.

\bibitem[SY19]{srikant2019finite}
R.~Srikant and L.~Ying.
\newblock Finite-time error bounds for linear stochastic approximation andtd learning.
\newblock In {\em Conference on Learning Theory}, pages 2803--2830. PMLR, 2019.

\bibitem[Sze22]{szepesvari2022algorithms}
Cs. Szepesv{\'a}ri.
\newblock {\em Algorithms for reinforcement learning}.
\newblock Springer Nature, 2022.

\bibitem[Tan24]{tang2024fine}
W.~Tang.
\newblock Fine-tuning of diffusion models via stochastic control: entropy regularization and beyond.
\newblock {\em arXiv preprint arXiv:2403.06279}, 2024.

\bibitem[Tro11]{tropp2011freedman}
J.~A. Tropp.
\newblock Freedman's inequality for matrix martingales.
\newblock {\em Electronic Communications in Probability}, 16:262--270, 2011.

\bibitem[Tsy08]{tsybakov2008introduction}
A.~B. Tsybakov.
\newblock {\em Introduction to Nonparametric Estimation}.
\newblock Springer Science \& Business Media, 2008.

\bibitem[TVR97]{tsitsiklis1997analysis}
J.~N. Tsitsiklis and B.~Van~Roy.
\newblock Analysis of temporal-diffference learning with function approximation.
\newblock In {\em Advances in Neural Information Processing Systems}, pages 1075--1081, 1997.

\bibitem[UZB{\etalchar{+}}24]{uehara2024fine}
M.~Uehara, Y.~Zhao, K.~Black, E.~Hajiramezanali, G.~Scalia, N.~L. Diamant, A.~M. Tseng, T.~Biancalani, and S.~Levine.
\newblock Fine-tuning of continuous-time diffusion models as entropy-regularized control.
\newblock {\em arXiv preprint arXiv:2402.15194}, 2024.

\bibitem[UZBL24]{uehara2024understanding}
M.~Uehara, Y.~Zhao, T.~Biancalani, and S.~Levine.
\newblock Understanding reinforcement learning-based fine-tuning of diffusion models: A tutorial and review.
\newblock {\em arXiv preprint arXiv:2407.13734}, 2024.

\bibitem[Wan05]{wang2005character}
F.-Y. Wang.
\newblock A character of the gradient estimate for diffusion semigroups.
\newblock {\em Proceedings of the American Mathematical Society}, 133(3):827--834, 2005.

\bibitem[WZ20]{wang2020continuous}
H.~Wang and X.~Y. Zhou.
\newblock Continuous-time mean--variance portfolio selection: A reinforcement learning framework.
\newblock {\em Mathematical Finance}, 30(4):1273--1308, 2020.

\bibitem[WZZ20]{wang2020reinforcement}
H.~Wang, T.~Zariphopoulou, and X.~Y. Zhou.
\newblock Reinforcement learning in continuous time and space: A stochastic control approach.
\newblock {\em The Journal of Machine Learning Research}, 21(1):8145--8178, 2020.

\bibitem[ZCZ{\etalchar{+}}24]{zhao2024scores}
H.~Zhao, H.~Chen, J.~Zhang, D.~D. Yao, and W.~Tang.
\newblock Scores as actions: a framework of fine-tuning diffusion models by continuous-time reinforcement learning.
\newblock {\em arXiv preprint arXiv:2409.08400}, 2024.

\bibitem[ZHL21]{zhou2021actor}
M.~Zhou, J.~Han, and J.~Lu.
\newblock Actor-critic method for high dimensional static {H}amilton--{J}acobi--{B}ellman partial differential equations based on neural networks.
\newblock {\em SIAM Journal on Scientific Computing}, 43(6):A4043--A4066, 2021.

\end{thebibliography}

\appendix
\section{Proofs of some auxiliary results}
We collect the proofs of some auxiliary results from the main paper in this appendix.

\subsection{Proof of \Cref{lemma:matrix-conc-markov}}\label{subsec:proof-lemma-matrix-conc-markov}
Let us first prove the result for the special case where each $Y_k$ are $\mbasis \times \mbasis$ symmetric square matrices. We will then extend the result to general matrices.

Letting $N_0 \mydefn \lceil \frac{4}{\lammin \poincare \stepsize} \rceil$ and $L \mydefn \lfloor \frac{N}{2 N_0} \rfloor$, we split the Markov process into two alternating processes, and define the partial sums
\begin{align*}
  Z_\ell^{(0)} \mydefn \sum_{j = 0}^{N_0 - 1} Y_{2 N_0 \ell + j}, \quad \mbox{and} \quad Z_\ell^{(1)} \mydefn \sum_{j = 0}^{N_0 - 1} Y_{2 N_0 \ell + N_0 + j}  \quad \mbox{for $\ell = 0,1,2, \cdots, L - 1$}.
\end{align*}
We also define the residual term
\begin{align*}
  Z' \mydefn \sum_{j = 2 N_0 L + 1}^N Y_{N_0 \ell + j}.
\end{align*}
Clearly, we have
\begin{align*}
   \opnorm{Z' - \Exs [Z']} \leq 2 N_0 \Big\{ \max_{0 \leq k \leq N} \opnorm{Y_k} + \opnorm{\Exs [Y_0]} \Big\} \leq 2 N_0 B.
\end{align*}
To bound the concentration of the discrete summation, we use the following matrix Bernstein inequality for Markov chains from~\cite{neeman2024concentration}.
\begin{proposition}[Theorem 2.6 of \cite{neeman2024concentration}]\label{prop:neeman-matrix-bernstein-mkv}
  Let $(\Omega_k)_{k \geq 0}$ be a discrete-time stationary Markov chain in any state spaces, with stationary distribution $\stationary$. If we have the contraction condition
  \begin{align*}
    \Exs \Big[\abss{\Exs \big[ f (\Omega_1) \mid \Omega_0 \big]}^2 \Big] \leq \lambda^2 \Exs \big[f (\Omega_0)^2\big], \quad \mbox{for any $f \in \ltwospace (\stationary)$},
  \end{align*}
  for some $\lambda \in (0, 1)$. Let $F$ be a function that maps the statespace to $\usedim \times \usedim$ symmetric matrices. If $F$ satisfies $\opnorm{F (\Omega)} \leq M$ almost surely and $\opnorm{\Exs_\stationary [F^2 (\Omega)]} \leq V$, we have
  \begin{align*}
    \Prob \Big\{ \lammax \Big( \sum_{k = 1}^n F (\Omega_k) \Big) \geq t \Big\} \leq d^{2 - \frac{\pi}{4}} \exp \Bigg( \frac{ - \pi^2 (1 - \lambda) t^2 }{32 (1 + \lambda)\numobs V + \frac{256}{\pi} M t} \Bigg),
  \end{align*}
  for any $t > 0$.
\end{proposition}
In order to apply this result, we construct two discrete-time Markov chains that take value in the space of trajectories with length $ N_0 \stepsize$.
\begin{align*}
  \Omega_\ell^{(0)} \mydefn \big\{ X_t:  2 \ell N_0 \stepsize \leq t \leq (2 \ell + 1) N_0 \stepsize \big\}, \quad \mbox{and} \quad \Omega_\ell^{(1)} \mydefn \big\{ X_t:  (2 \ell + 1) N_0 \stepsize \leq t \leq (2 \ell + 2) N_0 \stepsize \big\},
\end{align*}
for $\ell = 0,1,2, \cdots$. Furthermore, we define the matrix-valued function as
\begin{align*}
  F (\{ x_t : 0 \leq t \leq N_0 \stepsize \}) \mydefn  \sum_{j = 0}^{N_0 - 1} \Upsilon \big( \{ x_{j \stepsize + t}: t \in [0, \stepsize]\} \big).
\end{align*}
In order to verify the contraction property of the Markov chain $(\Omega_\ell^{(0)})_{\ell \geq 0}$, we note that $f(\Omega^{(0)}_1)$ is measurable in $\sigma (X_t: t \geq 2 N_0 \stepsize)$ while $\Omega_0^{(0)}$ is measurable in $\sigma (X_t: t \leq N_0 \stepsize)$. We can therefore use \Cref{lemma:poincare-contraction} to find that
\begin{align*}
 \Exs \Big[ \abss{\Exs \big[ f (\Omega^{(0)}_1) \mid \Omega_0^{(0)} \big] }^2 \Big] = \Exs \Big[ \abss{\Exs \big[ f (\Omega^{(0)}_1) \mid X_{N_0 \stepsize} \big] }^2 \Big] \leq e^{- \frac{\lammin \poincare}{4} N_0 \stepsize} \Exs \big[ \abss{f (\Omega^{(0)}_0)}^2 \big],
\end{align*}
for any square integrable function $f$. Since $N_0 \geq \frac{4}{\lammin \poincare}$, the contraction assumption is satisfied by $(\Omega_\ell^{(0)})_{\ell \geq 0}$ with $\lambda = e^{-1}$. Similarly, by stationarity of the underlying process $(\State_t)_{t \geq 0}$, the process $(\Omega_\ell^{(1)})_{\ell \geq 0}$ also satisfies the contraction assumption with $\lambda = e^{-1}$. So we are ready to apply \Cref{prop:neeman-matrix-bernstein-mkv}. Note that
\begin{align*}
  \opnorm{Z^{(0)}_0} &\leq N_0 B , \quad \mbox{almost surely, and}\\
  \opnorm{\Exs [ (Z_0^{(0)})^2]} &\leq \sum_{j = 0}^{N_0 - 1} \sum_{k = 0}^{N_0 - 1} \opnorm{\Exs \big[ Y_j Y_k \big]} = N_0 \matsnorm{\Exs \big[ Y_0^2 \big]}{\op, H} + \sum_{i = 1}^{N_0 - 1} (N_0 - i) \opnorm{\Exs \big[ Y_0 Y_i \big]}\\
  &\leq N_0 \cdot \Big\{ \sum_{i = 0}^{N_0 - 1} \opnorm{\Exs \big[ Y_0 Y_i \big]} \Big\}.
\end{align*}
Applying \Cref{prop:neeman-matrix-bernstein-mkv} to both $(Z^{(0)}_\ell)_{1 \leq \ell \leq L}$ and $(- Z^{(0)}_\ell)_{1 \leq \ell \leq L}$, with probability $1 - \delta$, we have that
\begin{align*}
  \opnorm{\sum_{\ell = 0}^{L - 1} \big( Z_\ell^{(0)} - \Exs[Z_0^{(0)}] \big) } \leq 8 \Big( \sum_{i = 0}^{N_0 - 1} \opnorm{\Exs \big[ Y_0 Y_i \big]} \Big)^{1/2} \sqrt{N_0 L \log (\mbasis / \delta)} + 30 N_0 B \log (\mbasis / \delta).
\end{align*}
Similarly, with probability $1 - \delta$, we have
\begin{align*}
  \opnorm{\sum_{\ell = 0}^{L - 1} \big( Z_\ell^{(1)} - \Exs[Z_0^{(1)}] \big) } \leq 8 \Big( \sum_{i = 0}^{N_0 - 1} \opnorm{\Exs \big[ Y_0 Y_i \big]} \Big)^{1/2} \sqrt{N_0 L \log (\mbasis / \delta)} + 30 N_0 B \log (\mbasis / \delta).
\end{align*}
Combining them together with the bound for the residual term, we conclude that
\begin{align}
  \opnorm{\frac{1}{N} \sum_{j = 0}^{N - 1} \big( Y_j - \Exs [Y] \big)} &\leq 8 \Big( \sum_{i = 0}^{N_0 - 1} \opnorm{\Exs \big[ Y_0 Y_i \big]} \Big)^{1/2} \sqrt{\frac{\log (\mbasis / \delta)}{N}} + 32 \frac{N_0 B}{N} \log (\mbasis / \delta), \nonumber\\
  &\leq 8 \Big( \sum_{i = 0}^{+ \infty} \opnorm{\Exs \big[ Y_0 Y_i \big]} \Big)^{1/2} \sqrt{\frac{\log (\mbasis / \delta)}{N}} + \frac{160 B}{\lammin \poincare \stepsize N} \log (\mbasis / \delta). \label{eq:mkv-matrix-concentration-symmetric-case}
\end{align}
with probability $1 - \delta$, which completes the proof in the symmetric case.

Now we extend the result for general (non-square) matrices. We consider the Hermitian dilation
\begin{align*}
  \widetilde{Y}_k = \begin{bmatrix}
    0 & Y_k\\ Y_k^\top & 0
  \end{bmatrix},
\end{align*}
which is a $(m_1 + m_2)$-dimensional symmetric matrix. It is easy to see that
\begin{align*}
  \opnorm{\widetilde{Y}_k} \leq 2 \opnorm{Y_k}, \quad \mbox{and} \quad \opnorm{\Exs \big[ \widetilde{Y}_0 \widetilde{Y}_j \big]} \leq \max \big\{ \opnorm{\Exs [Y_0^\top Y_j]}, \opnorm{\Exs [Y_0 Y_j^\top]} \big\}.
\end{align*}
Substituting to Eq~\eqref{eq:mkv-matrix-concentration-symmetric-case} completes the proof of \Cref{lemma:matrix-conc-markov}.

\subsection{Proof of \Cref{lemma:variance-matrix-conc-in-martingale-proof}}\label{subsec:proof-lemma-variance-matrix-conc-in-martingale-proof}
Define the random matrices
\begin{align*}
   M_k &\mydefn H_1^{-1/2} \psi (\MyState_{\ell (k) \stepsize}) \psi (\MyState_{\ell (k) \stepsize})^\top H_1^{-1/2} \cdot \int_0^{(\numerOrder - 1) \stepsize} \semigroup_t \Big( \mathrm{Tr} \big( \nabla \psi H_1^{-1} \nabla \psi^\top \big) \Big) (\MyState_{\ell (k) \stepsize}) dt, \quad \mbox{and}\\
   \widetilde{M}_k &\mydefn \vecnorm{\psi (\MyState_{\ell (k) \stepsize})}{H_1^{-1}}^2 \int_0^{(\numerOrder - 1) \stepsize} H_1^{-1/2}  \Exs \Big[ \nabla \psi (\MyState_{\ell (k) \stepsize + t}) \nabla \psi (\MyState_{\ell (k) \stepsize + t})^\top \mid \MyState_{\ell (k) \stepsize} \Big] H_1^{-1/2}  dt.
\end{align*}
Both $M_k$ and $\widetilde{M}_k$ are functionals of a stationary Markov process. For any unit vector $u \in \sphere^{\mbasis - 1}$, we note that
\begin{align*}
  u^\top \Exs [M_k] u &= \sum_{j \in [\mbasis]} \int_0^{(\numerOrder - 1) \stepsize} \Exs \Big[ \big( u^\top H_1^{-1/2} \psi (\MyState_{\ell (k) \stepsize}) \big)^2 \cdot  \vecnorm{ \coordinate_j^\top H_1^{-1/2} \nabla \psi (\MyState_{\ell (k) \stepsize + t})}{2}^2 \Big] dt\\
  &\leq \sum_{j \in [\mbasis]} \int_0^{(\numerOrder - 1) \stepsize} \lpstatnorm{u^\top H_1^{-1/2} \psi}{4}^2 \lpstatnorm{\coordinate_j^\top H_1^{-1/2} \nabla \psi}{4}^2 dt\\
  &\leq \numerOrder \stepsize \tau^4 \sum_{j \in [\mbasis]} \statnorm{u^\top H_1^{-1/2} \psi}^2 \statnorm{\coordinate_j^\top H_1^{-1/2} \nabla \psi}^2\\
  &\leq \numerOrder \stepsize \tau^4 \mbasis,
\end{align*}
and similarly,
\begin{align*}
  u^\top \Exs [\widetilde{M}_k] u = \sum_{j \in [\mbasis]} \int_0^{(\numerOrder - 1) \stepsize} \Exs \Big[ \big( \coordinate_j^\top H_1^{-1/2} \psi (\MyState_{\ell (k) \stepsize}) \big)^2 \cdot  \vecnorm{ u^\top H_1^{-1/2} \nabla \psi (\MyState_{\ell (k) \stepsize + t})}{2}^2 \Big] dt
  \leq \numerOrder \stepsize \tau^4 \mbasis.
\end{align*}
In order to derive upper bounds on the matrix sums, it suffices to prove concentration inequalities for $M_k$ and $\widetilde{M}_k$. By Assumption~\ref{assume:bounded-feature}, we have
\begin{align*}
  \max \big( \opnorm{M_k}, \opnorm{\widetilde{M}_k} \big) \leq \numerOrder \stepsize D_m^4.
\end{align*}
For the cross covariance, we note that for any vector $u \in \sphere^{\mbasis - 1}$, we have
\begin{align*}
  &u^\top \Exs \big[ M_k^2 \big] u \\
  &= \Exs \Big[  \big( u^\top H_1^{-1/2} \psi (\MyState_{\ell (k) \stepsize}) \big)^2 \vecnorm{H^{-1/2} \psi (\MyState_{\ell (k) \stepsize})}{2}^2 \Big\{ \int_0^{(\numerOrder - 1) \stepsize} \matsnorm{\nabla \psi (\MyState_{\ell (k) \stepsize + t}) H_1^{-1/2}}{F}^2 dt \Big\}^2 \Big]\\
  &\leq \numerOrder \stepsize \int_0^{(\numerOrder - 1) \stepsize} \sum_{j \in [\mbasis]} \Exs \Big[  \big( u^\top H_1^{-1/2} \psi (\MyState_{\ell (k) \stepsize}) \big)^2 \big( \coordinate_j^\top H^{-1/2} \psi (\MyState_{\ell (k) \stepsize}) \big)^2 \Big\{ \sum_{j' \in [\mbasis]} \vecnorm{\coordinate_{j'}^\top \nabla \psi (\MyState_{\ell (k) \stepsize + t}) H_1^{-1/2}}{2}^2 \Big\}^2 \Big]\\
  &\leq \numerOrder^2 \stepsize^2 \sum_{j_1, j_2, j_3 \in [\mbasis]} \lpstatnorm{u^\top H_1^{-1/2} \psi}{8}^2 \cdot  \lpstatnorm{\coordinate_{j_1}^\top H_1^{-1/2} \psi}{8}^2 \lpstatnorm{\coordinate_{j_2}^\top H_1^{-1/2} \nabla \psi}{8}^2 \lpstatnorm{\coordinate_{j_3}^\top H_1^{-1/2} \nabla \psi}{8}^2 \\
  &\leq \tau^8 \numerOrder^2 \stepsize^2 \mbasis^3.
\end{align*}
For time index $k \geq 1$, we note that
\begin{align*}
  &u^\top \Exs \big[ M_0 M_k \big] u\\
  &= \sum_{j \in [\mbasis]} \Exs \Big[ \big( u^\top H_1^{-1/2} \psi (\MyState_{\ell (0) \stepsize}) \big)  \big( \coordinate_j^\top H_1^{-1/2} \psi (\MyState_{\ell (0) \stepsize}) \big) \big( u^\top H_1^{-1/2} \psi (\MyState_{\ell (k) \stepsize}) \big)  \big( u^\top H_1^{-1/2} \psi (\MyState_{\ell (k) \stepsize}) \big) \\
  &\qquad \qquad \cdot \Big\{ \int_0^{(\numerOrder - 1) \stepsize} \matsnorm{\nabla \psi (\MyState_{\ell (0) \stepsize + t}) H_1^{-1/2}}{F}^2 dt \Big\} \Big\{ \int_0^{(\numerOrder - 1) \stepsize} \matsnorm{\nabla \psi (\MyState_{\ell (k) \stepsize + t}) H_1^{-1/2}}{F}^2 dt \Big\}\Big].
\end{align*}
Invoking the contraction result in \Cref{lemma:poincare-contraction}, we have
\begin{align*}
  &\var \Bigg( \Exs \Big[ \big( u^\top H_1^{-1/2} \psi (\MyState_{\ell (k) \stepsize}) \big)  \big( u^\top H_1^{-1/2} \psi (\MyState_{\ell (k) \stepsize}) \big) \Big\{ \int_0^{(\numerOrder - 1) \stepsize} \matsnorm{\nabla \psi (\MyState_{\ell (k) \stepsize + t}) H_1^{-1/2}}{F}^2 dt \Big\} \mid \State_{\ell (1) \stepsize} \Big] \Bigg)\\
  &\leq e^{- \frac{\poincare \lammin}{2} (k -1) \stepsize (\numerOrder - 1)} \Exs \Big[ \big( u^\top H_1^{-1/2} \psi (\MyState_{\ell (k) \stepsize}) \big)^2  \big( u^\top H_1^{-1/2} \psi (\MyState_{\ell (k) \stepsize}) \big)^2 \Big\{ \int_0^{(\numerOrder - 1) \stepsize} \matsnorm{\nabla \psi (\MyState_{\ell (k) \stepsize + t}) H_1^{-1/2}}{F}^2 dt \Big\}^2 \Big].
\end{align*}
Substituting to the expression for $u^\top \Exs \big[ M_0 M_k \big] u$ and invoking Cauchy--Schwarz inequality for each term, we obtain that
\begin{align*}
  &\abss{u^\top \Exs \big[ (M_0 - \Exs [M_0]) (M_k - \Exs [M_k]) \big] u}\\
  & \leq e^{- \frac{\poincare \lammin}{2} \ell (k - 1) \stepsize} \numerOrder^2 \stepsize^2 \sum_{j_1, j_2, j_3 \in [\mbasis]} \lpstatnorm{u^\top H_1^{-1/2} \psi}{8}^2 \cdot  \lpstatnorm{\coordinate_{j_1}^\top H_1^{-1/2} \psi}{8}^2 \lpstatnorm{\coordinate_{j_2}^\top H_1^{-1/2} \nabla \psi}{8}^2 \lpstatnorm{\coordinate_{j_3}^\top H_1^{-1/2} \nabla \psi}{8}^2 \\
  &\leq e^{- \frac{\poincare \lammin}{4} \ell (k - 1) \stepsize} \tau^8 \numerOrder^2 \stepsize^2 \mbasis^3.
\end{align*}
So we have
\begin{align*}
  \sum_{k \geq 0} \opnorm{\Exs \big[ (M_0 - \Exs [M_0]) ( M_k - \Exs [M_k]) \big]} \leq \numerOrder^2 \stepsize^2 \mbasis^3 \sum_{k \geq 0} e^{- \frac{\poincare \lammin}{4} \ell (k - 1) \stepsize} \leq \frac{4 \tau^8}{\poincare \lammin} \numerOrder \stepsize \mbasis^3.
\end{align*}
Similarly, we have the bound
\begin{align*}
  \sum_{k \geq 0} \opnorm{\Exs \big[ \big( \widetilde{M}_0 - \Exs[\widetilde{M}_0] \big) \big( \widetilde{M}_k - \Exs [\widetilde{M}_k] \big) \big]} \leq \numerOrder^2 \stepsize^2 \mbasis^3 \sum_{k \geq 0} e^{- \frac{\poincare \lammin}{4} \ell (k - 1) \stepsize} \leq \frac{4 \tau^8}{\poincare \lammin} \numerOrder \stepsize \mbasis^3.
\end{align*}
Invoking \Cref{lemma:matrix-conc-markov}, for any fixed $n \in [1, N]$, with probability $1 - \delta$, we have
\begin{align*}
  \opnorm{\sum_{k = 0}^{n - 1} M_k} &\leq n \opnorm{\Exs [M_0]} + c \sqrt{n \log (\mbasis / \delta)}\Big\{ \sum_{k \geq 0} \opnorm{\Exs \big[ (M_0 - \Exs [M_0]) ( M_k - \Exs [M_k]) \big]} \Big\}^{1/2} \\ &\qquad \qquad + c \numerOrder \stepsize D_\mbasis^4 \log (\mbasis / \delta)\\
  &\leq \tau^4 \numerOrder \mbasis T + c \numerOrder \tau^4 \mbasis \sqrt{\frac{T \mbasis \log (\mbasis / \delta)}{\poincare \lammin}} + c \numerOrder \stepsize D_\mbasis^4 \log (\mbasis / \delta),
\end{align*}
and similarly, with probability $1 - \delta$, we have
\begin{align*}
  \opnorm{\sum_{k = 0}^{n - 1} \widetilde{M}_k} &\leq n \opnorm{\Exs [\widetilde{M}_0]} + c \sqrt{n \log (\mbasis / \delta)}\Big\{ \sum_{k \geq 0} \opnorm{\Exs \big[ (\widetilde{M}_0 - \Exs [\widetilde{M}_0]) ( \widetilde{M}_k - \Exs [\widetilde{M}_k]) \big]} \Big\}^{1/2} \\ &\qquad \qquad + c \numerOrder \stepsize D_\mbasis^4 \log (\mbasis / \delta)\\
  &\leq \tau^4 \numerOrder \mbasis T + c \numerOrder \tau^4 \mbasis \sqrt{\frac{T \mbasis \log (\mbasis / \delta)}{\poincare \lammin}} + c \numerOrder \stepsize D_\mbasis^4 \log (\mbasis / \delta).
\end{align*}
By taking union bound over $n \in \{1,2, \cdots, N\}$, with probability $1 - \delta$, we have that
\begin{align*}
  \max \big\{ \opnorm{W_n}, \opnorm{\widetilde{W}_n} \big\} \leq \tau^4 \numerOrder \mbasis T + c \numerOrder \tau^4 \mbasis \sqrt{\frac{T \mbasis \log (\mbasis N / \delta)}{\poincare \lammin}} + c \numerOrder \stepsize D_\mbasis^4 \log (\mbasis N / \delta)
\end{align*}

Substituting with a trajectory length satisfying $T \geq c \frac{\mbasis}{\poincare \lammin} \log (\mbasis N / \delta)$, we complete the proof of this lemma.

\subsection{Proof of \Cref{lemma:variance-additional-term-bound}}\label{subsec:app-proof-of-lemma-variance-additional-term-bound}
Note that $\Exs [\widetilde{\zeta}_{\ell (k)}] = \Exs [\widebar{\zeta}_{\ell (k)}] = 0$. We start by decomposing the variance in two terms using Young's inequality.
\begin{align*}
  &\Exs \big[ \vecnorm{ \frac{1}{N} \sum_{k = 0 }^{N / (\numerOrder - 1)} H_1^{-1/2} \psi (X_{{\ell (k)} \stepsize}) \big( \widetilde{\zeta}_{\ell (k)} - \widebar{\zeta}_{\ell (k)} \big)}{2}^2 \big]\\
  &\leq 2 \sum_{j = 1}^\mbasis \var \Big(  \frac{1}{N} \sum_{k = 0 }^{N / (\numerOrder - 1)} \coordinate_j^\top H_1^{-1/2} \psi (\State_{{\ell (k)} \stepsize}) \frac{1}{\stepsize} \int_0^{(\numerOrder - 1) \stepsize} e^{- \discount t} \big( \discount - \generator\big) (\ftil - \fbar) (\MyState_{\ell (k) \stepsize + t}) dt  \Big)\\
  &\qquad + 2 \sum_{j = 1}^\mbasis \var \Big( \frac{1}{N} \sum_{k = 0 }^{N / (\numerOrder - 1)} \coordinate_j^\top H_1^{-1/2} \psi (X_{{\ell (k)} \stepsize}) \frac{1}{\stepsize} \int_0^{(\numerOrder - 1) \stepsize} e^{- \discount t} \big( \widebar{\reward}_{\ell (k) \stepsize + t} - \reward (\MyState_{ \ell (k) \stepsize + t}) \big) dt \Big).
\end{align*}
For the first term, we note that
\begin{align*}
  &\var \Big(  \frac{1}{N} \sum_{k = 0 }^{N / (\numerOrder - 1)} \coordinate_j^\top H_1^{-1/2} \psi (X_{{\ell (k)} \stepsize}) \frac{1}{\stepsize} \int_0^{(\numerOrder - 1) \stepsize} e^{- \discount t} \big( \discount - \generator\big) (\ftil - \fbar) (\MyState_{k \stepsize + t}) dt  \Big)\\ 
  &\leq \frac{2}{N^2} \sum_{k = 0}^{N / (\numerOrder - 1)} (\tfrac{N}{\numerOrder - 1} - k) \Bigg| \cov \Big(   \coordinate_j^\top H_1^{-1/2} \psi (X_{{\ell (0)} \stepsize}) \frac{1}{\stepsize} \int_0^{(\numerOrder - 1) \stepsize} e^{- \discount t} \big( \discount - \generator\big) (\ftil - \fbar) (\MyState_{\ell (0) \stepsize + t}) dt ,\\
  & \qquad \qquad \qquad \qquad  \qquad \qquad \qquad \coordinate_j^\top H_1^{-1/2} \psi (X_{{\ell (k)} \stepsize}) \frac{1}{\stepsize} \int_0^{(\numerOrder - 1) \stepsize} e^{- \discount t} \big( \discount - \generator\big) (\ftil - \fbar) (\MyState_{\ell (k) \stepsize + t}) dt \Big) \Bigg| \\ 
  &\leq \frac{1}{N} \sum_{k = 0}^{+ \infty} \abss{\mu_k (\coordinate_j^\top H_1^{-1/2} \psi, \ftil - \fbar)}\\ 
  &\leq \frac{\tau^2}{T} \Big\{  \constScaryReg (T_0) \sobopstatnorm{\ftil - \fbar}{2p}^2 \log (1 / \stepsize) + \frac{\constScary}{\poincare} e^{- \frac{\poincare \lammin }{4} T_0} \sobopstatnorm{\ftil - \fbar}{2p} \cdot \sobokpstatnorm{\ftil - \fbar}{2}{2p} + \stepsize \constScary \sobokpstatnorm{\ftil - \fbar}{2}{2p}^2 \Big\}.
\end{align*}
By applying Assumptions~\fakerefassumehyper{$\tfrac{4p}{p - 1}$}{$\tau$},~\ref{assume:basis-condition}, for stepsize satisfying $\stepsize \leq \mbasis^{- 2 \omega}$, we have that
\begin{align}
  &\sum_{j = 1}^\mbasis \var \Big(  \frac{1}{N} \sum_{k = 0 }^{N / (\numerOrder - 1)} \coordinate_j^\top H_1^{-1/2} \psi (X_{{\ell (k)} \stepsize}) \frac{1}{\stepsize} \int_0^{(\numerOrder - 1) \stepsize} e^{- \discount t} \big( \discount - \generator\big) (\ftil - \fbar) (\MyState_{k \stepsize + t}) dt  \Big) \nonumber \\ 
  &\leq \tau^4 \frac{\mbasis}{T} \sobonorm{\ftil - \fbar}^2  \Big\{ \constScaryReg (T_0)  \log (1 / \stepsize) + \constScary  \poincare^{-1} \mbasis^{\omega} + \constScary \stepsize \mbasis^{2 \omega} \Big\} \nonumber \\ 
  &\leq \frac{\Tthres (\mbasis, T_0)}{T} \sobonorm{\ftil - \fbar}^2.\label{eq:variance-additional-term-bound-first-term}
\end{align}
For the second term in the decomposition, we note that
\begin{align*}
  &\var \Big( \frac{1}{N} \sum_{k = 0 }^{N / (\numerOrder - 1)} \coordinate_j^\top H_1^{-1/2} \psi (X_{{\ell (k)} \stepsize}) \frac{1}{\stepsize} \int_0^{(\numerOrder - 1) \stepsize} e^{- \discount t} \big( \widebar{\reward}_{\ell (k) \stepsize + t} - \reward (\MyState_{ \ell (k) \stepsize + t}) \big) dt \Big) \\ 
  &\leq \frac{1}{N} \sum_{k = 0}^{N / (\numerOrder - 1)}  \Bigg| \cov \Big( \coordinate_j^\top H_1^{-1/2} \psi (X_{{\ell (0)} \stepsize}) \frac{1}{\stepsize} \int_0^{(\numerOrder - 1) \stepsize} e^{- \discount t} \big( \widebar{\reward}_{\ell (0) \stepsize + t} - \reward (\MyState_{ \ell (0) \stepsize + t}) \big) dt ,\\
  & \qquad \qquad \qquad \qquad  \qquad \qquad \qquad \coordinate_j^\top H_1^{-1/2} \psi (X_{{\ell (k)} \stepsize}) \frac{1}{\stepsize} \int_0^{(\numerOrder - 1) \stepsize} e^{- \discount t} \big( \widebar{\reward}_{\ell (k) \stepsize + t} - \reward (\MyState_{ \ell (k) \stepsize + t}) \big) dt \Big) \Bigg|.
\end{align*}
For $k \geq 1$, by \Cref{lemma:poincare-contraction} and Cauchy--Schwarz inequality, we have
\begin{align*}
  &\Bigg|\cov \Big( \coordinate_j^\top H_1^{-1/2} \psi (X_{{\ell (0)} \stepsize}) \frac{1}{\stepsize} \int_0^{(\numerOrder - 1) \stepsize} e^{- \discount t} \big( \widebar{\reward}_{\ell (0) \stepsize + t} - \reward (\MyState_{ \ell (0) \stepsize + t}) \big) dt, \\
  &\qquad \qquad  \coordinate_j^\top H_1^{-1/2} \psi (X_{{\ell (k)} \stepsize}) \frac{1}{\stepsize} \int_0^{(\numerOrder - 1) \stepsize} e^{- \discount t} \big( \widebar{\reward}_{\ell (k) \stepsize + t} - \reward (\MyState_{ \ell (k) \stepsize + t}) \big) dt \Big)\Bigg|\\
  &= \Bigg|\cov \Big( \coordinate_j^\top H_1^{-1/2} \psi (X_{{\ell (0)} \stepsize}) \frac{1}{\stepsize} \int_0^{(\numerOrder - 1) \stepsize} e^{- \discount t} \big( \widebar{\reward}_{\ell (0) \stepsize + t} - \reward (\MyState_{ \ell (0) \stepsize + t}) \big) dt, \\
  &\qquad \qquad  \coordinate_j^\top H_1^{-1/2} \psi (X_{{\ell (k)} \stepsize}) \frac{1}{\stepsize} \int_0^{(\numerOrder - 1) \stepsize} e^{- \discount t} \Exs \big[ \widebar{\reward}_{\ell (k) \stepsize + t} - \reward (\MyState_{ \ell (k) \stepsize + t}) \mid \State_{\ell (k) \stepsize} \big] dt \Big)\Bigg|\\
  &\leq e^{- \frac{\poincare \lammin}{4} (\numerOrder - 1)  (k - 1) \stepsize } \var \Big(  \coordinate_j^\top H_1^{-1/2} \psi (X_{{\ell (0)} \stepsize}) \frac{1}{\stepsize} \int_0^{(\numerOrder - 1) \stepsize} e^{- \discount t} \big( \widebar{\reward}_{\ell (0) \stepsize + t} - \reward (\MyState_{ \ell (0) \stepsize + t}) \big) dt  \Big)^{1/2} \\
  & \qquad \qquad \cdot \var \Big(  \coordinate_j^\top H_1^{-1/2} \psi (X_{{\ell (k)} \stepsize}) \frac{1}{\stepsize} \int_0^{(\numerOrder - 1) \stepsize} e^{- \discount t} \Exs \big[ \widebar{\reward}_{\ell (k) \stepsize + t} - \reward (\MyState_{ \ell (k) \stepsize + t}) \mid \State_{\ell (k) \stepsize} \big] dt  \Big)^{1/2}.
\end{align*}
For the two variance terms, we note that
\begin{align*}
  &\Exs \Big[ \abss{ \coordinate_j^\top H_1^{-1/2} \psi (X_{{\ell (0)} \stepsize}) \frac{1}{\stepsize} \int_0^{(\numerOrder - 1) \stepsize} e^{- \discount t} \big( \widebar{\reward}_{\ell (0) \stepsize + t} - \reward (\MyState_{ \ell (0) \stepsize + t}) \big) dt }^2 \Big]\\
  &\leq 4 (\numerOrder - 1)^2 \Exs  \Big[ \big( \coordinate_j^\top H_1^{-1/2} \psi (X_{{\ell (0)} \stepsize}) \big)^2 \sup_{0 \leq t \leq (\numerOrder - 1) \stepsize} \abss{\reward (\State_{\ell (0) \stepsize + t}) - \reward (\State_{\ell (0) \stepsize })}^2 \Big]\\
  &\leq 4 (\numerOrder - 1)^2 \tau^2 \statnorm{\coordinate_j H_1^{-1/2} \psi}^2 \cdot \Exs \Big[ \sup_{0 \leq t \leq (\numerOrder - 1) \stepsize} \abss{\reward (\State_{\ell (0) \stepsize + t}) - \reward (\State_{\ell (0) \stepsize }) }^4 \Big]^{1/2}.
\end{align*}
Applying It\^{o}'s formula to the reward function and invoking the Burkholder--Davis--Gundy inequality  associated continuous martingales, we have
\begin{align*}
  &\Exs \Big[ \sup_{0 \leq t \leq (\numerOrder - 1) \stepsize} \abss{\reward (\State_{\ell (0) \stepsize + t}) - \reward (\State_{\ell (0) \stepsize }) }^4 \Big]\\
  &\leq 8 \Exs \Big[\abss{ \int_0^{(\numerOrder - 1) \stepsize}  \drift (\State_{\ell (0) \stepsize + t})^\top \nabla \reward (\State_{\ell (0) \stepsize + t}) + \frac{1}{2} \mathrm{Tr} \Big( \covMat (\State_{\ell (0) \stepsize + t}) \nabla^2 \reward (\State_{\ell (0) \stepsize + t}) \Big) dt }^4 \Big]\\
  &\qquad + 8 \Exs \Big[ \sup_{0 \leq t \leq (\numerOrder - 1) \stepsize } \abss{ \int_0^{t} \nabla \reward (\State_{\ell (0) \stepsize + s})^\top \covMat (\State_{\ell (0) \stepsize + s})^{1/2} dB_s }^4 \Big]\\
  &\leq 8 \usedim^4 \big( \smoothness_0^\drift \smoothness_1^\reward + \smoothness_0^\covMat \smoothness_2^\reward \big)^4 (\numerOrder - 1)^4 \stepsize^4 + 8 c \usedim^4 \big( \smoothness_1^\reward \smoothness_0^\covMat \big)^2 (\numerOrder - 1)^2 \stepsize^2.
\end{align*}
Therefore, we have
\begin{align*}
   \Exs \Big[ \abss{ \coordinate_j^\top H_1^{-1/2} \psi (X_{{\ell (0)} \stepsize}) \frac{1}{\stepsize} \int_0^{(\numerOrder - 1) \stepsize} e^{- \discount t} \big( \widebar{\reward}_{\ell (0) \stepsize + t} - \reward (\MyState_{ \ell (0) \stepsize + t}) \big) dt }^2 \Big] \leq  \tau^2 \constScary \stepsize \statnorm{\coordinate_j^\top H_1^{-1/2} \psi}^2.
\end{align*}
On the other hand, for $k \geq 1$, we have
\begin{align*}
  \abss{\Exs \big[ \widebar{\reward}_{\ell (k) \stepsize + t} - \reward (\MyState_{ \ell (k) \stepsize + t}) \mid \State_{\ell (k) \stepsize} \big]}
  = \abss{ \Big( \sum_{i = 0}^{\numerOrder - 1} W_i (t) \semigroup_{i \stepsize} \reward - \semigroup_t \reward \Big) (\MyState_{\ell (k) \stepsize}) }
  \leq \constScary \stepsize^{\numerOrder},
\end{align*}
where the last step follows from Lemma 1 of the paper~\cite{mou2024bellman}, and the constant $\constScary$ depends on the derivatives bounds in Assumption~\fakerefassumelip{$(\numerOrder)$}. So we have
\begin{multline*}
  \var \Big(  \coordinate_j^\top H_1^{-1/2} \psi (X_{{\ell (k)} \stepsize}) \frac{1}{\stepsize} \int_0^{(\numerOrder - 1) \stepsize} e^{- \discount t} \Exs \big[ \widebar{\reward}_{\ell (k) \stepsize + t} - \reward (\MyState_{ \ell (k) \stepsize + t}) \mid \State_{\ell (k) \stepsize} \big] dt  \Big) \\ \leq (\numerOrder - 1)^2 \constScary^2 \stepsize^{2 \numerOrder} \statnorm{\coordinate_j^\top H_1^{-1/2} \psi}^2.
\end{multline*}
Collecting the bounds together, we have
\begin{align*}
  &\sum_{j = 1}^\mbasis \var \Big( \frac{1}{N} \sum_{k = 0 }^{N / (\numerOrder - 1)} \coordinate_j^\top H_1^{-1/2} \psi (X_{{\ell (k)} \stepsize}) \frac{1}{\stepsize} \int_0^{(\numerOrder - 1) \stepsize} e^{- \discount t} \big( \widebar{\reward}_{\ell (k) \stepsize + t} - \reward (\MyState_{ \ell (k) \stepsize + t}) \big) dt \Big)\\
  &\leq \constScary \mathrm{Tr} \big( H_1^{-1} H_0 \big) \Big\{ \frac{\tau^2 \stepsize}{N} + \frac{1}{N} \sum_{k = 1}^{+ \infty} e^{- \frac{\poincare \lammin}{4} (\numerOrder - 1) (k - 1) \stepsize} \sqrt{\tau^2 \stepsize^2 \cdot \stepsize^{2 \numerOrder}} \Big\}\\
  &\leq \constScary \mathrm{Tr} \big( H_1^{-1} H_0 \big) \Big\{ \frac{\tau^2 \stepsize^2}{T} + \frac{\tau \stepsize^{1 + \numerOrder}}{T \poincare \lammin} \Big\}.
\end{align*}
Combining with Equation~\eqref{eq:variance-additional-term-bound-first-term}, we conclude that
\begin{multline*}
  \Exs \big[ \vecnorm{ \frac{1}{N} \sum_{k = 0 }^{N / (\numerOrder - 1)} H_1^{-1/2} \psi (X_{{\ell (k)} \stepsize}) \big( \widetilde{\zeta}_{\ell (k)} - \widebar{\zeta}_{\ell (k)} \big)}{2}^2 \big]\\
  \leq \frac{1}{T} \Big\{ \sobonorm{\ftil - \fbar}^2 \Tthres (\mbasis, T_0) + \tau^4 \constScary \mathrm{Tr} \big( H_1^{-1} H_0 \big) \Big( \stepsize^2 +  \frac{ \stepsize^{1 + \numerOrder}}{\poincare \lammin} \Big) \Big\},
\end{multline*}
which completes the proof of this lemma.

\subsection{Proof of \Cref{lemma:relate-approx-error-to-best-err}}\label{subsec:app-proof-of-lemma-relate-approx-error-to-best-err}
By definition, $\ftil$ satisfies the orthogonality condition
\begin{align*}
  \statinprod{\mathcal{L} \big(\ftil - \ValTrue \big)}{h} = 0, \quad \mbox{for any $h \in \LinSpace$}.
\end{align*}
where the operator $\mathcal{L}$ is defined as
\begin{align*}
  \mathcal{L} \mydefn \frac{\IdMat - e^{- \discount t} \semigroup_{(\numerOrder - 1) \stepsize}}{(\numerOrder - 1) \stepsize}.
\end{align*}
By Lemma 5, 6 and 7 of the paper~\cite{mou2024bellman}, the semigroup operator satisfies
\begin{subequations}
\begin{align}
  \statinprod{g}{\mathcal{L} g} &\geq c_1 \sobonorm{g}^2, \quad \mbox{for $g \in \LinSpace$}, \label{eq:lower-bound-in-approx-err-lemma-proof} \\
  \abss{\statinprod{f}{\mathcal{L} g} } &\leq c_2 \sobonorm{f} \sobonorm{g}, \quad \mbox{for $f, g \in \mathbb{H}^1 (\stationary)$},\label{eq:upper-bound-in-approx-err-lemma-proof}
\end{align}
\end{subequations}
for a pair of constants $c_1, c_2 > 0$ depending on the parameters in Assumptions~\ref{assume:uniform-elliptic}, \ref{assume:smooth-stationary} and~\fakerefassumelip{$(2)$}.

By the orthogonality condition $\statinprod{\mathcal{L} \big(\ftil - \ValTrue \big)}{\ftil - \projecttolinhil (\ValTrue)} = 0$, we have
\begin{align*}
  &c_1 \sobonorm{\ftil - \projecttolinhil (\ValTrue)}^2 \overset{(i)}{\leq} \statinprod{\mathcal{L} \big(\ftil - \projecttolinhil (\ValTrue) \big)}{\ftil - \projecttolinhil (\ValTrue) } \\
  &=  \statinprod{\mathcal{L} \big(\ValTrue - \projecttolinhil (\ValTrue) \big)}{\ftil - \projecttolinhil (\ValTrue) }\overset{(ii)}{\leq} c_2  \sobonorm{\ValTrue - \projecttolinhil (\ValTrue)} \sobonorm{\ftil - \projecttolinhil (\ValTrue)},
\end{align*}
where $(i)$ follows from Equation~\eqref{eq:lower-bound-in-approx-err-lemma-proof}, and $(ii)$ follows from Equation~\eqref{eq:upper-bound-in-approx-err-lemma-proof}. By rearranging the terms, we have
\begin{align*}
  \sobonorm{\ftil - \projecttolinhil (\ValTrue)} \leq \frac{c_2}{c_1} \sobonorm{\ValTrue - \projecttolinhil (\ValTrue)} = \frac{c_2}{c_1} \inf_{f \in \LinSpace} \sobonorm{f - \ValTrue}.
\end{align*}
By triangle inequality, we have
\begin{align*}
  \sobopstatnorm{\ftil - \ValTrue}{q} &\leq \sobopstatnorm{\ftil - \projecttolinhil (\ValTrue)}{2p} +  \sobopstatnorm{\projecttolinhil (\ValTrue) - \ValTrue}{q}\\
  & \leq \frac{c_2}{c_1} \tau \inf_{f \in \LinSpace} \sobonorm{f - \ValTrue} + \sobopstatnorm{\projecttolinhil (\ValTrue) - \ValTrue}{q},
\end{align*}
which proves the first bound in this lemma.

As for higher derivatives, we note that
\begin{align*}
  \sobokpstatnorm{\ftil - \ValTrue}{2}{q} &\leq \sobokpstatnorm{\ftil - \projecttolinhil (\ValTrue)}{2}{q} + \sobokpstatnorm{\projecttolinhil (\ValTrue) - \ValTrue}{2}{q}\\
   &\leq \tau \sobohilnorm{\ftil - \projecttolinhil (\ValTrue)}{2} + \sobokpstatnorm{\projecttolinhil (\ValTrue) - \ValTrue}{2}{q}\\
  &\leq \tau \mbasis^\omega \sobonorm{\ftil - \projecttolinhil (\ValTrue)} + \sobokpstatnorm{\projecttolinhil (\ValTrue) - \ValTrue}{2}{q}\\
  &\leq \frac{c_2 \tau}{c_1} \mbasis^\omega \inf_{f \in \LinSpace} \sobonorm{f - \ValTrue} + \sobokpstatnorm{\projecttolinhil (\ValTrue) - \ValTrue}{2}{q},
\end{align*}
which proves the second bound in this lemma.

For the last bound, we define the auxiliary function $\valfun^\numerOrder$ as the unique fixed-point to the following operator equation
\begin{align*}
  \valfun^\numerOrder = \sum_{i = 0}^{\numerOrder - 1} \int_0^{(\numerOrder - 1) \stepsize} e^{- \discount t} W_i (t) \semigroup_{i \stepsize} \reward dt + e^{- \discount (\numerOrder - 1) \stepsize} \semigroup_{(\numerOrder - 1) \stepsize} \valfun^\numerOrder.
\end{align*}
Following Theorem 1 of the paper~\cite{mou2024bellman}, such an operator is a contraction mapping under $L^{\infty}$-norm, and the fixed-point $\valfun^\numerOrder$ exists and is unique.

By Lemma 9 of the paper~\cite{mou2024bellman}, under Assumption~\fakerefassumelip{$(\numerOrder + 1)$}, we have
\begin{align*}
  \sobonorm{\valfun^\numerOrder - \ValTrue} \leq \constScary \stepsize^{\numerOrder}.
\end{align*}
On the other hand, since $\fbar$ is the fixed point to the projected time-discretized Bellmen operator, we have the orthogonality condition
\begin{align*}
  \statinprod{\mathcal{L} (\fbar - \valfun^\numerOrder)}{h} = 0, \quad \mbox{for any $h \in \LinSpace$}.
\end{align*}

To relate the error in the projected space to the time discretization error $\valfun^\numerOrder - \ValTrue$, we note that
\begin{align*}
  c_1 \sobonorm{\ftil - \fbar}^2 &\overset{(i)}{\leq} \statinprod{\mathcal{L} (\ftil - \fbar)}{\ftil - \fbar}\\
  &= \statinprod{\mathcal{L} (\ValTrue - \valfun^\numerOrder)}{\ftil - \fbar} + \statinprod{\mathcal{L} (\ftil - \ValTrue)}{\ftil - \fbar} +  \statinprod{\mathcal{L} (\valfun^\numerOrder - \fbar)}{\ftil - \fbar}  \\
  &\overset{(ii)}{=}  \statinprod{\mathcal{L} (\ValTrue - \valfun^\numerOrder)}{\ftil - \fbar}\\
  &\overset{(iii)}{\leq} c_2 \sobonorm{\ValTrue - \valfun^\numerOrder} \sobonorm{\ftil - \fbar},
\end{align*}
where $(i)$ follows from Equation~\eqref{eq:lower-bound-in-approx-err-lemma-proof}, $(ii)$ follows from the orthogonality conditions, and $(iii)$ follows from Equation~\eqref{eq:upper-bound-in-approx-err-lemma-proof}. By rearranging the terms and combining with the discretization error bound, we have
\begin{align*}
  \sobonorm{\ftil - \fbar} \leq \frac{c_2}{c_1} \sobonorm{\ValTrue - \valfun^\numerOrder} \leq \frac{c_2}{c_1} \constScary \stepsize^{\numerOrder},
\end{align*}
which proves the last bound in this lemma.

\section{Proofs of some examples}
In this section, we collect the proofs of some results related to the examples presented in the main text.
\subsection{Proof of \Cref{prop:advantage-func}}\label{subsec:app-proof-of-prop-advantage-func}
Define the finite-stepsize advantage function as
\begin{align*}
  \advFunc_{\Delta t} (\state, \action) \mydefn \Exs_\state \Big[ \int_0^{\Delta t} e^{- \discount t} \reward (\MyState_t^\action) dt + e^{- \discount \Delta t} \ValFun^\policy (\MyState^\action_{\Delta t}) - \ValFun^\policy (\state)\Big].
\end{align*}
Within the time interval $[0, \Delta t]$, the process $\State_t^\action$ evolves according to the dynamics
\begin{align*}
  d \MyState_t^\action = \ctrDrift (\MyState_t^\action, \action) dt + \covMat^{1/2} (\MyState_t^\action) dB_t.
\end{align*}
Applying It\^{o}'s formula, we have
\begin{align*}
  \advFunc_{\Delta t} (\state, \action)  \mydefn  \int_0^{\Delta t} e^{- \discount t}  \Exs_\state \Big[ \reward (\MyState_t^\action)  - \discount \ValFun^\policy (\MyState^\action_t) + \inprod{\nabla \ValFun^\policy (\MyState_t^\action)}{\ctrDrift (\MyState_t^\action, \action)} + \frac{1}{2} \mathrm{Tr} \big( \covMat (\MyState_t^\action) \cdot \nabla^2 \ValFun^\policy (\MyState_t^\action) \big) \Big] dt.
\end{align*}
On the other hand, the value function $\ValFunc^\policy$ satisfies the elliptic equation
\begin{align*}
\discount \ValFun^\policy - \generator^\policy \ValFun^\policy = \reward.
\end{align*}
Substituting into above expression, we obtain
\begin{align*}
  \advFunc_{\Delta t} (\state, \action) \mydefn  \int_0^{\Delta t} e^{- \discount t}  \Exs_\state \Big[ \inprod{\nabla \ValFun^\policy (\MyState_t^\action)}{\ctrDrift (\MyState_t^\action, \action) - \drift^\policy (\MyState_t^\action)}  \Big] dt.
\end{align*}
Taking $\Delta t \rightarrow 0$, we conclude the proof of this proposition.

\subsection{Proof of \Cref{cor:fourier-example}}\label{subsec:proof-cor-fourier-example}
Recall that the basis functions are given by
\begin{align*}
  \psi (x) = \exp (2 \pi i \inprod{\alpha}{x}) \quad \mbox{for $\vecnorm{\alpha}{1} \leq \numobs$}.
\end{align*}
The cardinality $\mbasis$ of this set of basis function satisfies
\begin{align}
  \Big( \frac{2 n}{\usedim} - 1 \Big)^\usedim \leq \mbasis \leq (2 n + 1)^\usedim.\label{eq:mbasis-upper-lower-in-fourier-example}
\end{align}
Since the stationary distribution $\stationary$ is uniform, we have
\begin{align*}
 H_0 = I_\mbasis , \quad \mbox{and} \quad H_1 = \mathrm{diag} \Big\{ 1 + \vecnorm{\alpha}{2}^2 \Big\}_{\vecnorm{\alpha}{1}\leq n} .
\end{align*}
and consequently, Assumption~\ref{assume:basis-condition} is satisfied with $\omega = 1 / \usedim$, and Assumption~\ref{assume:bounded-feature} is satisfied with $D_\mbasis = c_d \mbasis^{\frac{1}{2} + \frac{1}{\usedim}}$. And we have the bound
\begin{align*}
  \mathrm{Tr} \big( H_1^{-1} H_0 \big) &\leq\sum_{\alpha~:~ \vecnorm{\alpha}{1} \leq n} \frac{1}{1 + \vecnorm{\alpha}{2}^2} \leq   \sum_{\alpha~:~ \vecnorm{\alpha}{1} \leq n} \frac{1}{1 + \vecnorm{\alpha}{1}^2 / \usedim} =  \sum_{j = 0}^{\numobs} \frac{1}{1 + j^2 / \usedim} \abss{\{ \alpha \in \integers^\usedim: \vecnorm{\alpha}{1} = j\}}.
\end{align*}
So we have the upper bound
\begin{align*}
  \mathrm{Tr} \big( H_1^{-1} H_0 \big) \leq c_\usedim \sum_{j = 0}^\numobs \frac{j^{\usedim - 1}}{1 + j^2} \leq c_\usedim \cdot \begin{cases}
    1 & \usedim = 1,\\
    \log \numobs & \usedim = 2,\\
    \numobs^{\usedim - 2} & \usedim \geq 3.
  \end{cases}
\end{align*}
By Equation~\eqref{eq:mbasis-upper-lower-in-fourier-example}, we have
\begin{align*}
  \mathrm{Tr} \big( H_1^{-1} H_0 \big) \leq c_\usedim \begin{cases}
  1 & \usedim = 1,\\
  \log \mbasis & \usedim = 2,\\
  \mbasis^{1 - \frac{ 2}{\usedim}} & \usedim \geq 3.
  \end{cases}
\end{align*}
We use the function $g_\usedim (\mbasis)$ to denote such an upper bound.

For the approximation error, we use the Jackson inequality for trigonometric approximation.
\begin{proposition}[\cite{ganzburg1981multidimensional}]\label{prop:jackson-inequality}
  Let $f$ be an order-$k$ H\"{o}lder function. For any $n \geq 1$, let $\mathcal{T}_n (f)$ be the best $n$-th order trigonometric polynomial approximation to $f$ in the $\sup$-norm. Then there exists a constant $\constScary > 0$ depending on $d$ and $k$ such that
  \begin{align*}
   \sup_{x \in \torus^\usedim } \abss{f (x) - \mathcal{T}_n f (x)} \leq \constScary \frac{1}{n^k}.
  \end{align*}
\end{proposition}
Taking this result as given, let us now bound the approximation error in the Fourier basis. Let $\Pi_n$ be the orthonormal projection operator onto the $n$-th order trigonometric polynomial space. For an order-$k$ H\"{o}lder function $f$, we have
\begin{align*}
  \lpstatnorm{f - \Pi_n f}{p} &\leq \lpstatnorm{f - \mathcal{T}_n (f)}{p} + \lpstatnorm{\mathcal{T}_n (f) - \Pi_n f}{p}\\
  &= \lpstatnorm{f - \mathcal{T}_n (f)}{p} + \lpstatnorm{\Pi_n \big( \mathcal{T}_n (f) - f \big)}{p}\\
  &\leq \vecnorm{f - \mathcal{T}_n (f)}{\infty} \Big\{1 + \sup_{\vecnorm{h}{\infty} \leq 1} \lpstatnorm{\Pi_n (h)}{p} \Big\}.
\end{align*}
For $p \in (1, + \infty)$, the Fourier partial sum operator within a re-scaled convex polytope region satisfies a uniformly strong $(p, p)$ operator norm bound (see~\cite{duoandikoetxea2024fourier}, Section 3.5). So there exists a constant $c > 0$ that depends only on $p$ and $d$, such that 
\begin{align*}
  \sup_{\vecnorm{h}{\infty} \leq 1} \lpstatnorm{\Pi_n (h)}{p} \leq \sup_{\lpstatnorm{h}{p} \leq 1} \lpstatnorm{\Pi_n (h)}{p} \leq c.
\end{align*}
Consequently, we have
\begin{align*}
  \lpstatnorm{f - \Pi_n f}{p} \leq (1 + c) \vecnorm{f - \mathcal{T}_n (f)}{\infty} \leq (1 + c) \constScary n^{- k} \leq c' \mbasis^{- k / \usedim}.
\end{align*}
Applying the same arguments to gradient and Hessian of $f$, we have
\begin{align*}
  \lpstatnorm{\nabla f - \Pi_n (\nabla f)}{p} & \leq (1 + c) \constScary n^{- k + 1} \leq c' \mbasis^{- (k - 1) / \usedim}, \quad \mbox{and}\\
   \lpstatnorm{\nabla^2 f - \Pi_n (\nabla^2 f)}{p} & \leq (1 + c) \constScary n^{- k + 2} \leq  c' \mbasis^{- (k - 2) / \usedim}.
\end{align*}

In order to make use of this result, we need to relate the Fourier projection to the Sobolev projection. Given a function
\begin{align*}
  \ValTrue = \sum_{\alpha \in \integers^\usedim} \thetastar_\alpha \psi_\alpha,
\end{align*}
for any approximating function $\valfun = \sum_{\vecnorm{\alpha} \leq \numobs} \theta_\alpha \psi_\alpha$, we have
\begin{align*}
  \sobonorm{\ValTrue - \valfun}^2 &= \statnorm{\ValTrue - \valfun}^2 + \statnorm{\nabla \ValTrue - \nabla \valfun}^2\\
  &= \sum_{\vecnorm{\alpha}{1} \leq \numobs} (1 + \vecnorm{\alpha}{2}^2) \abss{\thetastar_\alpha - \theta_\alpha}^2 + \sum_{\vecnorm{\alpha} \leq \numobs} (1 + \vecnorm{\alpha}{2}^2) \abss{\thetastar_\alpha}^2.
\end{align*}
The best approximation is given by matching all the Fourier coefficients for multi-indices satisfying $\vecnorm{\alpha}{1} \leq n$, and we have
\begin{align*}
  \projecttolinhil (\ValTrue) = \projecttolin (\ValTrue) = \Pi_\numobs \ValTrue.
\end{align*}
Following this argument, we also note that
\begin{align*}
  \Pi_n (\nabla \ValTrue) = \nabla \projecttolinhil (\ValTrue), \quad \mbox{and} \quad   \Pi_n (\nabla^2 \ValTrue) = \nabla^2 \projecttolinhil (\ValTrue).
\end{align*}
Therefore, we have that
\begin{align*}
  \sobokpstatnorm{\Delstar}{k}{p}^p = \sum_{j = 0}^k \lpstatnorm{\nabla^j \ValTrue - \Pi_n (\nabla^j \ValTrue)}{p}^p, \quad \mbox{for $k = 0,1,2$}.
\end{align*}
By substituting the bound to Eq~\eqref{eq:simplified-bound-bad-case}, we have
\begin{align*}
  \Exs \Big[ \sobonorm{\valuehat_T - \ValTrue}^2 \bm{1}_\Event \Big] &\leq c \mbasis^{- (2 k - 2) / d} + c \frac{\mbasis \log^2 (1 / \stepsize)}{T} \Big\{ \poincare^{-1} \mbasis^{- ( 2k - 3) / \usedim} + \stepsize \mbasis^{- (2k - 4) / \usedim} \Big\} \\
  &\qquad \qquad +  \frac{\tau^4 \constScary}{T} \mathrm{Tr} \big( H_1^{-1} H_0 \big)  \big( \statnorm{\nabla \ValTrue}^2 + \stepsize \big) + \constScary \stepsize^{2 \numerOrder}\\
  &\leq c \mbasis^{- (2 k - 2) / d} + \frac{c'}{T} g_\usedim (\mbasis) + \constScary \stepsize^{2 \numerOrder},
\end{align*}
as long as $T \gtrsim \frac{1}{\poincare} \mbasis^{1 + 2 / \usedim} \log^{3/2} \big( \frac{\mbasis}{\delta \stepsize} \big)$. We therefore completed the proof of \Cref{cor:fourier-example}.

\end{document}